\documentclass[twoside,11pt]{article}

\usepackage{blindtext}

\usepackage{jmlr2e}
\newcommand{\Bplus}{\bar{\mathcal{B}}^+}

\usepackage{amsmath,amsfonts}
\usepackage{cleveref}
\usepackage{mathtools}
\usepackage{algorithmic}
\usepackage{graphicx}
\usepackage{braket}
\usepackage{caption}
\usepackage{subcaption}
\usepackage{cleveref}
\usepackage{breqn}
\usepackage{eucal}
\usepackage{adjustbox}
\usepackage{comment}
\usepackage{todonotes}

\usepackage{etoolbox}
\usepackage{autonum}
\usepackage{enumitem}
\usepackage{amsmath}
\newcommand{\subscript}[2]{$#1 _ #2$}

\DeclareFontFamily{U}{matha}{\hyphenchar\font45}
\DeclareFontShape{U}{matha}{m}{n}{
<-6> matha5 <6-7> matha6 <7-8> matha7
<8-9> matha8 <9-10> matha9
<10-12> matha10 <12-> matha12
}{}
\DeclareSymbolFont{matha}{U}{matha}{m}{n}

\DeclareFontFamily{U}{mathx}{\hyphenchar\font45}
\DeclareFontShape{U}{mathx}{m}{n}{
<-6> mathx5 <6-7> mathx6 <7-8> mathx7
<8-9> mathx8 <9-10> mathx9
<10-12> mathx10 <12-> mathx12
}{}
\DeclareSymbolFont{mathx}{U}{mathx}{m}{n}

\DeclareMathDelimiter{\vvvert} {0}{matha}{"7E}{mathx}{"17}%

\DeclarePairedDelimiterX{\normiii}[1]
{\vvvert}
{\vvvert}
{\ifblank{#1}{\:\cdot\:}{#1}}

\usepackage{lastpage}
\usepackage{subfiles}

\ShortHeadings{Quantum Reservoir Computing and Risk Bounds}{Chmielewski, Amini and Mikael}
\firstpageno{1}

\begin{document}

\title{Quantum Reservoir Computing and Risk Bounds} 

\author{\name Naomi Mona Chmielewski \email naomi-mona.chmielewski@centralesupelec.fr \\
       \addr Université Paris-Saclay \\ CNRS-CentraleSupélec, L2S, France \\
       EDF Lab, France
       \AND
       \name Nina Amini \email nina.amini@centralesupelec.fr \\
       \addr Université Paris-Saclay \\ CNRS-CentraleSupélec, L2S, France
       \AND
       \name Joseph Mikael \email
       joseph.mikael@gmail.com \\
       \addr EDF Lab, France
       }


\maketitle

\begin{abstract}

We propose a way to bound the generalisation errors of several classes of quantum reservoirs using the Rademacher complexity. We give specific, parameter-dependent bounds for two particular quantum reservoir classes. We analyse how the generalisation bounds scale with growing numbers of qubits. Applying our results to classes with polynomial readout functions, we find that the risk bounds converge in the number of training samples. The explicit dependence on the quantum reservoir and readout parameters in our bounds can be used to control the generalisation error to a certain extent. It should be noted that the bounds scale exponentially with the number of qubits $n$. The upper bounds on the Rademacher complexity can be applied to other reservoir classes that fulfill a few hypotheses on the quantum dynamics and the readout function. 

\end{abstract}


\section{Introduction}
\label{sec:introduction}
As an alternative to classical machine learning models which tend to require a high number of parameters and consequently large data sets, quantum machine learning (QML) models for near-term quantum devices present a promising lead for understanding the potential of quantum technologies. Many current near-term QML methods are based on some form of variational quantum circuits and can be used for different learning tasks, such as classification (\cite{schuld:20}), autoencoding (\cite{romero:17}) and generative models (\cite{benedetti:19}). Phenomena like barren plateaus (\cite{mcclean:18}) and general poor trainability (\cite{anschuetz:22}) of variational quantum methods have motivated the search for circuit designs and quantum machine learning models that circumvent these issues. 

\textit{Reservoir computing (RC)} has recently garnered some attention as an energy and data efficient alternative to more mainstream variants of recurrent neural networks (RNN) for time series forecasting problems (\cite{tanaka:19, gilpin:23}). Instead of a large neural network with trainable weights, RC uses a fixed-weight reservoir described by its input-dependent dynamics and a final readout layer, the latter of which is usually a simple model such as a linear or polynomial regression. The RC models that we are concerned with in this paper possess a memory of past inputs and are able to adapt to the dynamics of the input process. While initially conceived as a type of neural network (\cite{jaeger:01}), physical implementations of RC such as the use of silicone octopus arms have become a novel way to introduce different reservoir dynamics (\cite{nakajima:20}). 

Two important properties that we typically require of a class of RC models are the so called \textit{echo state property (ESP)} and the \textit{fading memory property (FMP)}, both of which are characterisations of how the reservoir's ``memory'' of past inputs should behave over time. The ESP ensures that the reservoir forgets its initial conditions in the infinite time limit. Additionally, a reservoir with the FMP will yield similar outputs for two input processes that behaved differently in the distant past but similarly in the recent past.

Recent propositions for \textit{quantum reservoir computing (QRC)} propose to leverage the complex dynamics of quantum systems to reproduce the behaviour of complex input data dynamics. Since the only trainable part of a quantum reservoir is in the classical linear regression, the trainability issues of other QML models mentioned earlier are no longer of concern. As pointed out by \cite{chen:19}, for the quantum system to have ESP, it must be dissipative (see e.g. \cite{fujii:16}). As an example, \cite{suzuki:22} make use of the so called depolarising error. 

\medskip

One of the properties that is often studied in machine learning is a model's ability to generalise on unseen data. We call the \textit{generalisation error} or \textit{risk} of a hypothesis class the expected value of the loss over all possible samples according to some data distribution. Since this distribution is generally unknown, it is bounded through the maximal difference within the class of hypothesis functions between the generalisation error and the empirical error, that is, the average of the model's errors evaluated on a finite number of training samples. Essentially, an upper bound on this quantity can give an idea of how different the loss on a random data sample will be, compared to the empirical risk. This upper bound is often called a risk bound. Ideally, the risk bound should decrease as the number of samples in the empirical risk increases. The risk bound can also give some information on how to scale the parameters of the learning model in order to improve its ability to generalise. 
A common way to bound the risk of a hypothesis class is through the \textit{Rademacher complexity}, which is essentially a measure of the ``richness'' of the class by quantifying how well the class can adapt to random noise. The Rademacher complexity is defined on a finite number of samples, and for a hypothesis class to exhibit good generalisation, we expect the Rademacher complexity to go down as the number of samples goes up. 

If the hypothesis class is universal, the difference between the in-class minimum and the minimum over all measurable functions can be made arbitrarily small. 
Formally, a universal class of RC is defined to be a class such that for any process characterised by a dynamical map with fading memory, there exists a member of the reservoir class that can approximate the fading memory map arbitrarily well. Some popular reservoir classes have been proven to be universal by \cite{grigoryeva:18a, grigoryeva:18b}.
Universal reservoir classes based on quantum dynamics have been analysed by \cite{chen:19, chen:20, sannia:24}. \cite{monzani:24} give universality conditions for reservoir classes based on quantum dynamics. For our analysis, we restrict ourselves to quantum reservoir classes with a fixed number of qubits, similarly to \cite{bu:21}, which is a more realistic setting. This is in contrast to the standard procedure when analysing universality properties of quantum reservoir classes where the number of qubits is generally unbounded.

In this paper, we propose to bound the Rademacher complexity of a general parametrised quantum reservoir class with bounded parameter space equipped with a general readout class, and show how the choice of readout functions impacts the scaling of the Rademacher complexity by upper bounding the latter for three different readout classes. As stated above, the specific classes of quantum reservoirs that we study employ polynomial readouts to prove universality. We show that this choice leads to an unfavourable scaling of the Rademacher complexity in the number of qubits. We show that, using a linear readout function, the scaling can be improved upon, though it remains exponential. Another method to introduce the structure of a polynomial algebra that has been suggested by \cite{sannia:24} and \cite{monzani:24} is spatial multiplexing. We compare the scaling of a polynomial readout and spatial multiplexing.

\cite{gonon:20} propose a certain number of hypotheses which allow them to establish risk bounds for reservoir classes equipped with a general readout class. Using our bounds on the Rademacher complexity, we establish risk bounds for different quantum reservoirs by showing that the dynamics of two popular classes of quantum reservoirs, which are governed by \textit{completely positive trace preserving (CPTP)} maps, verify the aforementioned hypotheses. We provide an analysis of the scaling of the risk bound in the number of training samples

\medskip

 The generalisation abilities of certain parametrised quantum circuit configurations have been studied for example by \cite{bu:21} and \cite{caro:22}, using the Rademacher complexity and covering numbers respectively. The learning problem in these cases is based on independent and identically distributed samples and is thus not immediately applicable to reservoir computing.
 \cite{gonon:23} establish approximation error bounds for trainable variational quantum circuits as well as quantum extreme learning machines (a form of reservoir computing that does not take into account past input states) using the Fourier transform. To our knowledge, risk bounds for the specific case of reservoir computing with quantum dynamics for time series forecasting have not yet been studied.

\medskip

This paper is organised as follows: In \Cref{sec:framework} we introduce the mathematical background and recall some definitions. In \Cref{sec:general_rademacher_bound} we establish some general conditions on the quantum reservoir classes as well as the  input and target processes, and state the first main result, which is an upper bound on the Rademacher complexity of this general quantum reservoir class. In \Cref{sec:rademacher_bound_specific} we establish bounds on the Rademacher complexity for any quantum reservoir that verifies the conditions established in \Cref{sec:general_rademacher_bound} when equipped specifically with a polynomial readout class, as well as for linear and spatial multiplexing readout classes. In \Cref{sec:PTR_RRR} we define slightly modified versions of the QRC classes introduced by \cite{chen:19} and \cite{chen:20} and show that they verify the hypotheses from \Cref{sec:general_rademacher_bound}. Finally, in \Cref{sec:generalisation_bounds} we establish risk bounds for the reservoir classes from \Cref{sec:PTR_RRR} that explicitly depend on the reservoir and readout parameters.

\section{Framework}
\label{sec:framework}
\subsection{General definitions}
\label{sec:general}

We first define some of the notation that will be used in the remainder of the paper, namely the reservoir system and its filter and functional, the input-output processes, the echo state and fading memory property, which are two important properties of a reservoir system, and some norms that we use in the remainder of the document.

\paragraph{Hilbert Space.} For the Hilbert space $\mathcal{H} = \mathbb{C}^{2^n \times 2^n},\, n \in \mathbb{N} $, define the set $\mathcal{B}(\mathcal{H})$ (which we abbreviate as $\mathcal{B}$ in the following) of bounded self-adjoint trace-class linear operators that act on $\mathcal{H}$, as well as the dual space $\mathcal{B}^*(\mathcal{H})$ of all possible linear transformations $T(\mathcal{B}) : \mathcal{B} \mapsto \mathcal{B}$ which we abbreviate as $\mathcal{B}^*$ and which forms a finite Banach space with induced norm $\lVert T \rVert_1 = \underset{A \in \mathcal{B}, A \neq 0}{\text{sup}} \frac{\lVert T(A) \rVert_1}{\lVert A \rVert_1}$.

\paragraph{Density Matrices.} Define the set of quantum \textit{density matrices} \\ $\bar{\mathcal{B}}^+(\mathcal{H}) := \{ \rho \in \mathcal{B} : \rho \geq 0 , \text{tr}[\rho] = 1 \}$, as well as the hyperplane of traceless operators $\mathcal{B}_0(\mathcal{H}) := \left\{ A \in \mathcal{B}: \text{tr} \left[ A \right] = 0 \right\} $ which we abbreviate as $\bar{\mathcal{B}}^+$ and $\mathcal{B}_0$ respectively. In the following we may also call elements in $\bar{\mathcal{B}}^+$ \textit{quantum states}.

\paragraph{Input and Output Sequences.} We now define the classical input and output sequences as well as the quantum reservoir states. Let $\mathbb{Z}_- = \{ \ldots, -2, -1, 0\}$ be the set of all negative integers. We consider the semi-infinite \textit{input sequences} \\ $ \pmb v = ( \ldots, v_{-1}, v_0 ) \in \left( D_v \right)^{\mathbb{Z}_-} \subset \left(\mathbb{R}^{p_v}\right)^{\mathbb{Z}_-}$, where $D_v$ is a compact subset of $\mathbb{R}^{p_v}$ for some $p_v \in \mathbb{N}^*$, and \textit{output sequences} $\pmb y = ( \ldots, y_{-1}, y_0 ) \in \left(\mathbb{R}^{p_y}\right)^{\mathbb{Z}_-}$, $p_y \in \mathbb{N}^*$ as well as \textit{reservoir states} $\pmb\rho = ( \ldots, \rho_{-1}, \rho_0 ) \in \left( \bar{\mathcal{B}}^+ \right)^{\mathbb{Z}_-}$. 

\paragraph{Input and Output Processes.} To model the stochastic processes of the input and output sequences, we consider random variable inputs $\pmb V = (V_t)_{t \in \mathbb{Z}_-}$ and outputs $\pmb Y = (Y_t)_{t \in \mathbb{Z}_-}$ with a causal Bernoulli shift structure, that is, for $ q_v, q_y \in \mathbb{N}^*$ and measurable functions $G^v \colon \left( \mathbb{R}^{q_v} \right)^{\mathbb{Z}_-} \to D_v$ 
and $G^y \colon \left( \mathbb{R}^{q_y} \right)^{\mathbb{Z}_-} \to \mathbb{R}^{p_y}$ of independent and identically distributed random variables $\xi = \left( \left( \xi_t^v, \xi_t^y \right)_{t \in \mathbb{Z}_-} \right)$ with values in $\mathbb{R}^{q_v} \times \mathbb{R}^{q_y}$ such that for any $t \in \mathbb{Z}_-$ we have 
\begin{equation}
\label{eq:bernoulli_shift}
\begin{aligned}
    V_t = G^v(\ldots, \xi_{t-1}^v, \xi_t^v ) \\
    Y_t = G^y(\ldots, \xi_{t-1}^y, \xi_t^y )
\end{aligned} \; .
\end{equation}

We call the $\xi_t^v$ and $\xi_t^y$ innovations. The reason for this choice of input-output processes lies in its relative generality in time series analysis in the reservoir computing framework. It models the temporal dependence of the processes, and the causal structure is adapted to the causal structure of reservoirs. Examples of well-known, common processes that fall into this framework include autoregressive models, and more generally ARMA models, which are used frequently in time series analysis (see for example \cite{chen:95} for applications in load forecasting or \cite{atyabi:16} for the modelling of EEG signals) and ARCH processes (typically used for financial time series, see for example \cite{brooks:08}).

See \cite[Section 3.1]{dedecker:07} for more details on Bernoulli shifts and more examples.

\paragraph{Quantum Reservoir System.} The \textit{reservoir system } is defined by the system of equations

\begin{equation}
\label{eq:reservoir_system}
  \begin{cases}
    \rho_t &= T(v_t, \rho_{t-1}) \\
    y_t &= h(\rho_t)
  \end{cases}
\end{equation}

where $h: \mathcal{B} \to \mathbb{R}^{p_v}$ is called the readout function and $T \colon D_v \times \mathcal{B}  \to \mathcal{B}$ is the \textit{completely positive trace preserving (CPTP)} map that describes the dynamics of the quantum reservoir. A CPTP map is a completely positive linear mapping that preserves the trace of any operator. 

\begin{figure}[!h]
  \centering
  \includegraphics[trim={0 10cm 0 10cm}, clip, width=1\linewidth]{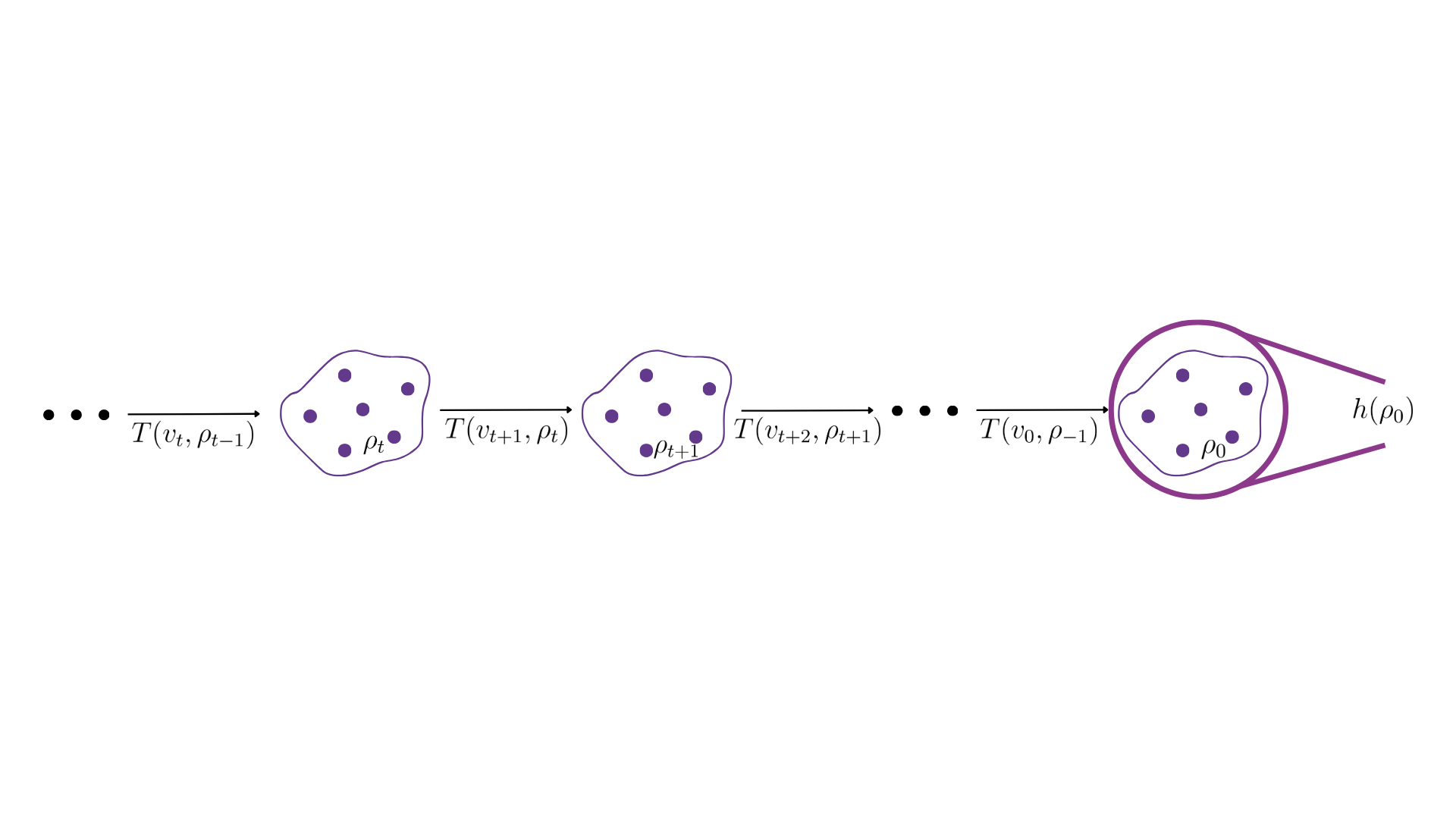}
  \caption{Schematic of the temporal evolution of a quantum reservoir. The purple dots designate qubits, the outline determines the reservoir. At time $t$, the reservoir is in state $\rho_t$. A new input $v_{t+1}$ is injected and the full reservoir is made to evolve according to the dynamics induced by the CPTP map $T$. After all the data has been injected in this way, the final reservoir state $\rho_0$ is processed in a readout function $h$ which produces the prediction.}
  \label{fig:general_quantum_reservoir}
\end{figure}

Conceptually, the reservoir at time $t$ is described by its reservoir state $\rho_t$ and the subsequent input $v_{t+1}$ is injected into the dynamics of the reservoir, which is left to evolve for some unit of time. At time $t+1$, the new reservoir state $\rho_{t+1}$ encodes all of the previously seen inputs, transformed through the reservoir dynamics (see Fig.~\ref{fig:general_quantum_reservoir}).

\paragraph{Echo State Property.} A reservoir system satisfies the \textit{echo state property (ESP) }if for any $\pmb v \in (D_v)^{\mathbb{Z}_-}$, \eqref{eq:reservoir_system} has a unique solution $\pmb\rho \in \left( \bar{\mathcal{B}}^+\right)^{\mathbb{Z}_-}$ when restricted to the space $\Bplus$ of density matrices. This translates the idea that the reservoir dynamics are independent of its initial conditions. 
 
\paragraph{Reservoir Filter and Functional.} In the case where the system satisfies the ESP, it induces a causal and time invariant reservoir filter $M_h^T$ as shown by \cite{grigoryeva:18a} which allows us to describe the input transformation of all past inputs, as opposed to the CPTP map which only describes the transformation between time steps $t-1$ and $t$:

\begin{align}
\label{eq:filter_general}
    M_h^T \colon (D_v)^{\mathbb{Z}_-} &\to \left( \mathbb{R}^{p_y} \right)^{\mathbb{Z}_-} \\
    \pmb v &\mapsto h \circ M^T(\pmb v)
\end{align}
where $M^T(\pmb v) := \left( \ldots, T \left( v_{-1}, \rho_{-2} \right), T \left( v_{0}, \rho_{-1} \right) \right) $, with the corresponding functional

\begin{align}
\label{eq:funtional_general}
    H_h^T \colon (D_v)^{\mathbb{Z}_-} &\to \mathbb{R}^{p_y} \\
    \pmb v &\mapsto h \circ H^T(\pmb v)
\end{align}
where we define $H^T(\pmb v) := M^T(\pmb v)_0$, and the subscript $0$ indicates the last element in the semi-infinite sequence, that is, $M^T(\pmb v)_0 = T \left( v_{0}, \rho_{-1} \right) = \rho_0 $. The subscript $h$ and superscript $T$ are meant to highlight the dependence of the filter and functional on the CPTP map $T$ and the readout function $h$. In the following we will omit this subscript and superscript of the reservoir functional whenever the dependence on $T$ and $h$ is clear from the definitions.

\paragraph{Fading Memory.} Let $\pmb w = \{w_t\}_{t \geq 0}$ be a decreasing positive-valued sequence such that $ \underset{t \to \infty}{\text{lim}} w_t = 0 $. Then we call the norm $\lVert \cdot \rVert_{1, \pmb w}$ defined as $\lVert \pmb v \rVert_{1, \pmb w} := \sum_{t \in \mathbb{Z}_-} \Vert v_t \Vert_2 w_{-t}$ for $\pmb v \in(D_v)^{\mathbb{Z}_-} $ the $(1, \pmb w)$-norm and define the space $\ell_-^{1, w} := \left\{ \pmb v \in \left( \mathbb{R}^{p_v} \right)^{\mathbb{Z}_-} \Big\vert \left\lVert \pmb v \right\rVert_{1, \pmb w} < \infty \right\} $. We say that a reservoir map has $\pmb w$\textit{-fading memory }if for any $\pmb v, \pmb s \in (D_v)^{\mathbb{Z}_-}$, for all $\varepsilon > 0$ there exists $\delta > 0$ such that $ \left\Vert M_h^T(\pmb v)_0 - M_h^T(\pmb s)_0 \right\Vert_2 < \varepsilon $ whenever $ \sup_{t \in \mathbb{Z}_-}  \Vert v_t - s_t \Vert_2 w_{-t} < \delta $ (alternatively, we say that a reservoir map has the fading memory property (FMP) if there exists a decreasing positive-valued weighting sequence $\pmb w$ such that the reservoir map has $\pmb w$-fading memory). This formalises the idea that a reservoir map should focus more on recent inputs and forget the distant past over time. 

\paragraph{Norms.} For an operator $\Phi$ we write $\lVert \Phi \rVert_2 = \sqrt{ \text{tr} [ \Phi^{\dag} \Phi ] }$ for the Schatten-$2$ norm. Note that for a matrix $A$, the Schatten-$2$ norm $\lVert A \rVert_2$ is the same as the Frobenius norm $\lVert A \rVert_F = \sqrt{\sum_{i,j}A_{ij}^2} $ as well as the $L^2$-norm of the vectorised matrix $\lVert \text{vec}(A)\rVert_2=\sqrt{\sum_k (\text{vec}(A))_k^2}$.  For the remainder of the text, we will use the notation $\lVert \cdot \rVert_2$ to denote all three of these norms. This is justified by the fact that any CPTP map $T$ is a linear map, meaning that there exists a matrix $\hat{T}$ such that we can write the linear application as a matrix multiplication $T(A) = \hat{T} A$. 
We also denote by $\normiii{ \cdot }_2$ the operator norm induced by the Schatten-$2$ norm, i.e. $\normiii{ T }_2 :=  \mathop{\text{sup}_{\substack{A \neq 0}}} \frac{\lVert T(A)\rVert_2}{\lVert A \rVert_2} = \mathop{\text{sup}_{\substack{\lVert A \rVert_2 =1}}} \lVert T(A)\rVert_2$, and we write $\normiii[Big]{ T \bigr|_{\mathcal{B}_0}}_2 :=  \mathop{\text{sup}_{\substack{ A \in  \mathcal{B}_0 \\ A \neq 0}}} \frac{\lVert T(A)\rVert_2}{\lVert A \rVert_2} = \mathop{\text{sup}_{\substack{ A \in  \mathcal{B}_0 \\ \lVert A \rVert_2 = 1}}} \lVert T(A)\rVert_2$.

\subsection{Empirical Risk and Rademacher Complexity}
\label{sec:empirical_risk}

In this subsection we formally define the generalisation error, which measures the ability of the reservoir to generalise on unseen data, and then introduce the empirical risk. We also define the Rademacher complexity, the measure that we use to bound the generalisation error. 

We consider loss functions of the form $ \ell( \hat{y}, y ) = \sum_{i=1}^{p_y} f_i( \hat{y}_i - y_i) $ where the $f_i : \mathbb{R} \to \mathbb{R}$ are $L_{\ell}/\sqrt{p_y}$-Lipschitz-continuous and such that $f_i(0) = 0$. Once again, this condition is necessary for the proofs of the theorems in \Cref{sec:generalisation_bounds}. Several of the commonly used loss functions are Lipschitz-continuous, such as the mean absolute error, the Huber loss and the hinge loss. For any hypothesis functional $H \colon D_v \to \mathbb{R}^{p_y}$, the generalisation error is given by 
\begin{equation}
    R(H) := \mathop{\mathbb{E}_{\substack{(\pmb V, \pmb Y) \sim P}}}  [ \ell( H(\pmb V), Y_0 ) ] 
\end{equation}
where $P$ is the joint distribution of $\pmb V$ and $\pmb Y$. As this quantity is generally intractable since $P$ is unknown, the standard procedure to bound the generalisation error is to uniformly bound the maximal difference between the true generalisation error and the empirical risk error for some hypothesis class. The empirical risk error $\hat{R}_m(H)$ of the hypothesis functional $H$ is typically defined on $m$ independent and identically distributed samples. In the context of time series, we do not usually have such i.i.d. samples; instead, a common way to define the empirical risk is to consider the $m$ subsequences $ (V_{-m+1}, \ldots, V_{-i-1}, V_{-i})$ for $i = 0$ to $m-1$ of the input process to construct $m$ samples. Note that the reservoir filter is defined on an infinite sequence, so that we must append the subsequence with zeros. We thus define $\pmb V_{-m+1:-i} := (\ldots, 0, 0, V_{-m+1}, \ldots, V_{-i-1}, V_{-i})$ and the empirical risk of hypothesis $H$ 
\begin{equation}
    \hat{R}_m(H) := \frac{1}{m} \sum_{i=0}^{m-1} \ell ( H(\pmb V_{-m+1:-i}), Y_{-i} ) \ .
\end{equation}

The goal of this paper is to bound the quantity
\begin{equation}
\label{eq:risk_difference}
\begin{aligned}
    \sup_{H \in \mathcal{H} } \left| R(H) - \hat{R}_m(H) \right|
\end{aligned}
\end{equation}
for specific classes $\mathcal{H}$ of quantum reservoirs with dependencies in the parameters of the CPTP and readout maps.

One common way to bound \eqref{eq:risk_difference} is to find a bound that depends on the Rademacher complexity, and to subsequently bound the Rademacher complexity. In the case of time series, the standard definition of the Rademacher complexity needs to be adapted. Usually, the Rademacher complexity is defined on $k$ independent identically distributed samples. In our case, we only have one sample (the time series). We thus make use of the version defined in \cite[Section 4.1]{gonon:20} which introduces ghost samples, which are ``imagined copies'' of the input process: For $k \in \mathbb{N}$, denote by $\Tilde{\pmb V}^{(0)}, \ldots, \Tilde{\pmb V}^{(k-1)}$ independent copies of $\pmb V$, then the Rademacher complexity of the hypothesis class $\mathcal{H}$ is defined as
\begin{equation}
    \mathcal{R}_k(\mathcal{H}) = \frac{1}{k}\mathbb{E} \left[ \sup_{H \in \mathcal{H}} \left\lVert \sum_{j=0}^{k-1} \varepsilon_j H \left( \Tilde{\pmb V}^{(j)} \right) \right\rVert_2 \right]
\end{equation}
where $\varepsilon_0, \ldots, \varepsilon_{k-1}$ are independent and identically distributed Rademacher random variables, independent of $\Tilde{\pmb V}^{(0)}, \ldots, \Tilde{\pmb V}^{(k-1)}$.

\section{A general class of quantum reservoirs and an upper bound on its Rademacher complexity} 
\label{sec:general_rademacher_bound}

In this section, we establish general conditions for the quantum reservoir classes that we wish to analyse and provide a bound on the Rademacher complexity of this general class. The conditions that we require of the reservoir map are notably contractivity in the space of quantum density matrices and Lipschitz-continuity in the input space, whereas the conditions on the readout map are mostly related to boundedness and Lipschitz-continuity. We also require Lipschitz-continuity of both CPTP and readout maps in the parameters.

To be able to bound the generalisation error of the reservoir classes that we describe in \Cref{sec:PTR_RRR}, we first recall the assumptions on the data distribution introduced by \cite{gonon:20}:

\begin{enumerate}[label=(\subscript{A^{IO}}{{\arabic*}})]
    \item For $I = y, v$ the functional $G^I$ as defined in \eqref{eq:bernoulli_shift} is $L_I$-Lipschitz continuous when restricted to $(\ell_{-}^{1,w^I}(\mathbb{R}^{q_I}), \lVert \cdot \rVert_{1, \pmb w^I} )$ for some strictly decreasing weighting sequence $\pmb w^I : \mathbb{N} \to (0,1]$ with finite mean, that is, $\sum_{j \in \mathbb{N}} j w_j^I < \infty $. More specifically, there exists $L_I > 0$ such that for all $\pmb u^I = ( u_t^I )_{t \in \mathbb{Z}_-} \in \ell_{-}^{1, \pmb w^I}(\mathbb{R}^{q_I})$ and $\pmb s^I = ( s_t^I )_{t \in \mathbb{Z}_-} \in \ell_{-}^{1, \pmb w^I}(\mathbb{R}^{q_I})$ it holds that 
    \begin{equation}
        \left\Vert G^I(\pmb u^I)-G^I(\pmb s^I) \right\Vert_2 \leq L_I \lVert \pmb u^I - \pmb s^I \rVert_{1, \pmb w^I} \ .
    \end{equation}
    \item Additionally, let the innovations in \eqref{eq:bernoulli_shift} satisfy $\mathbb{E}\left[ \left\lVert \xi_0^I \right\rVert_2 \right] < \infty$ for $I = y, v$. 
    \item  We also assume that $D_{\pmb w^y} := \sup_{i \in \mathbb{N}} \frac{w_{i+1}^y}{w_i^y} < 1$ and $D_{\pmb w^v} := \sup_{i \in \mathbb{N}} \frac{w_{i+1}^v}{w_i^v} < 1$ and that there exists $M_{\xi} > 0$ such that  for all $t \in \mathbb{Z}_- $, $ \lVert \xi_t^I \rVert_2 < M_{\xi} \ ,\ $ where $I = v, y $. 
\end{enumerate}

We designate the set of conditions on the input-output processes as \\ $A^{IO} = \left\{ (A^{IO}_1), (A^{IO}_2), (A^{IO}_3) \right\}$.

Assumption $(A^{IO}_1)$ essentially imposes a restriction on the strength of the dependence of the process on past innovations. The faster the weighting sequence decreases, the weaker the influence of past innovations. Assumption $(A^{IO}_3)$ imposes a regularity constraint on the speed at which the influence of past innovations decreases.

Consider for example the infinite moving average process, defined as $V_t := \sum_{j=0}^{\infty} a_j \xi_{t-j}$, for some infinite sequence $\left\{ a_j \right\}_{j \geq 0}$, and assume $a_j \in (0,1)$ for all $j \geq 0$. We can identify $G^v(\pmb u^v) = \sum_{j=0}^{\infty} a_j u_{-j}^v $ and write $\left\Vert G^v(\pmb u^v) - G^v(\pmb s^v) \right\Vert_2 \leq \sum_{j=0}^{\infty} a_j \left\Vert u_{-j}^v - s_{-j}^v \right\Vert_2$. It is clear that this process verifies assumption $(A^{IO}_1)$ as long as $\left\{ a_j \right\}_{j \geq 0}$ decreases faster than the weighting sequence $\pmb w^v$ and assumption $(A^{IO}_3)$ is verified if we set $a_j = \lambda^j, j \geq 0$, for some $\lambda \in (0,1)$. In this case, we can identify the weighting sequence as $\pmb w = \left\{ \lambda^j \right\}_{j \geq 0}$ and we verify $\sum_{j=0}^{\infty} j w_j^v = \lambda / (1 - \lambda)^2 < \infty$ as well as $w_{j+1}/w_j = \lambda < 1$.
In particular, this applies to the linear autoregressive model of order one, described as $V_t := \lambda V_{t-1} + \xi_t$.

Additionally, suppose that the inputs are in some set $S \subset \left(D_v \right)^{\mathbb{Z}_-} \cap \left( \ell_{-}^{1,w^v}(\mathbb{R}) \right)$, and suppose that the reservoir system in \eqref{eq:reservoir_system} has a solution in $ S \times \left( \Bplus \right)^{\mathbb{Z}_-}$. Then, if the reservoir map is Lipschitz-continuous in the inputs and strictly contractive in the reservoir states, the associated reservoir functional verifies Assumption $(A^{IO}_1)$ as shown by \cite[Example 1]{gonon:20}.

\subsection{A General Class of Quantum Reservoirs}
\label{sec:general_class}
We begin by defining a general class of quantum reservoirs and readout maps.
First, we suppose that the parameter space $\Theta$ which parametrises the quantum reservoir maps is a bounded subset of $\mathbb{R}^{\text{dim}(\Theta)}$ for finite $\text{dim}(\Theta)$:

\begin{enumerate}[label=(\subscript{A^{\Theta}}{{}})]
    \item We require the parameter space $\Theta$ to be a bounded subset of $\mathbb{R}^{\text{dim}(\Theta)}$ for some $\text{dim}(\Theta) < \infty$. In the following, we write $D_{\Theta} := \sup \left\{ \left\Vert \theta_1 - \theta_2 \right\Vert_2; \ \theta_1, \theta_2 \in \Theta \right\}$ for the diameter of $\Theta$.
\end{enumerate}

We then define, for a fixed number $n$ of qubits, the general class $\mathcal{F}_n^{\text{QRC}} (\Theta)$ of parametrised quantum reservoir maps on an $n$-qubit system with parameters in $\Theta$ as the set of CPTP maps $T^{\theta}$ that map an input $v \in D_v$ as well as a self-adjoint trace class linear operator $A$ to another self-adjoint trace class linear operator, and that meets the two following conditions:

\begin{enumerate}[label=(\subscript{A^T}{{\arabic*}})]
    \item We require all CPTP maps to be $L_R$-Lipschitz-continuous on the compact input space $D_v \subset \mathbb{R}^{p_v} $, that is, there exists a constant $0 < L_R$ such that for all $T^{\theta} \in \mathcal{F}_n^{\text{QRC}} (\Theta)$, for any fixed density operator $\rho \in \bar{\mathcal{B}}^+$ and any fixed parameter $\theta \in \Theta$, for all inputs $v_1, v_2 \in D_v$ we have $\left\lVert T^{\theta}(v_1, \rho) -  T^{\theta}(v_2, \rho) \right\rVert_2 \leq L_R \left\Vert v_1 - v_2\right\Vert_2$.
    \item We further require all CPTP maps to be $r$-contractions in the space of density operators, that is, there exists a constant $0 < r < 1$ such that for all $T^{\theta} \in \mathcal{F}_n^{\text{QRC}} (\Theta)$, for any fixed $v \in D_v$ and for any fixed $\theta \in \Theta$, for all $\rho_1, \rho_2 \in \bar{\mathcal{B}}^+ $, we have \\ $\left\lVert  T^{\theta}(v, \rho_1) -  T^{\theta}(v, \rho_2) \rVert_2 \leq r \lVert \rho_1 - \rho_2 \right\rVert_2$. Note that this condition implies that the reservoir system has ESP and FMP according to \cite[Proposition 1]{pena:23}.
    \item We additionally require all CPTP maps to be $L_{\Theta}$-Lipschitz-continuous in the compact parameter space $\Theta$, that is, there exists a constant $ 0 < L_{\Theta}$ such that for any fixed $v \in D_v$ and for any fixed $\rho \in \bar{\mathcal{B}}^+$, for all $\theta_1, \theta_2 \in \Theta $, we have $\lVert  T_{\theta_1}(\rho, v) -  T_{\theta_2}(\rho, v) \rVert_2 \leq L_{\Theta} \lVert \theta_1 - \theta_2 \rVert_2$.
\end{enumerate}
We designate the set of conditions on the CPTP maps as $A^T = \left\{ (A^T_1), (A^T_2), (A^T_3) \right\}$.

For a CPTP map $T^{\theta}$ that verifies condition $(A^T_2)$ we can define \\ $\overrightarrow{ \prod}_{i=0}^{\infty} T^{\theta} (v_{-i})\rho_{-\infty} := \text{lim}_{N \to \infty} T^{\theta}( v_0 , T^{\theta}( v_{-1}, \ldots T^{\theta} (v_{-N} , \rho_{-N} ) ) ) $. The following lemma shows why this limit always exists:

\begin{lemma}
\label{lemma:chen19}
    Let $T$ be a CPTP map that verifies $(A^T_2)$. Then, for any input sequence $\pmb v$ and any initial state $\rho_{-\infty}$, the infinite composition of the CPTP map
    \begin{align}
        \overrightarrow{ \prod}_{i=0}^{\infty} T (v_{-i})\rho_{-\infty} := \text{lim}_{N \to \infty} T( v_0 , T( v_{-1}, \ldots T (v_{-N} , \rho_{-N} ) ) ) 
    \end{align}
    exists and converges to the state $\overrightarrow{ \prod}_{i=0}^{\infty} T (v_{-i})\left( \frac{I}{2^n} \right)$.
\end{lemma}
The proof is provided in Appendix \ref{app:lemma_convergence}.

We can thus set the initial reservoir state to be the completely mixed state, i.e. $\rho_{-\infty} = \frac{I}{2^n}$, which makes the reservoir maps in \eqref{eq:general_qrc_functional} independent of the initial condition. This means that our analyses are independent of the initial state, and we can establish general results. 

We can write the infinite composition of reservoir maps $T^{\theta}$ as 

\begin{equation}
\label{eq:general_qrc_functional}
        H^{T^{\theta}}(\pmb v) := \overrightarrow{ \prod}_{i=0}^{\infty} T^{\theta} (v_{-i}) \rho_{-\infty} \ . 
\end{equation}

With the assumptions $(A^T_2)$ and $(A^T_3)$, it follows that the infinite composition of the CPTP map $T$ is also Lipschitz-continuous with respect to $\theta$:

\begin{lemma}
\label{lemma:lipschitz_infinite_comp}
    Let $T: \Theta \times D_v \times \bar{\mathcal{B}}^+ \to  \bar{\mathcal{B}}^+$ be a completely positive trace preserving map such that assumptions $(A^T_2)$ and $(A^T_3)$ hold. Then, the infinite composition of CPTP maps is Lipschitz-continuous w.r.t. $\theta$, with Lipschitz constant
   $ 
        L'_{\Theta} = \frac{L_{\Theta}}{1-r} \ .
   $ 
    Note in particular that if all $T$ in the hypothesis class of reservoir CPTP maps have the same Lipschitz and contractivity constant, then all infinite compositions generated by the hypothesis class of reservoir CPTP maps have the same Lipschitz constant.
\end{lemma}
\begin{proof}
     For all $\theta, \theta' \in \Theta$, for any fixed $\rho_{-\infty} \in \bar{\mathcal{B}}^+$ and for any fixed $\pmb v \in \left( D_v \right)^{\mathbb{Z}_-}$ we have
    \begin{align}
        & \left\lVert T_{\theta} \left( v_0, T_{\theta} \left( v_{-1}, \ldots, T_{\theta} \left( v_{-\infty}, \rho_{-\infty} \right)  \right) \right) -  T_{\theta'} \left( v_0, T_{\theta'} \left( v_{-1}, \ldots, T_{\theta'} \left( v_{-\infty}, \rho_{-\infty}\right)  \right) \right) \right\rVert_2 \\
        & = \Big\lVert T_{\theta} \left( v_0, T_{\theta} \left( v_{-1}, \ldots, T_{\theta} \left( v_{-\infty}, \rho_{-\infty} \right)  \right) \right) -  T_{\theta'} \left( v_0, T_{\theta} \left( v_{-1}, \ldots, T_{\theta} \left( v_{-\infty}, \rho_{-\infty} \right)  \right) \right) \\
        & + T_{\theta'} \left( v_0, T_{\theta} \left( v_{-1}, \ldots, T_{\theta} \left( v_{-\infty}, \rho_{-\infty} \right)  \right) \right) -  T_{\theta'} \left( v_0, T_{\theta'} \left( v_{-1}, \ldots, T_{\theta'} \left( v_{-\infty}, \rho_{-\infty} \right)  \right) \right) \Big\rVert_2 \\
        & \leq L_{\Theta} \left\lVert \theta - \theta' \right\rVert_2 + r \left\lVert T_{\theta} \left( v_{-1}, \ldots, T_{\theta} \left( v_{-\infty}, \rho_{-\infty} \right)  \right) - T_{\theta'} \left( v_{-1}, \ldots, T_{\theta'} \left( v_{-\infty}, \rho_{-\infty} \right)  \right)  \right\rVert_2 \\
        & \leq L_{\Theta} \left\lVert \theta - \theta' \right\rVert_2 \\
        & \hspace{3mm} + r \left( L_{\Theta} \left\lVert \theta - \theta' \right\rVert_2 + r \left\lVert T_{\theta} \left( v_{-2}, \ldots, T_{\theta} \left( v_{-\infty}, \rho_{-\infty} \right)  \right) - T_{\theta'} \left( v_{-2}, \ldots, T_{\theta'} \left( v_{-\infty}, \rho_{-\infty} \right)  \right) \right\rVert_2 \right) \\
        & \leq \ldots \leq L_{\Theta} \sum_{\ell = 0}^{\infty} r^\ell \left\lVert \theta - \theta' \right\rVert_2 = \frac{L_{\Theta}}{1-r} \left\lVert \theta - \theta' \right\rVert_2
    \end{align}
where we used the triangle inequality as well as the Lipschitz-continuity w.r.t. $\theta$ and the contractivity w.r.t. $\rho$ in the first inequality, and iterated over the infinite composition.
\end{proof}

\begin{remark} 
\label{remark:separability_T}
    In general, we also require the class $\mathcal{F}_n^{\text{QRC}} (\Theta)$ to be separable in the space of bounded continuous functions when equipped with the supremum norm so that we can conclude that the supremum over the reservoir class is measurable. In the case of quantum reservoirs, the classes that we consider are always subsets of the set of bounded operators that act on the finite-dimensional Hilbert space $\mathcal{H}$ with fixed number of qubits $n$, which itself is a vector space and becomes a finite-dimensional metric space when equipped with the supremum norm.
\end{remark}

We denote by $\mathcal{F}_n^O(\Omega)$ a general class of parametrised readout functions $h_{\omega} : \bar{\mathcal{B}}^+ \cup 0 \to \mathbb{R}$ on an $n$-qubit system (where 0 denotes the $2^n \times 2^n$ matrix with all zeros) with parameters in $\Omega$ that verifies the following conditions: 

\begin{enumerate}[label=(\subscript{A^h}{{\arabic*}})]
    \item $\mathcal{F}_n^O(\Omega)$ contains the zero function.
    \item $\mathcal{F}_n^O(\Omega)$ is separable in the space of bounded continuous functions when equipped with the supremum norm.
    \item The maps $h_{\omega}$ in $\mathcal{F}_n^O(\Omega)$ are $L_h$-Lipschitz continuous in the inputs where there exists $\Bar{L}_h>0 $ such that $L_h \leq \Bar{L}_h $ for all $h \in \mathcal{F}_n^O$.
    \item There exists $L_{h,0} > 0$ such that $|h_{\omega}(0)| \leq L_{h,0}$.
    \item The parameter space $\Omega$ is a bounded subset of $\mathbb{R}^{\text{dim}(\Omega)}$ for some $\text{dim}(\Omega) < \infty$. In the following we write $D_{\Omega}:= \sup \left\{ \left\Vert \omega_1 - \omega_2 \right\Vert_2; \ \omega_1, \omega_2 \in \Omega \right\}$ for the diameter of $\Omega$.
    \item The readout functions are $L_{\Omega}$-Lipschitz-continuous in $\Omega$, that is, there exists a constant $0 < L_{\Omega}$ such that for all $\rho \in \bar{\mathcal{B}}^+ \cup 0$, for all $\omega_1 , \omega_2 \in \Omega$ we have \\ $\left\vert h_{\omega_1}(\rho) - h_{\omega_2}(\rho) \right\vert \leq L_{\Omega} \left\Vert \omega_1 - \omega_2 \right\Vert_2 $.
\end{enumerate}
We designate the set of conditions on the readout maps as $A^h = \left\{ (A^h_1), (A^h_2), (A^h_3), (A^h_4), (A^h_5), (A^h_6) \right\}$.

Finally, we define the general class of reservoir functionals as
\begin{equation}
\label{eq:general_reservoirclass}
    \mathcal{H}_{n}^{\text{QRC}}:= \left\{ H : (D_v)^{\mathbb{Z}_-} \to \mathbb{R} \: | \: H(\pmb v )= h \left(  H^{T^{\theta}} (\pmb v) \right), h \in \mathcal{F}_n^O(\Omega), T^{\theta} \in \mathcal{F}_n^{\text{QRC}} (\Theta) \right\} \ .
\end{equation}

We can bound the Rademacher complexity of this general class:
\begin{theorem}
\label{thm:general_rademacher_dudley}
    The class $\mathcal{H}_{n}^{\text{QRC}}$ has Rademacher complexity bounded by
    \begin{align}
    \label{eq:rademacher_general_bound_thm}
        \mathcal{R}_k\left( \mathcal{H}_{n}^{\text{QRC}} \right) \leq \frac{48 \cdot 0.89 \left( \sqrt{\text{dim} (\Theta)} + \sqrt{\text{dim} (\Omega)} \right) \cdot \max\left( \Bar{L}_h L_{\Theta}' D_{\Theta} \, , \, D_{\Omega} L_{\Omega} \right)} {\sqrt{k} } \ .
    \end{align}
\end{theorem}
The proof is provided in Appendix \ref{app;thm_general_rademacher_dudley}.

\begin{remark}
    Here we have restricted the set of readout functions to one-dimensional outputs for ease of notation and readability of the proofs. 
    It is easy to extend Theorem \ref{thm:general_rademacher_dudley} to the multi-dimensional output case $h_{\omega} \colon \Bplus \cup 0 \to \mathbb{R}^{p_y} \, , \, p_y \geq 1$.
    In this case, it suffices to bound the euclidean norm by the 1-norm at the very beginning of the proof of Theorem \ref{thm:general_rademacher_dudley}, writing
    \begin{align}
        \sup_{\theta \in \Theta, \omega \in \Omega} \left\Vert \sum_{j=0}^{k-1} \varepsilon_j h_{\omega} \left( \rho_0^{\theta, j} \right) \right\Vert_2 & \leq \sup_{\theta \in \Theta, \omega \in \Omega} \sum_{\ell=1 }^{p_y} \left\vert \sum_{j=0}^{k-1} \varepsilon_j h_{\omega}^{\ell} \left( \rho_0^{\theta, j} \right) \right\vert \\
        & \leq  \sum_{\ell=1 }^{p_y} \sup_{\theta \in \Theta, \omega \in \Omega}\left\vert \sum_{j=0}^{k-1} \varepsilon_j h_{\omega}^{\ell} \left( \rho_0^{\theta, j} \right) \right\vert \ , 
    \end{align}
    where we write $ h_{\omega}^{\ell} \colon \Bplus \cup 0 \to \mathbb{R}$ for the $\ell$-th component function of the $p_y$-dimensional vector-valued function, and to proceed exactly in the same way for the process
    \begin{equation}
        \left( Z_{\theta, \omega}^k \right)_{\ell} := \sum_{j=0}^{k-1} \varepsilon_j h_{\omega}^{\ell} \left( \rho_0^{\theta, j} \right)
    \end{equation}
    as we did for $Z_{\theta, \omega}^k$. (Note that, if $h_{\omega}$ is Lipschitz-continuous, then so is $ h_{\omega}^{\ell}$.) This will incur a multiplicative factor of $p_y$ in the bound in \eqref{eq:rademacher_general_bound_thm}.
\end{remark}
In order to understand how the bound in Theorem \ref{thm:general_rademacher_dudley} scales for different choices of readout functions, which impacts the values of $\sqrt{\text{dim}(\Omega)}$, $\bar{L}_h$, $D_{\Omega}$ and $L_{\Omega}$, we will examine these values for three commonly employed readout functions in the following section.

\section{Bounds on the Rademacher Complexity for Specific Readout Functions}
\label{sec:rademacher_bound_specific}
We now move on to analyse the Rademacher complexity of quantum reservoirs with specific classes of readout functions. We first begin by defining the polynomial readout class since it has been used multiple times to prove universality of quantum reservoirs (see e.g. \cite{chen:19, chen:20}). We find a bound on the Lipschitz constants of the members of the class, which allows us to establish a general bound on the Rademacher complexity of all reservoir classes which employ this readout class, and where the reservoir maps verify the hypotheses highlighted in \Cref{sec:general_rademacher_bound}. Additionally, we analyse the Rademacher complexity of two alternative readout classes, namely a linear readout and spatial multiplexing and compare them to the Rademacher complexity of the polynomial readout.

\subsection{Polynomial Readout Map}
\label{sec:readout_map}

The polynomial nature of the readout class used in the reservoir classes by \cite{chen:19} and \cite{chen:20} is used to establish universality of the quantum reservoir classes. In fact, a common way to prove universality in the case of quantum reservoirs is to evoke the Stone-Weierstrass theorem, which, amongst other things, requires the class to be an algebra. The polynomial readout class generates a polynomial algebra. 

We define a class of multivariate polynomial readout functions with a slight modification to the one mentioned above, specifically with a maximal degree and a bound on the absolute value of the constant bias: 

\begin{equation}
\label{eq:universalReadoutClass}
\begin{aligned}
        & \mathcal{F}_{n, R_{\max}, C_{\max}}^{\text{poly}} = \Big\{h_{C, \pmb w}: \Bplus \cup 0 \rightarrow \mathbb{R} \: | \:  \\
        & \hspace{5mm} h_{C, \pmb w}(\rho) = C + \sum_{d=1}^{R_{\text{max}}}\sum_{i_1=1}^n\sum_{i_2=i_1+1}^n \cdots \sum_{i_n = i_{n-1}+1}^n \sum_{r_{i_1}+\cdots + r_{i_n}=d} w_{i_1,\ldots,i_n}^{r_{i_1},\ldots,r_{i_n}} \text{tr}\left[ \sigma_Z^{(i_1)} \rho \right]^{r_{i_1}} \cdots \text{tr}\left[ \sigma_Z^{(i_n)}\rho \right]^{r_{i_n}}, \\
        & \hspace{5mm} 0 \leq |C| \leq C_{\max}, w_{i_1,\ldots,i_n}^{r_{i_1},\ldots,r_{i_n}} \in[0,1] \Big\}
\end{aligned}
\end{equation} 
where $\text{tr}\left[ \sigma_Z^{(i)} \rho \right]$ denotes the expectation of the projective measurement along the $Z$-axis on the $i$-th qubit; $C_{\max}$ is a bound on the maximal magnitude of the constant bias $C$; $R_{\max}$ and the $r_i$ are positive valued integers; $R_{\max}$ is a bound on the degree of the polynomials; and the $w_{i_1,\ldots,i_n}^{r_{i_1},\ldots,r_{i_n}}$ are real valued weight parameters, and we have written $\pmb w = \left\{ w_{i_1,\ldots,i_n}^{r_{i_1},\ldots,r_{i_n}} \right\} $. The reason for this modification of the readout function is analogous to the reason for the restriction in the number of qubits.

Note that this is a class of polynomials with a constant bound on the maximal degree, meaning that it is a finite dimensional vector space so that it is separable and assumption $(A^h_2)$ in \Cref{sec:general_class} is immediately verified. Additionally, it is clear that this class contains the zero function by setting all parameters to zero, so that assumption $(A^h_1)$ is also verified.  We also have $\left| h_{C, \pmb w}(0) \right| = | C | \leq C_{\max}$ for all $h_{C, \pmb w} \in \mathcal{F}_{n, R_{\max}, C_{\max}}^{\text{poly}}$ meaning that $(A^h_4)$ is verified. $\left(A_5^h\right)$ is easily verified by identifying the parameter space of $\mathcal{F}_{n, \text{R}_{\max}, \text{C}_{\text{max}}}^{\text{poly}}$, which we call $\Omega_{\text{poly}}$ as $\Omega_{\text{poly}} = \left[-C_{\max}, C_{\max} \right] \times \left[0, 1\right]^{\mathcal{N}_{\text{poly}}}$, where we have written $\mathcal{N}_{\text{poly}} < \infty$ for the total number of weights, which we will see in the proof of Lemma \ref{lemma:lipschitz_poly} is equal to $\mathcal{N}_{\text{poly}} = \binom{n+R_{\max}}{R_{\max}} - 1$. We can also calculate the diameter of $\Omega_{\text{poly}}$:
\begin{align}
\label{eq:diameter_omega_poly}
    D_{\Omega}^{\text{poly}} = \sqrt{\sum_{i=1}^{\mathcal{N}_{\text{poly}}} (1-0)^2 + \left(2 C_{\text{max}} \right)^2} = \sqrt{\mathcal{N}_{\text{poly}} + 4 C_{\text{max}}^2} < \infty \ .
\end{align}

It remains to show assumptions $(A^h_3)$ and $(A_6^h)$, and to find the explicit expressions for the Lipschitz constants, which we do in the following propositions:
\begin{lemma}
\label{lemma:lipschitz_poly}
    The polynomial readout class
    $\mathcal{F}_{n, R_{\max}, C_{\max}}^{\text{poly}}$ introduced in \eqref{eq:universalReadoutClass}, with a fixed number $n \in \mathbb{N}$ of qubits, a maximal degree $R_{\max}$ of the polynomial readout, and a bound $C_{\max}$ on the maximal absolute value of the constant term in the readout verifies assumption $(A^h_3)$, that is, all maps $h_{C, \pmb w} \in \mathcal{F}_{n, R_{\max}, C_{\max}}^{\text{poly}}$ are Lipschitz-continuous w.r.t. the input, with bounded Lipschitz constants $L_h \leq \Bar{L}_h^{\text{poly}}$, where we write
    \begin{equation}
    \label{eq:L_h_Bar_poly}
        \Bar{L}_h^{\text{poly}} := n \sqrt{2^n}  R_{\max} \cdot \mathcal{N}_{\text{poly}} = n \sqrt{2^n}  R_{\max} \cdot \left( \binom{n+R_{\max}}{R_{\max}} - 1 \right) \ .
    \end{equation}
\end{lemma}
The proof is provided in Appendix \ref{app:lemma_lipschitz_poly}.

\begin{lemma}
\label{lemma:lipschitz_poly_omega}
    The polynomial readout class
    $\mathcal{F}_{n, R_{\max}, C_{\max}}^{\text{poly}}$ introduced in \eqref{eq:universalReadoutClass}, with a fixed number $n \in \mathbb{N}$ of qubits, a maximal degree $R_{\max}$ of the polynomial readout, and a bound $C_{\max}$ on the maximal absolute value of the constant term in the readout verifies assumption $(A^h_6)$, that is, all maps $h_{C, \pmb w} \in \mathcal{F}_{n, R_{\max}, C_{\max}}^{\text{poly}}$ are Lipschitz-continuous w.r.t. the parameters $C$ and $\pmb w$, with Lipschitz constants $L_{\Omega}^{\text{poly}}$, where we write $L_{\Omega}^{\text{poly}} := \sqrt{ \mathcal{N}_{\text{poly}} + 1} = \sqrt{\binom{n+R_{\max}}{R_{\max}}}$.
\end{lemma}
\begin{proof}
    For ease of notation, we write $x_{r_{i_j}} = \text{tr}\left[ \sigma_Z^{(i_j)} \rho \right]^{r_{i_j}}$.
    For any fixed $\rho \in \Bplus \cup 0 \rightarrow \mathbb{R}$, for all $(C, \pmb w), (\tilde{C}, \tilde{\pmb w}) \in \Omega_{\text{poly}}$, we have
    \begin{align}
        & \left\vert h_{C, \pmb w} (\rho) - h_{\tilde{C}, \tilde{\pmb w}} (\rho) \right\vert \\
        &= \left\vert C - \tilde{C} + \sum_{d=1}^{R_{\text{max}}}\sum_{i_1=1}^n\sum_{i_2=i_1+1}^n \cdots \sum_{i_n = i_{n-1}+1}^n \sum_{r_{i_1}+\cdots + r_{i_n}=d} \left( w_{i_1,\ldots,i_n}^{r_{i_1},\ldots,r_{i_n}} - \tilde{w}_{i_1,\ldots,i_n}^{r_{i_1},\ldots,r_{i_n}}  \right) x_{r_{i_1}} \cdots x_{r_{i_n}} \right\vert \\
        & \leq \left\vert C - \tilde{C} \right\vert +  \sum_{d=1}^{R_{\text{max}}}\sum_{i_1=1}^n\sum_{i_2=i_1+1}^n \cdots \sum_{i_n = i_{n-1}+1}^n \sum_{r_{i_1}+\cdots + r_{i_n}=d} \left\vert\left( w_{i_1,\ldots,i_n}^{r_{i_1},\ldots,r_{i_n}} - \tilde{w}_{i_1,\ldots,i_n}^{r_{i_1},\ldots,r_{i_n}}  \right)\right\vert \left\vert x_{r_{i_1}} \cdots x_{r_{i_n}} \right\vert \\
        & \leq \left\Vert (C, \pmb w) - (\tilde{C}, \tilde{\pmb w}) \right\Vert_1 \leq  \sqrt{\mathcal{N}_{\text{poly}} + 1 } \left\Vert (C, \pmb w) - (\tilde{C}, \tilde{\pmb w}) \right\Vert_2
    \end{align}
    where we have used the fact that the traces of the density matrices are all between $-1$ and $1$ in the second to last inequality, and we have bounded the one norm via the two norm.
\end{proof}

We can now apply Theorem \ref{thm:general_rademacher_dudley} to the polynomial readout class:
\begin{proposition}
\label{prop:rademacher_polynomial}
    Consider the class of quantum reservoirs \\
    \begin{align}
        \mathcal{H}_{n, R_{\max}, C_{\max}}^{\text{poly}} := \Big\{ & H : (D_v)^{\mathbb{Z}_-} \to \mathbb{R} \: | \: \\
        & H(\pmb v )= h_{C, \pmb w} \left( H^{T^{\theta}} (\pmb v) \right), h_{C, \pmb w} \in \mathcal{F}_{n, R_{\max}, C_{\max}}^{\text{poly}}, T^{\theta} \in \mathcal{F}_n^{\text{QRC}} (\Theta) \Big\} \ ,
    \end{align}
    that is, the general class of reservoir maps equipped with the polynomial readout class introduced in \eqref{eq:universalReadoutClass}, with a fixed number $n \in \mathbb{N}$ of qubits, a maximal degree $R_{\max}$ of the polynomial readout, and a bound $C_{\max}$ on the maximal absolute value of the constant term in the readout. Then the Rademacher complexity is bounded as 
    \begin{align}
        \mathcal{R}_k \left( \mathcal{H}_{n, R_{\max}, C_{\max}}^{\text{poly}} \right) \leq & \frac{48 \cdot 0.89}{\sqrt{k}} \left(\sqrt{\text{dim} (\Theta)} + \sqrt{\mathcal{N}_{\text{poly}}+1} \right) \\
        & \cdot \max\left( n \sqrt{2^n} R_{\max} \mathcal{N}_{\text{poly}} L_{\Theta}' D_{\Theta} \, , \, \sqrt{\mathcal{N}_{\text{poly}} + 4 C_{\text{max}}^2} \sqrt{\mathcal{N}_{\text{poly}} + 1} \right) \ .
    \end{align}
\end{proposition}
We can see that the Rademacher complexity of the polynomial readout function scales polynomially in $n$ through the factor $\mathcal{N}_{\text{poly}}$, and exponentially in $n$ through the factor $\sqrt{2^n}$.

We can get rid of the exponential factor in Proposition \ref{prop:rademacher_polynomial} if we restrict to a finite parameter space:
\begin{proposition}
\label{cor:rademacher_polynomial}
    Consider the same class of quantum reservoirs as in Proposition \ref{prop:rademacher_polynomial}, with the additional restriction that the parameter space $\Theta$ of the CPTP maps be of finite cardinality, that is, $\vert \Theta \vert < \infty$. Then the Rademacher complexity is bounded as 
    \begin{equation}
    \label{eq:rademacher_L_h_bar_poly}
        \mathcal{R}_k \left( \mathcal{H}_{n, R_{\max}, C_{\max}}^{\text{poly}}\Biggr|_{\vert \Theta \vert < \infty} \right) \leq  \frac{1 }{\sqrt{k}} \left( \left| \Theta \right| \cdot R_{\max} \cdot \mathcal{N}_{\text{poly}} + C_{\max} \right)
    \end{equation}
    for any $k \geq 2$.    
\end{proposition}
The proof is provided in Appendix \ref{app:proof_rademacher_polynomial}.
We see that the exponential factor is replaced by the size of the parameter space, which contributes linearly to the Rademacher complexity. We may choose a very low-degree polynomial so that $\mathcal{N}_{\text{poly}}$ scales more slowly in $n$.
\begin{remark}
    Proposition \ref{cor:rademacher_polynomial} is in fact true for a much larger class of quantum reservoirs. The assumptions $\left(A_5^h\right) and \left( A_6^h \right)$ are in fact unnecessary.
\end{remark}

\subsection{Alternative Readout Functions}
\label{sec:alternative_readouts}

\subsubsection{Linear Readout}

One might hope to improve upon the bounds in Proposition \ref{prop:rademacher_polynomial} and Proposition \ref{cor:rademacher_polynomial} by considering a low-degree polynomial, or even a linear readout, similar to what was proposed by \cite{yasuda:23}, that is, a readout class defined as
\begin{align}
\label{eq:linear_readout_class}
    \mathcal{F}_{n, C_{\max}}^{\text{lin}} &= \Big\{h_{C, \pmb w}: \Bplus \cup 0 \rightarrow \mathbb{R} \: | \: h_{C, \pmb w}(\rho) = C + \sum_{i=1}^{n} w_i \text{tr}\left[ \sigma_Z^{(i)} \rho \right] \ , \ 0 \leq |C| \leq C_{\max}, w_i \in[0,1] \Big\} \ ,
\end{align} 
where we write $\pmb w = \left\{ w_i \right\}_{i=1}^n$. 
Then we find the upper bound on the Rademacher complexity:
\begin{corollary}
\label{cor:rademacher_linear}
    Consider the class of quantum reservoirs
    \begin{align}
        \mathcal{H}_{n, R_{\max}, C_{\max}}^{\text{lin}}:= & \Big\{ H : (D_v)^{\mathbb{Z}_-} \to \mathbb{R} \: | \: \\
        & \hspace{3mm} H(\pmb v )= h_{C, \pmb w} \left( H^{T^{\theta}} (\pmb v) \right), h_{C, \pmb w} \in \mathcal{F}_{n, R_{\max}, C_{\max}}^{\text{lin}}, T^{\theta} \in \mathcal{F}_n^{\text{QRC}} (\Theta) \Big\} \ ,
    \end{align}
    that is the general class of reservoir maps equipped with the linear readout class introduced in \eqref{eq:linear_readout_class}, with a fixed number $n \in \mathbb{N}$ of qubits and a bound $C_{\max}$ on the maximal absolute value of the constant term in the readout. Then the Rademacher complexity is bounded as 
    \begin{align}
        &\mathcal{R}_k \left( \mathcal{H}_{n, R_{\max}, C_{\max}}^{\text{lin}} \right) \\
        &\leq \frac{48 \cdot 0.89}{\sqrt{k}} \left(\sqrt{\text{dim} (\Theta)} + \sqrt{n+1} \right) \cdot \max\left( n \sqrt{2^n} L_{\Theta}' D_{\Theta} \, , \, \sqrt{n + 4 C_{\text{max}}^2} \sqrt{n + 1} \right) \ .
    \end{align}
\end{corollary}
The proof is provided in Appendix \ref{app:rademacher_linear}.
We see that, while the explicit polynomial dependence in $n$ has been replaced by a linear one, the exponential dependence is still there. This polynomial scaling is caused by the projective measurements and cannot be improved upon simply by replacing the polynomial readout through a linear one.

\subsubsection{Spatial Multiplexing}

Another popular readout function is called spatial multiplexing. As shown by \cite{monzani:24}, this class also introduces a polynomial algebra. For a total number $n$ of qubits, suppose we have a finite number $\ell$ of independent reservoir systems, each with $n/ \ell$ qubits (where $\ell$ is such that $n/ \ell$ is an integer). After injecting the input into each reservoir system, we consider the resulting reservoir state $\rho$ to be the joint state $ \rho = \rho_1 \otimes \cdots \otimes \rho_{\ell}$ of each individual reservoir state $\rho_1, \ldots, \rho_{\ell}$. Then the readout function of the complete state is obtained by multiplying linear readouts of the individual states, and adding a constant bias $C$, that is,

\begin{align}
\label{eq:h_sm_original}
    h_{C, \pmb w, \ell}(\rho) &= \left( \sum_{i_1 = 1}^{n/\ell} w_{i_1} \text{tr}\left[ \sigma_Z^{(i_1)} \rho \right] \right) \cdots \left( \sum_{i_{\ell} =  \left( \ell - 1 \right) \frac{n}{\ell} + 1}^{n} w_{i_{\ell}} \text{tr}\left[ \sigma_Z^{(i_{\ell})} \rho \right] \right) + C \\
    & = \prod_{j=1}^{\ell} \left( \sum_{i_j \in S_j} w_{i_j} \text{tr}\left[ \sigma_Z^{(i_j)} \rho \right] \right) + C \ ,
\end{align}
where we write $S_j = \left\{ (j-1) \frac{n}{\ell} + 1, \ldots, j \frac{n}{\ell} \right\}$. Note that we can equivalently write this function as $h_{C, \pmb W, \pmb b}(\rho) = \prod_{j=1}^{\ell_{\max}} \left( \sum_{i=1}^n W_{j,i} \text{tr}\left[ \sigma_Z^{(i)} \rho \right] + b_j \right) + C$, for some $\ell_{\max} \geq \ell$, where we write $\pmb W = \left( W_{j,i} \right)_{1 \leq j \leq \ell_{\max} \atop 1 \leq i \leq n} \in [0,1]^{\ell_{\max} \times n} $ and we have replaced the parameter $\ell$ by another parameter $\pmb b = (b_j)_{1 \leq j \leq \ell_{\max}} \in [0,1]^{\ell_{\max}}$, and we set 
\begin{itemize}
    \item $W_{j,i} = 0$ and $b_j = 1$ whenever $j > \ell$
    \item $W_{j,i} = 0$ and $b_j = 0$ whenever $j \leq \ell $ and $i \notin S_j$
    \item $W_{j,i} = w_{i_j}$ and $b_j = 0$ whenever $j \leq \ell $ and $i \in S_j$.
\end{itemize}
 We can define the readout class
\begin{equation}
\label{eq:spatial_multiplexing_readout}
\begin{aligned}
    \mathcal{F}_{n, \ell_{\max}, C_{\max}}^{\text{SM}} = & \Big\{ h_{C, \pmb w, \pmb b} : \Bplus \cup 0 \rightarrow \mathbb{R} \: | \: \\
    & \hspace{3mm} h_{C, \pmb w}(\rho) = h_{C, \pmb W, \pmb b}(\rho) = \prod_{j=1}^{\ell_{\max}} \left( \sum_{i=1}^n W_{j,i} \text{tr}\left[ \sigma_Z^{(i)} \rho \right] + b_j \right) + C \ , \\
    & \hspace{3mm} 0 \leq |C| \leq C_{\max}\, , \, \pmb W = \left( W_{j,i} \right)_{1 \leq j \leq \ell_{\max} \atop 1 \leq i \leq n} \in [0,1]^{\ell_{\max} \times n} \, , \, \\
    & \hspace{3mm} \pmb b = (b_j)_{1 \leq j \leq \ell_{\max}} \in [0,1]^{\ell_{\max}} \Big\} \ ,
\end{aligned}
\end{equation} 
where $C_{\max}$ is an upper bound on the absolute value of the constant bias. Note that this class necessarily contains all functions of the form \eqref{eq:h_sm_original} by setting the parameters $W_{j,i}$ and $b_j$ according to the rules established above. Conditions $(A^h_1)$ and $(A^h_2)$ are verified in the same way as for the readout class $\mathcal{F}_{n, R_{\max}, C_{\max}}^{\text{poly}}$ discussed in \Cref{sec:readout_map}. We also have $\left\vert h_{C, \pmb W, \pmb b}(0) \right\vert = \left\vert \prod_{j=\ell + 1}^{\ell_\max} b_j + C\right\vert \leq 1 + C_{\max}$ for all $h_{C, \pmb W, \pmb b} \in  \mathcal{F}_{n, \ell_{\max}, C_{\max}}^{\text{SM}} $ meaning that $(A^h_4)$ is verified. $\left( A_5^h \right)$ is easily verified by identifying the parameter space of $ \mathcal{F}_{n, \ell_{\max}, C_{\max}}^{\text{SM}} $, which we call $\Omega_{\text{SM}}$ as $\Omega_{\text{SM}} = [-C_{\max}, C_{\max}] \times [0,1]^{\ell_{\max}\cdot n} \times [0,1]^{\ell_\max}$, so that we have $\text{dim} \left( \Omega_{\text{SM}} \right) = 1 + \ell_{\max}(n+1)$, and the diameter is $D_{\Omega}^{\text{SM}} = \sqrt{\ell_{\max}(n+1) + 4 C_{\max}^2}$.

We verify conditions $(A^h_3)$ and $(A_6^H)$ in the following two lemmas:
\begin{lemma}
\label{proposition:lipschitz_sm_inputs}
    The spatial multiplexing readout class
    $\mathcal{F}_{n, \ell_{\max}, C_{\max}}^{\text{SM}} $ introduced in \eqref{eq:spatial_multiplexing_readout}, with a fixed number $n \in \mathbb{N}$ of qubits, a maximum polynomial degree $\ell_{\max}$ and a bound $C_{\max}$ on the maximal absolute value of the constant term in the readout verifies assumption $(A^h_3)$, that is, all maps $h_{C, \pmb w, \pmb b} \in \mathcal{F}_{n, \ell_{\max}, C_{\max}}^{\text{SM}}$ are Lipschitz-continuous w.r.t. the input with bounded Lipschitz constants $L_h \leq \Bar{L}_h^{\text{SM}}$, where we write $\Bar{L}_h^{\text{SM}} := n(n+1)^{\ell_{\max} - 1} \sqrt{2^n} \ell_{\max}$.
\end{lemma}
The proof is provided in Appendix \ref{app:lipschitz_sm}.

\begin{lemma}
\label{proposition:lipschitz_sm_omega}
    The spatial multiplexing readout class
    $\mathcal{F}_{n, \ell_{\max}, C_{\max}}^{\text{SM}} $ introduced in \eqref{eq:spatial_multiplexing_readout}, with a fixed number $n \in \mathbb{N}$ of qubits, a maximum polynomial degree $\ell_{\max}$ and a bound $C_{\max}$ on the maximal absolute value of the constant term in the readout verifies assumption $(A^h_6)$, that is. all maps $h_{C, \pmb w, \pmb b} \in \mathcal{F}_{n, \ell_{\max}, C_{\max}}^{\text{SM}}$ are Lipschitz-continuous w.r.t. the parameters $\omega \in \Omega_{\text{SM}}$, with bounded Lipschitz constants $L_{\Omega}$, where we write
    \begin{equation}
    \label{eq:L_Omega_SM}
        L_{\Omega}^{\text{SM}} := \sqrt{(n+1)^{2 \ell_{\max} - 1} \ell_{\max} + 1} \ .
    \end{equation}
\end{lemma}
The proof is provided in Appendix \ref{app:lipschitz_sm_parameters}.

We can bound the Rademacher complexity using Theorem \ref{cor:rademacher_polynomial}:
\begin{corollary}
\label{lemma:SM_Lipschitz}
    Consider the class of quantum reservoirs \\
    $\mathcal{H}_{n, \ell_{\max}, C_{\max}}^{\text{SM}}:= \left\{ H : (D_v)^{\mathbb{Z}_-} \to \mathbb{R} \: | \: H(\pmb v )= h \left( H^{T^{\theta}} (\pmb v) \right), h \in \mathcal{F}_{n, \ell_{\max}, C_{\max}}^{\text{SM}}, T^{\theta} \in \mathcal{F}_n^{\text{QRC}} (\Theta) \right\} $, i.e. the general class of reservoir maps equipped with the spatial multiplexing readout class as defined in \eqref{eq:spatial_multiplexing_readout}. Then the Rademacher complexity is bounded as 
    \begin{align}
    \label{eq:rademacher_L_h_bar_sm}
        \mathcal{R}_k(\mathcal{H}_{n, \ell_{\max}, C_{\max}}^{\text{SM}}) \leq & \frac{48 \cdot 0.89}{\sqrt{k}} \left( \sqrt{\text{dim}(\Theta)} + \sqrt{1 + \ell_{\max}(n+1)} \right) \\
        & \cdot \max \Biggl( n(n+1)^{\ell_{\max}-1} \sqrt{2^n} L_{\Theta}' D_{\Theta} \, , \,\sqrt{(\ell_{\max}(n+1) + 4 C_{\max}^2) ((n+1)^{2 \ell_{\max} - 1} \ell_{\max} + 1)} \Biggr)
    \end{align}
    for any $k \geq 2$. 
\end{corollary}

We can interpret $\ell_{\max}$ to be analogous to $R_{\max}$ in the class $ \mathcal{F}_{n, R_{\max}, C_{\max}}^{\text{poly}}$, as $\ell_{\max}$ limits the degree of the polynomial in the measurements of the reservoir state. 

In general, the reservoir systems of the individual subsystems are governed by different dynamics, so that the remaining results of the paper are not directly applicable to the general spatial multiplexing case. In particular, the tensor product of Lipschitz-continuous maps is not necessarily Lipschitz-continuous, which means that an ensemble of reservoir classes that individually verify the hypotheses $A^T$ does not necessarily verify the hypotheses $A^T$ when taken as a tensor product. A more thorough analysis is required to establish general risk bounds when using spatial multiplexing. For the remainder of the document we thus consider the polynomial readout class $\mathcal{F}_{n, R_{\max}, C_{\max}}^{\text{poly}}$.

From the results in this section, we see that the readout class that scales best in terms of the numbers of qubits is unsurprisingly the linear readout, however, it still induces an exponential scaling. Using a finite parameter space as in Proposition \ref{cor:rademacher_polynomial}, we may reduce this exponential scaling to a polynomial one, though the size of the parameter space might be large. As we will see in Section \ref{sec:generalisation_bounds}, the choice of the CPTP map $T^{\theta}$ also greatly influences the scaling of the risk bounds in $n$.

\section{Two Specialised Subclasses of Quantum Reservoirs}
\label{sec:PTR_RRR}

In closed quantum systems, time evolution is described by unitary transformations. 
In contrast, open quantum systems, that is to say systems interacting with an environment, are 
described by density matrices (see Section~\ref{sec:framework}) whose evolutions are 
governed by CPTP maps. As introduced earlier, quantum reservoir dynamics are modelled using such CPTP maps. These maps capture both coherent dynamics and irreversible effects such as decoherence and dissipation. For a more detailled introduction to open quantum systems, see for example \cite{breuer:02}.


We now introduce two more specific classes of quantum reservoirs, namely the \textit{Partial Trace Reservoir (PTR)} and the \textit{Random Reinitialisation Reservoir (RRR)}, which are adaptations of the reservoir classes by \cite{chen:19} and \cite{chen:20}, respectively. We use the properties of these reservoirs to establish risk bounds in \Cref{sec:generalisation_bounds} that depend on their parameters as well as the parameters of the polynomial readout class $\mathcal{F}_{n, R_{\text{max}, C_{\text{max}}} }^{\text{poly}}$ and the number of training samples.

\subsection{Partial Trace Reservoir}
\label{sec:PTR}

Before introducing the PTR class of quantum reservoirs, we recall that in closed quantum systems, time evolution is described by unitary transformations. Such unitary operations on density matrices can be expressed in terms of a Hamiltonian operator $H$ representing the energy of the system. In this case, the evolution of a state $\rho_t$ over a time $\tau > 0$ is given by $\rho_{t+1} = e^{-iH\tau} \rho_t e^{iH\tau}$. In contrast, quantum reservoirs are generally considered to be open systems and their evolution is therefore described by CPTP maps rather than purely unitary dynamics.

We now define the PTR subclass of the reservoir map from \cite{chen:19}, which slightly modifies the input injection and readout function from the first proposal of a QRC introduced by \cite{fujii:16}. Modified versions of this class are frequently considered in quantum reservoir computing (see e.g. \cite{yasuda:23}, \cite{sannia:24}).

Conceptually, we consider a quantum system composed of $n$ \textit{reservoir qubits}, carrying the reservoir state, together with an additional \textit{ancillary qubit} used to inject inputs. We write $\ket{0}$ and $\ket{1}$ for the computational basis states of a single qubit. 
The state space of $n$ qubits is the Hilbert space $(\mathbb{C}^2)^{\otimes n}$ 
with basis $\{\ket{0}, \ket{1}\}^{\otimes n}$. 
Given an input $v$, the ancilla qubit is prepared in the mixed state $v \ket{0}\bra{0} + (1-v) \ket{1}\bra{1}$. A joint unitary evolution is then applied to the combined reservoir and ancillary system for a time $\tau$. Subsequently, the ancillary qubit is discarded via a partial trace. The partial trace maps an operator acting on a joint Hilbert space to an operator acting on a subsystem. Concretely, let $\mathcal{H}_A$ and $\mathcal{H}_B$ be two Hilbert spaces, and consider an operator $O = M_A \otimes N_B$ that acts on $\mathcal{H} = \mathcal{H}_A \otimes \mathcal{H}_B$. Then, the partial trace $\text{tr}_B$ w.r.t. subsystem $B$ is defined as
\begin{align}
    \text{tr}_B : \mathcal{H} & \to \mathcal{H}_A \\
    O & \mapsto \textbf{tr}_B[M_A \otimes N_B] = M_A \; \text{tr}[N_B] \ ,
\end{align}
where $\text{tr}[N_B]$ is the standard scalar trace of the operator $N_B$. Using the partial trace on the ancillary qubit allows us to consider only the reservoir subsystem. This partial trace induces a non-unitary CPTP evolution on the reservoir state. For a more in-depth discussion on the partial trace, see for example \cite{lidar}.

\begin{figure}[!h]
  \centering
  \includegraphics[trim={0 5cm 0 5cm}, clip, width=1\linewidth]{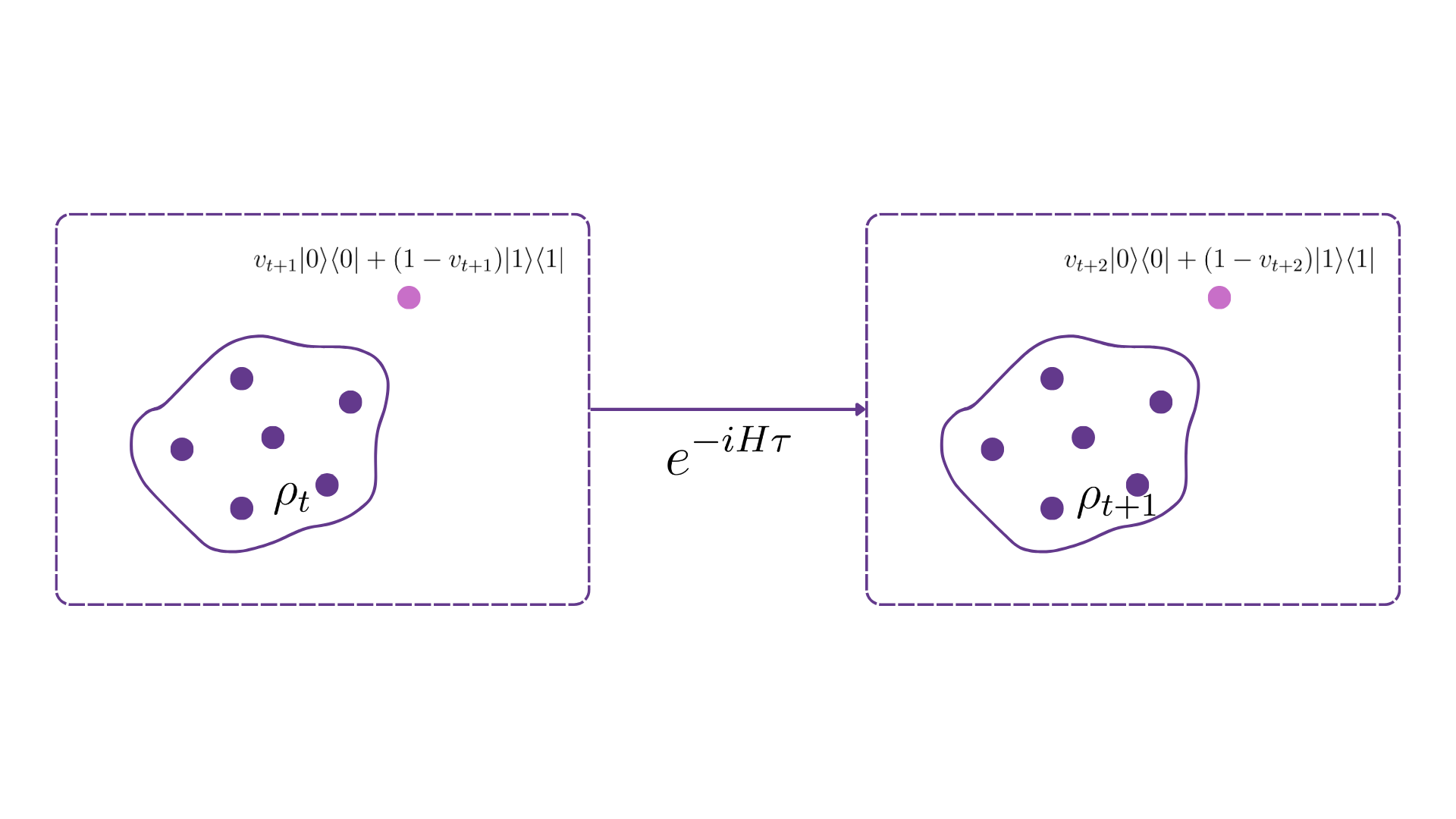}
  \caption{Schematic of the CPTP map of a partial trace reservoir. The purple dots designate the reservoir qubits, the solid outline determines the reservoir. The single pink dot designates the ancillary ``input qubit'', and the dashed line designates the system on which we apply the unitary evolution determined by the XY-Hamiltonian. At time $t$, the reservoir is in state $\rho_t$. A new input $v_{t+1}$ is injected in the ancillary qubit by setting it to the mixed state $v_{t+1} \ket{0}\bra{0} + (1-v_{t+1}) \ket{1}\bra{1}$ and the reservoir qubits along with the ancillary qubit are made to evolve according to the unitary map induced by the Hamiltonian. After time $\tau$, the new reservoir state $\rho_{t+1}$ is obtained by tracing out the ancilla qubit. Finally, the next input is injected into the ancilla qubit.}
  \label{fig:ptr}
\end{figure}

More concretely, define the input space \\ 
$D_v^{\text{PTR}} = \left[ \frac{1}{2} \left(1-\sqrt{\sqrt{2}-1} \right) + \epsilon_{\text{PTR}}, \frac{1}{2} \left( 1+\sqrt{\sqrt{2}-1} \right) - \epsilon_{\text{PTR}} \right]$ for some fixed \\ $0 < \epsilon_{\text{PTR}} < \sqrt{ \sqrt{2} - 1}$. We consider an $n$-qubit system and define the CPTP map 

\begin{align}
    T^{J, \gamma, \tau} : D_v^{\text{PTR}} \times \mathcal{B} &\to \mathcal{B} \\
    (v, A) &\mapsto \text{tr}_0\left[ e^{-i H (J, \gamma) \tau} A \otimes \left(v \ket{0}\bra{0} + (1-v) \ket{1}\bra{1} \right) e^{i H(J, \gamma) \tau} \right]
\end{align}
where $\text{tr}_0$ denotes the partial trace with respect to the ancillary qubit, and $\tau \in \mathbb{R}_+$ is a time parameter. We choose for the Hamiltonian $H (J, \gamma)$ the $XY$-Hamiltonian of the form
\begin{align}
\label{eq:XY_hamiltonian}
    H (J, \gamma) = \sum_{i=0}^n \sum_{j=i+1}^n J^{i,j} \left( \sigma_X^{(i)} \sigma_X^{(j)} + \sigma_Y^{(i)} \sigma_Y^{(j)} \right) + \gamma \sum_{i=0}^n \sigma_Z^{(i)}
\end{align}
where $\sigma_X^{(i)}, \sigma_Y^{(i)}$ and $ \sigma_Z^{(i)} $ are Pauli-$X$, -$Y$ and -$Z$ matrices applied to the $i$-th qubit (i.e. $\sigma_X^{(i)} = I \otimes \cdots \otimes I \otimes \sigma_X \otimes I \otimes \cdots \otimes I $ and so on), and the parameters $J = \left\{ J^{ij} \right\}$ and $\gamma$ are real-valued constants. For the remainder of the document, we set $\tau = 1$ for readability and to simplify calculations.

Define $\Theta_{\text{PTR}} := [J_{\min}, J_{\max}]^{n(n-1)} \times [\gamma_{\min}, \gamma_{\max}] \in \mathbb{R}^{n(n-1)+1}$ as the space of bounded reservoir parameters $J$ and $\gamma$. For a fixed number $n$ of qubits, we can then define the class of PTR quantum reservoir maps $\mathcal{F}_n^{\text{PTR}} \left( \Theta_{\text{PTR}} \right) := \left \{ T^{J, \gamma} \: | \:  (J, \gamma) \in \Theta_{\text{PTR}} \right \}$.

In the following Lemma, we show that the class of reservoir maps $\mathcal{F}_n^{\text{PTR}}\left( \Theta_{\text{PTR}} \right)$ verifies Assumptions $A^T$. We use this Lemma in the next section to prove a generalisation bound for the PTR class of quantum reservoirs.
\begin{lemma}
\label{lemma:T_PTR_contractive_Lipschitz}
    For any fixed parameters $(J, \gamma) \in \Theta_{\text{PTR}}$, the CPTP map \\ $T^{J, \gamma, \tau} : D_v^{\text{PTR}} \times \mathcal{B} \to \mathcal{B} \, , \, 
    T^{J, \gamma} (v, A) \mapsto \text{tr}_0\left[ e^{-iH(J, \gamma)} A \otimes \left(v \ket{0}\bra{0} + (1-v) \ket{1}\bra{1} \right) e^{iH(J, \gamma)} \right]$ with Hamiltonian $H(J, \gamma)$ is strictly contractive in the space of density operators whenever $0 < \epsilon_{\text{PTR}} < \sqrt{ \sqrt{2} - 1}$, with contractivity constant \\ $ r_{\text{PTR}} (\epsilon_{\text{PTR}}) = 2 \sqrt{2} \left( \epsilon_{\text{PTR}}^2 - \epsilon_{\text{PTR}} \sqrt{ \sqrt{2} -1 } \right) + 1 $. Additionally, the map is Lipschitz-continuous in the space of inputs $D_v^{\text{PTR}}$ with Lipschitz-constant $L_R^{\text{PTR}} = 2$. Finally, for any fixed input $v \in D_v^{\text{PTR}}$ and any fixed density matrix $\rho \in \Bplus$, the map is Lipschitz-continuous in the parameter space $\Theta_{\text{PTR}}$ with Lipschitz-constant $ L_{\Theta}^{\text{PTR}} = 2^{n+1}\sqrt{ n (4n - 3)}$.
\end{lemma}
The proof is provided in Appendix \ref{app:T_PTR_contractive_Lipschitz}. Note that we can somewhat control the strength of the contractivity constant by choosing a convenient value for $\epsilon_{\text{PTR}}$.

It is also clear that the assumption $\left(A^{\Theta}\right)$ is verified, with $\text{dim}\left( \Theta_{\text{PTR}} \right) = n(n-1) + 1$, and $D_{\Theta}^{\text{PTR}} = \sqrt{n(n-1) \left( J_{\max} - J_{\min} \right)^2 + \left( \gamma_{\max} - \gamma_{\min} \right)^2}$.

We then define the class of reservoir functionals associated with the class of PTR reservoir maps as
\begin{equation}
\label{eq:ptr_reservoirclass}
   \mathcal{H}_{n, R_{\text{max}}, C_{\text{max}}}^{\text{PTR}}\! :=\! \left\{ H\! :\! (D_v)^{\mathbb{Z}_-}\! \to\! \mathbb{R} \: | \: H( v )= h \left( H^{T^{J, \gamma}} (v) \right), h \in \mathcal{F}_{n, R_{\text{max}}, C_{\text{max}}}^{\text{poly}}, T^{J, \gamma} \in \mathcal{F}_n^{\text{PTR}}\left( \Theta_{\text{PTR}} \right) \right\} \ .
\end{equation}

\subsection{Random Reinitialisation Reservoir}
\label{sec:RRR}

\begin{figure}[!h]
  \centering
  \includegraphics[trim={0 1cm 0 1cm}, clip, width=1\linewidth]{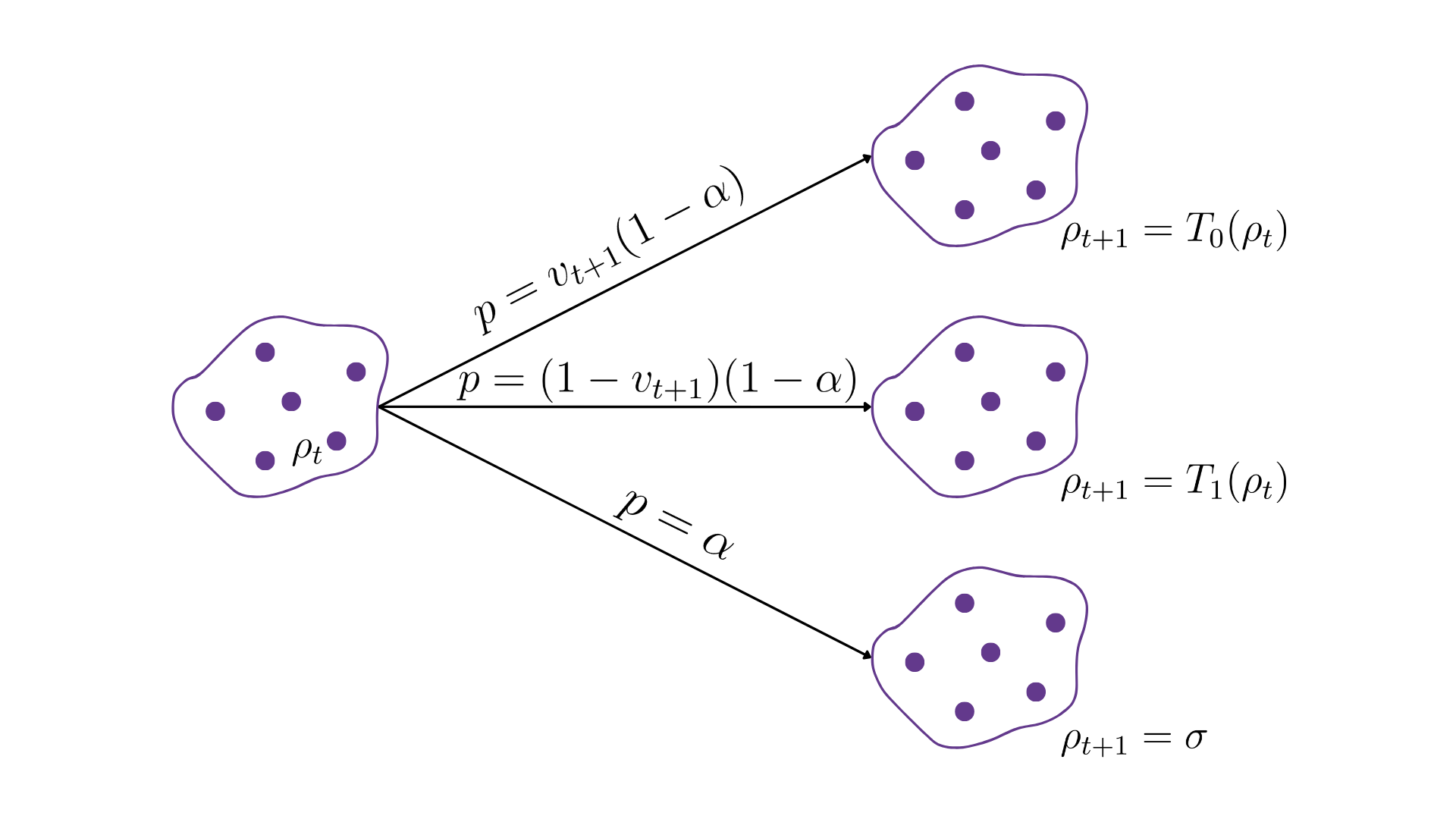}
  \caption{Schematic of the CPTP map of a Random Reintialisation Reservoir. The purple dots designate the reservoir qubits, the solid outline determines the reservoir. At time $t$, the reservoir is in state $\rho_t$. With probability $v_{t+1} (1 - \alpha) $ we apply the map $T_0$ to the reservoir to obtain the new reservoir state $\rho_{t+1} = T_0(\rho_t)$, with probability $(1 - v_{t+1}) (1 - \alpha) $ we apply the map $T_1$ to the reservoir to obtain the new reservoir state $\rho_{t+1} = T_1(\rho_t)$, and with probability $\alpha$ we reintialise the reservoir to some fixed quantum state $\sigma$.}
  \label{fig:rrr}
\end{figure}

We now define the RRR subclass of the reservoir map from \cite{chen:20}, which provides an interesting, more general alternative to the PTR class for quantum platforms on which the Hamiltonian \eqref{eq:XY_hamiltonian} or the input injection in a mixed state are not available. The RRR map can be interpreted as a probabilistic injection of the input $v$, where for some parameter $\alpha \in (0,1)$, with probability $(1 - \alpha)v$ we apply the CPTP map $T_0$ to the reservoir, and with probability $(1 - \alpha)(1 - v)$ we apply the CPTP map $T_1$ (see Fig.~\ref{fig:rrr} for a visual representation).

Mathematically, define the input space $D_v^{\text{RRR}} = \left[ \epsilon_{\text{RRR}}, \frac{1}{2} - \epsilon_{\text{RRR}} \right]$ for some fixed $0 < \epsilon_{\text{RRR}} < \frac{1}{4}$ . We consider an $n$-qubit system and define the CPTP map

\begin{align}
    T^{\alpha, \sigma} : D_v^{\text{RRR}} \times \mathcal{B} &\to \mathcal{B} \\
    (v, A) &\mapsto ( 1 - \alpha ) \left( v T_0 + (1-v) T_1 \right)(A) + \alpha \sigma 
\end{align}

where $\sigma$ is a fixed arbitrary quantum state, $\alpha$ is the parameter that determines the probability of reinitialising the reservoir state to $\sigma$ and $T_0 \neq T_1$ are fixed CPTP maps, contractive in the space of density matrices, with contraction constants $r_0$ and $r_1$ such that $r_0 + r_1 < 1$. If $r_0 = r_1$, we write $r_ 0 = r_1 = r_{0,1}$. If $r_0 \neq r_1$, we impose $r_0 > r_1$.

For a fixed number $n$ of qubits, we can then define the class of RRR quantum reservoir maps

\begin{equation}
\label{eq:class_rrr_maps}
\begin{aligned}
    \mathcal{F}_n^{\text{RRR}} (\Theta_{\text{RRR}}) := \{ T^{\alpha, \sigma} \: | \:  (\alpha, \sigma) \in \Theta_{\text{RRR}}  \} 
\end{aligned}
\end{equation}
where $\Theta_{\text{RRR}} := [\alpha_{\min}, \alpha_{\max}] \times \Sigma $ is some compact subset of $ (0, 1) \times \bar{\mathcal{B}}^+ $. For generality we will set $\Sigma = \Bplus$ for the remainder of the document.

In the following Lemma we show that the class of reservoir maps $\mathcal{F}_n^{\text{RRR}}(\Theta_{\text{RRR}})$ verifies Assumptions $A^T$. We use this Lemma in the next section to prove a generalisation bound for the RRR class of quantum reservoirs.

\begin{lemma}
\label{lemma:RRR_AT} 
    The class of RRR quantum reservoir maps $\mathcal{F}_n^{\text{RRR}}$ defined in \eqref{eq:class_rrr_maps} verifies assumptions $A^T$. More specifically, any CPTP map $T^{\alpha, \sigma} \in \mathcal{F}_n^{\text{RRR}}$ with arbitrary $n$ is strictly contractive in the space of density operators. If $r_0 = r_1 = r_{0,1}$, the contractivity constant is given by $r_{\text{RRR}} = (1 - \alpha_{\min}) r_{0,1} $. If $r_0 > r_1$, the contractivity constant is given by $r_{\text{RRR}} = 1 - \epsilon_{\text{RRR}}$. Furthermore, any CPTP map $T^{\alpha, \sigma} \in \mathcal{F}_n^{\text{RRR}}$ with arbitrary $n$ is Lipschitz-continuous in the space of inputs $D_v^{\text{RRR}}$ with Lipschitz-constant $L_R^{\text{RRR}} = 2 (1 - \alpha_{\min})$. Finally, for any fixed input $v \in D_v^{\text{RRR}}$ and any fixed density matrix $\rho \in \Bplus$, the map is Lipschitz-continuous in the parameter space $\Theta_{\text{RRR}}$ with Lipschitz-constant $ L_{\Theta}^{\text{RRR}} = 2$.
\end{lemma}
The proof is provided in Appendix \ref{app:T_RRR_contractive_Lipschitz}.

We can also verify that the assumption $\left(A^{\Theta} \right)$ is verified: Note that we can write any density matrix in the basis of the Bloch sphere
\begin{equation}
    \rho = \frac{I}{2^n} + \sum_{j_1, \ldots, j_n = \left\{0, 1, 2, 3, 4\right\} \atop j_1\cdots j_n \neq 0} \alpha_{j_1 j_2 \ldots j_n} \bigotimes_{i=1}^n \sigma_{j_i}^{(i)} \,
\end{equation}
where $\sigma_{j_i}^{(i)}$ represents the identity operator on the $i$th qubit if $j_i = 0$, the Pauli-X, -Y or -Z operator if $j_i = 1, 2$ or $3$ respectively (see for example \cite{chen:19} or \cite{kimura:03} for more details), the coefficients $\left\{ \alpha_{j_1 j_2 \ldots j_n} \right\}$ are real-valued, and we have $\left\Vert \alpha \right\Vert_2 \leq \sqrt{1 - \frac{1}{2^n}}$, where we have written $\alpha = \left\{ \alpha_{j_1 j_2 \ldots j_n} \right\}$. We can thus equivalently characterise the parameter space of the RRR reservoir map as $\Theta_{\text{RRR}} = [\alpha_{\min}, \alpha_{\max}] \times \mathcal{B}_{4^n - 1} \left(\sqrt{1 - \frac{1}{2^n}}\right)$, where we write $ \mathcal{B}_{4^n - 1} \left(\sqrt{1 - \frac{1}{2^n}}\right)$ for the $4^n - 1$-dimensional euclidean ball with radius $\sqrt{1 - \frac{1}{2^n}}$. We thus immediately get $\text{dim}\left( \Theta_{\text{RRR}} \right) = 4^n $, and $D_{\Theta}^{\text{RRR}} = \sqrt{(\alpha_{\max} - \alpha_{\min})^2 + 2}$, where we have used the fact that for any two density matrices $\sigma_1, \sigma_2$, we have
\begin{equation}
    \left\Vert \sigma_1 - \sigma_2 \right\Vert_2 \leq \sqrt{\text{tr} [\sigma_1]^2 + \text{tr}[\sigma_2]^2 } \leq \sqrt{2} \ .
\end{equation}

We then define the class of reservoir functionals associated with the class of RRR reservoir maps as
\begin{align}
\label{eq:rrr_reservoirclass}
    \mathcal{H}_{n, R_{\max}, C_{\max}}^{\text{RRR}}\! :=\! & \Big\{ H\! :\! \left( D_v^{\text{RRR}} \right)^{\mathbb{Z}_-}\! \to\! \mathbb{R} \: \Big| \: \\
    & \hspace{3mm} H(\pmb v )= h \left( H^{T^{\alpha, \sigma}} (\pmb v) \right) ,  h \in \mathcal{F}_{n, R_{\max}, C_{\max}}^{\text{poly}} ,  T^{\alpha, \sigma} \in \mathcal{F}_n^{\text{RRR}} \left(\Theta_{\text{RRR}} \right) \Big\} \ .
\end{align}

Using the results of this section as well as the previous sections, we can derive an upper bound on the generalisation error of the RRR and PTR quantum reservoir classes.

\section{Generalisation Bounds of the Quantum Reservoir Classes}
\label{sec:generalisation_bounds}

We are now ready to present the main results. We can apply \cite[Theorem 14]{gonon:20} to the PTR and RRR reservoirs, with the appropriate constants. To highlight the dependence in the reservoir parameters, and to simplify notation, we omit the dependencies on parameters that are not determined by the reservoir map or the readout map. The explicit expressions can be found in Appendix \ref{app:explicit_risk_bounds}.

\begin{theorem}
\label{cor:risk_bound_ptr_O}

 Let $n \in \mathbb{N}$ be a fixed number of qubits. Consider the PTR class of reservoir functionals $\mathcal{H}_{n, R_{\text{max}}, C_{\text{max}}}^{\text{PTR}}$ defined in (\ref{eq:ptr_reservoirclass}) as well as the input and target processes defined in \Cref{sec:framework}, fulfilling the hypotheses $A^{IO}$. Define $\zeta_{\text{max}} := \text{max} ( r_{\text{PTR}} (\epsilon_{\text{PTR}}), D_{w^y}, D_{w^v}) $. Then for all $m \in \mathbb{N}$ such that $\log(m) < m \log(\zeta_{\text{max}}^{-1})$ and for all $\delta \in (0, 1)$, with probability at least $1 - \delta$ we have
    \begin{align}
    \label{eq:T_PTR_risk_bound_big_O}
         & \ \sup_{H \in \mathcal{H}_{n, R_{\max}, C_{\max}}^{\text{PTR}}} \left| R(H) - \hat{R}_m(H) \right|  \\
         & \leq \mathcal{O} \left( \Bar{L}_h^{\text{poly}}\text{max} \left\{ \frac{P_1^p}{m} \ , \ P_2^p \frac{\log{m}}{m} \ , \ P_3^p \sqrt{\frac{\log{m}}{m}} \ , \ P_4^p \sqrt{\frac{\log{4/\delta}}{2m}}\right\} \right)
    \end{align}

    where the reservoir parameter dependent expressions are given by

    \begin{align}
        & P_1^p = \text{max} \left\{\frac{ r_{\text{PTR}} (\epsilon_{\text{PTR}})}{ 1 -  r_{\text{PTR}} (\epsilon_{\text{PTR}})} \, , \,  2 L_{\ell} \zeta_{\text{max}}^{-1} \, , \, L_{\ell} C_y \zeta_{\text{max}}^{-1} \right\} \\
        & P_2^p = \text{max} \left\{ 1 \, , \,  C_{\text{max}}  \, , \, 2 \zeta_{\text{max}}^{-1} C_v \right\} \\ \label{eq:constants_risk_bound_PTR}
        & P_3^p = \frac{1}{ \sqrt{\log{\zeta_{\text{max}}^{-1}}}} \max \left\{ n^2 \, , \, \sqrt{\mathcal{N}_{\text{poly}}}\right\} \\
        & \hspace{1.5cm} \cdot \max \left\{ \frac{n^2 2^{n} \left( J_{\max} - J_{\min} \right) }{r_{\text{PTR} \left( \epsilon_{\text{PTR}} \right)}} \, , \, \frac{n 2^{n} \left( \gamma_{\max} - \gamma_{\min} \right) }{r_{\text{PTR} \left( \epsilon_{\text{PTR}} \right)}}   \, , \, \frac{ \mathcal{N}_{\text{poly}} }{\Bar{L}_h^{\text{poly}}} \, , \, \frac{ C_{\max} \sqrt{\mathcal{N}_{\text{poly}}} }{\Bar{L}_h^{\text{poly}}} \right\} \\
        & P_4^p = \frac{ 1 }{ 1 -r_{\text{PTR}} (\epsilon_{\text{PTR}}) } \text{max} \left\{ r_{\text{PTR}} (\epsilon_{\text{PTR}})  \, , \, 2 M_{\xi} L_v \left\lVert w^v \right\rVert_1   \right\} 
    \end{align}

    and $\Bar{L}_h^{\text{poly}}$ is as in \eqref{eq:L_h_Bar_poly}, i.e. $\Bar{L}_h^{\text{poly}} =  R_{\text{max}} \cdot n \sqrt{2^n} \left( \binom{n+R_{\text{max}}}{R_{\text{max}}} - 1 \right)$. 
    
\end{theorem}

The proof can be found in Appendix \ref{app:risk_bound_ptr_O}.

From Theorem \ref{cor:risk_bound_ptr_O} we see that the main scaling issues of the generalisation error of the PTR class with respect to the number of qubits comes from the Lipschitz constant and the dimension and diameter of the parameter space of the polynomial readout, although $P_3^p$ also contains exponential and polynomial factors related to the Lipschitz constants and the parameter space of the reservoir map.

We can proceed analogously for the RRR reservoir class, distinguishing the case where $r_0 = r_1$ from the case where $r_0 > r_1$ to obtain the following result:

\begin{theorem}
\label{cor:risk_bound_rrr_O}
    Let $n \in \mathbb{N}$ be a fixed number of qubits. Consider the RRR class of reservoir functionals $\mathcal{H}_{n, R_{\max}, C_{\max}}^{\text{RRR}}$ defined in (\ref{eq:rrr_reservoirclass}) as well as the input and target processes defined in \Cref{sec:framework}, fulfilling the hypotheses $A^{IO}$. Define $\zeta_{\max} := \max (r_{\text{RRR}}, D_{w^y}, D_{w^v}) $. Then for all $m \in \mathbb{N}$ such that $\log(m) < m \log(\zeta_{\max}^{-1})$ and for all $\delta \in (0, 1)$, with probability at least $1 - \delta$ we have

    \begin{align}
    \label{eq:T_RRR_risk_bound_big_O}
         & \sup_{H \in \mathcal{H}_{n, R_{\max}, C_{\max}}^{\text{RRR}}} \left| R(H) - \hat{R}_m(H) \right|  \\
         & \leq \mathcal{O} \left( \Bar{L}_h^{\text{poly}} \max \left\{ \frac{P_1^r}{m} , \ P_2^r \frac{\log{m}}{m} \ , \ P_3^r \sqrt{\frac{\log{m}}{m}} \ , \ P_4^r \sqrt{\frac{\log{4/\delta}}{2m}}\right\} \right)
    \end{align}

    where the reservoir parameter dependent expressions are given by

    \begin{align}
    \label{eq:constants_risk_bound_RRR_cst}
        & P_1^r = 
        \begin{cases}
            \max \left\{ \frac{(1 - \alpha_{\min}) r_{0,1} }{1 - (1 - \alpha_{\min}) r_{0,1}} \ , \ 1 \right\}   & \text{ if } r_0 = r_1 = r_{0,1} \\
            \max \left\{ \frac{1 - \epsilon_{\text{RRR}}}{\epsilon_{\text{RRR}}} \ , \ 1 \right\} & \text{ if } r_0 > r_1 \\
        \end{cases} \\
        & P_2^r = \max \left\{ 1 \ , \ 2 (1 - \alpha_{\min})  \zeta_{\max}^{-1} L_{\ell} C_v \ , \ C_{\max} L_{\ell} \right\}  \\
        & P_3^r =  \frac{ \max \left\{ 2^n \, , \, \sqrt{\mathcal{N}_{\text{poly}}} \right\}}{ \sqrt{\log{\zeta_{\text{max}}^{-1}}}} 
        \begin{cases}
            \max \left\{ \frac{\alpha_{\max} - \alpha_{\min}}{(1 - \alpha_{\min})r_{0,1}} 
            \, , \, \frac{\mathcal{N}_{\text{poly}}}{\Bar{L}_h^{\text{poly}}} \, , \, \frac{C_{\max} \sqrt{\mathcal{N}_{\text{poly}}}}{\Bar{L}_h^{\text{poly}}} \right\} & \text{ if } r_0 = r_1 = r_{0,1} \\
            \max \left\{ \frac{\alpha_{\max} - \alpha_{\min}}{1 - \epsilon_{\text{RRR}}}  \, , \, \frac{\mathcal{N}_{\text{poly}}}{\Bar{L}_h^{\text{poly}}} \, , \, \frac{C_{\max} \sqrt{\mathcal{N}_{\text{poly}}}}{\Bar{L}_h^{\text{poly}}} \right\} & \text{ if } r_0 > r_1 
        \end{cases}\\
        & P_4^r = 
        \begin{cases}
             \frac{\left( 1 - \alpha_{\min} \right)}{ 1 - \left( \alpha_{\min} \right) r_{0,1} }  \max \left\{ r_{0,1} \ , \ 2 M_{\xi} L_v \left\lVert w^v \right\rVert_1 \right\} & \text{ if } r_0 = r_1 = r_{0,1} \\
            \frac{1}{\epsilon_{\text{RRR}}} \max \left\{ 1 - \epsilon_{\text{RRR}} \ , \ 2 \left( 1 - \alpha_{\min} \right) M_{\xi} L_v \left\lVert w^v \right\rVert_1 \right\} & \text{ if } r_0 > r_1 \\
        \end{cases} 
    \end{align}
    and $\Bar{L}_h^{\text{poly}}$ is as in \eqref{eq:L_h_Bar_poly}, i.e. $\Bar{L}_h^{\text{poly}} =  R_{\max} \cdot n \sqrt{2^n} \left( \binom{n+R_{\max}}{R_{\max}} - 1 \right)$.

\end{theorem}
The proof can be found in Appendix \ref{app:risk_bound_rrr_O}.

Comparing Theorems \ref{cor:risk_bound_ptr_O} and \ref{cor:risk_bound_rrr_O}, we can see how the choice of the reservoir map as well as the diameter and the dimension of the parameter space influence the risk bounds. Note that one may restrict the space $\Sigma$ of admissible density matrices in the RRR class in order to reduce the scaling in $P_3^r$.

\color{black}

\begin{remark}    
    Note that we have analysed the generalisation error in an idealised setting, without taking into account the estimation of the observables. In reality, multiple shots are necessary to average over the measurement results. The risk bound presented in this paper can help prevent overfitting in the idealised setting of stationary time series without noise. 
\end{remark}

\section{Discussion}
\label{sec:discussion}
In this paper, we have established bounds on the Rademacher complexity for both a general class of quantum reservoirs as well as for a general class of reservoir maps equipped with different specific readout classes. Using these results, we were able to derive risk bounds for the more specific quantum reservoir classes which we have introduced in \Cref{sec:PTR_RRR}, based on the classes introduced by \cite{chen:19} and \cite{chen:20}. 

If one wants to use the aforementioned classes for a forecasting task, our risk bounds can give an idea on how to choose a convenient set of parameters that are more likely to generalise well. 

On the one hand, we find risk bounds that scale as $\mathcal{O}(\sqrt{\log{m} / m})$ in the number of training samples $m$, which suggests a convergence rate for the generalisation error. It is important to note that tzhis scaling comes from the risk bounds established by \cite{gonon:20} in a classical setting. We were able to verify that the quantum reservoir classes analysed here fall into this setting.

On the other hand, our analysis shows that both the choice of the class of readout maps as well as the reservoir map, and in particular the parameter space, influence the way that the risk bound scales with the number of qubits. As shown in \Cref{sec:generalisation_bounds}, the choice of a polynomial readout class leads to the risk bounds scaling as $\mathcal{O}(n 2^{3n/2} \binom{n+R_{\max}}{R_{\max}} )$ in the fixed number of qubits $n$, where $R_{\max}$ is an upper bound on the degree of the polynomial. As discussed in \Cref{sec:alternative_readouts}, this problem can be somewhat alleviated by choosing either a linear readout function, which turns the polynomial factor into a linear one, or a polynomial readout with a finite sized parameter space, which replaces the exponential scaling in the qubits with a linear scaling in the size of the parameter space.

The risk bounds we provide are not directly comparable to the risk bounds established for universal classical reservoir classes, since the modified quantum reservoir classes studied here are a priori no longer universal. 

The bad scaling of the risk bound in the number of qubits motivates the search for universal reservoir classes that do not require a polynomial readout class. \cite{grigoryeva:18a} have shown that the reservoir class called State Affine System (SAS) is universal even with a linear readout. \cite{pena:23} showed that all input-dependent quantum reservoir maps can be written in a similar form, and thus universality results from \cite{grigoryeva:18a} can be applied to reservoir systems with linear readouts when the necessary conditions are verified. However, verifying these conditions is not trivial; apart from the difficulty of finding explicit matrix representations of the CPTP maps, which is necessary to apply the results from \cite{pena:23}, establishing risk bounds on those representations is not an easy undertaking, as these matrices necessarily depend on the input variables, making it difficult to separate inputs and reservoir states in the decomposition of the reservoir functional. Additionally, the risk bounds for SAS were only established for linear readout functions. In this paper we have extended the analysis to the popular polynomial readout class and establish risk bounds for the specific reservoir classes PTR and RRR with reservoir-dependent parameters.

Another potential avenue is to utilise the rather general results in \cite{sannia:24} which establishes conditions on the Lindbladian for the reservoir class to be universal when using spatial multiplexing. In forthcoming work, we develop device-specific classes that can harness the natural dynamics of specific quantum systems, including experimental verification of the theoretical results.

\acks{This work was supported by the European Union’s Horizon 2020 research and innovation programs under grant agreement No. 817482 (PASQuanS) and No. 101079862 (PASQuanS2) as well as the ANR projects Q-COAST (ANR-19-CE48-0003) and IGNITION (ANR-21-CE47-0015). Part of this research was conducted during a visit to the Institute for Mathematical and Statistical Innovation (IMSI), which is supported by the National Science Foundation (Grant No. DMS-1929348). 
The authors would like to thank Paulin Jacquot, Nadia Oudjane and Gérard Biau for useful discussions and technical advice.}


\bibliography{sample}

\appendix

\section{Proof of Lemma \ref{lemma:chen19}}
\label{app:lemma_convergence}

    We will show that the contractivity condition $(A^T_2)$ necessarily implies the condition on $T$ in \cite[Theorem 3]{chen:19}, namely that for all inputs $v \in D_v$, there exists $\varepsilon \in (0, 1]$ such that the input-dependent CPTP map $T(v, \cdot)$ restricted to the hyperplane $\mathcal{B}_0$ of $2^n \times 2^n$ traceless Hermitian operators satisfies $\normiii[Big]{ T \bigr|_{\mathcal{B}_0}}_2 \leq 1 - \varepsilon$.

    Let $v \in D_v$ be some fixed input. First, note that for any hermitian operator $A$, we can write its spectral decomposition $A = \sum_k \lambda_k \ket{u_k}\bra{u_k}$, where the $\lambda_k \in \mathbb{R}$ are the eigenvalues of $A$, and the $\ket{u_k}$ are the corresponding eigenvectors. We can rewrite this as
    \begin{equation}
        A = \sum_{\lambda_k > 0} \lambda_k \ket{u_k}\bra{u_k} + \sum_{\lambda_k<0} \lambda_k \ket{u_k}\bra{u_k} \ .
    \end{equation}
    Now define $A_+ := \sum_{\lambda_k > 0} \lambda_k \ket{u_k}\bra{u_k}$ and $A_- := -\sum_{\lambda_k<0} \lambda_k \ket{u_k}\bra{u_k}$. Then, both $A_+$ and $A_-$ are positive semidefinite and we can write $A = A_+ - A_-$. In the case when $A \in \mathcal{B}_0$, we have $\text{tr} [A] = 0$, that is, $\text{tr}[A_+] = \text{tr}[A_-] = c$, where $c$ is some real number. Define the matrices $\rho_+ := A_+/c$ and $\rho_- := A_-/c$. They are both hermitian as well as positive semi-definite and of unit trace, that is, they are both density matrices. Then, using Assumption $\left( A_2^T \right)$, we can write for any $A \in \mathcal{B}_0$ such that $\left\Vert A \right\Vert_2 = 1$, 
    \begin{align}
       \left\Vert T(v, A) \right\Vert_2 & = \left\Vert T(v, A_+) - T(v, A_-)\right\Vert_2 = \vert c \vert \left\Vert T \left(v, \rho_+\right) - T\left(v,\rho_-\right) \right\Vert_2 \\
       & \leq r \vert c \vert \left\Vert \rho_+ - \rho_- \right\Vert_2 = r \vert c \vert \left\Vert \frac{A_+ - A_-}{c} \right\Vert_2 = r \left\Vert A \right\Vert_2 = r \ ,
    \end{align}
    where we used the fact that $T$ is a linear map in its second argument.
    
    The final result follows by applying \cite[Theorem 3]{chen:19} for $\varepsilon = 1-r$ and its following remark, which states that any quantum system that satisfies the aforementioned condition on $T$, applied an infinite number of times, maps any initial state $\rho_{-\infty}$ to the state $\overrightarrow{ \prod}_{i=0}^{\infty} T (v_{-i}) \left( \frac{I}{2^n} \right)$.

\section{Proof of Theorem \ref{thm:general_rademacher_dudley}}
\label{app;thm_general_rademacher_dudley}
For ease of notation, we write $\rho_0^{\theta, j} := \overrightarrow{ \prod}_{i=0}^{\infty} T^{\theta} \left( \Tilde{V}^{(j)}_{-i} \right) \rho_{-\infty} $.

 First, note that for any fixed Rademacher sequence $\left( \varepsilon_j \right)_j$ and any input sequence $\left( \rho_0^{\theta, j} \right)_j$ we have
    \begin{align}
        \sup_{\theta \in \Theta, \omega \in \Omega} \left\vert \sum_{j=0}^{k-1} \varepsilon_j h_{\omega} \left( \rho_0^{\theta, j} \right) \right\vert &= \sup_{\theta \in \Theta, \omega \in \Omega} \max \left( \sum_{j=0}^{k-1} \varepsilon_j h_{\omega} \left( \rho_0^{\theta, j} \right) \, , \, - \sum_{j=0}^{k-1} \varepsilon_j h_{\omega} \left( \rho_0^{\theta, j} \right)\right) \\
        & \leq \sup_{\theta \in \Theta, \omega \in \Omega} \sum_{j=0}^{k-1} \varepsilon_j h_{\omega} \left( \rho_0^{\theta, j} \right) + \sup_{\theta \in \Theta, \omega \in \Omega} \sum_{j=0}^{k-1} \left( -\varepsilon_j\right) h_{\omega} \left( \rho_0^{\theta, j} \right) \ \text{a.s.} \ ,
    \end{align}
    where we used the fact that $0 \in \mathcal{F}_n^{O}$, that is, the supremum is always non-negative. Thus, we can write
    \begin{align}
        \mathbb{E} \left[ \sup_{\theta \in \Theta, \omega \in \Omega} \left\vert \sum_{j=0}^{k-1} \varepsilon_j h_{\omega} \left( \rho_0^{\theta, j}  \right) \right\vert \right] \leq \mathbb{E} \left[ \sup_{\theta \in \Theta, \omega \in \Omega} \sum_{j=0}^{k-1} \varepsilon_j h_{\omega} \left( \rho_0^{\theta, j} \right) \right] + \mathbb{E} \left[ \sup_{\theta \in \Theta, \omega \in \Omega} \sum_{j=0}^{k-1} \left( -\varepsilon_j\right) h_{\omega} \left( \rho_0^{\theta, j} \right) \right] \ .
    \end{align}

    Next, note that, because the Rademacher variables are i.i.d. and $\varepsilon_j \in \{-1, 1\}$ with $\mathbb{P}\left( \varepsilon_j = -1 \right) = \mathbb{P}\left( \varepsilon_j = 1 \right) = 1/2$ for all $j$, we can write, by conditioning,
    \begin{align}
        \mathbb{E} \left[ \sup_{\theta \in \Theta, \omega \in \Omega} \sum_{j=0}^{k-1} \left( -\varepsilon_j\right) h_{\omega} \left( \rho_0^{\theta, j} \right) \right] & = \mathbb{E}_{\pmb V} \left[ \mathbb{E}_{\varepsilon} \left[ \sup_{\theta \in \Theta, \omega \in \Omega} \sum_{j=0}^{k-1} \left( -\varepsilon_j\right) h_{\omega} \left( \rho_0^{\theta, j} \right) \Bigg\vert \pmb V \right] \right] \\
        & = \mathbb{E}_{\pmb V} \left[ \mathbb{E}_{\varepsilon} \left[ \sup_{\theta \in \Theta, \omega \in \Omega} \sum_{j=0}^{k-1} \varepsilon_j h_{\omega} \left( \rho_0^{\theta, j} \right) \Bigg\vert \pmb V \right] \right] \\  
        & = \mathbb{E} \left[ \sup_{\theta \in \Theta, \omega \in \Omega} \sum_{j=0}^{k-1} \varepsilon_j h_{\omega} \left( \rho_0^{\theta, j} \right) \right] \ .
    \end{align}
    (See for example the proof of \cite[Theorem 3.1]{mohri:12} for a similar argument). Thus, we have
    \begin{align}
        \mathbb{E} \left[ \sup_{\theta \in \Theta, \omega \in \Omega} \left\vert \sum_{j=0}^{k-1} \varepsilon_j h_{\omega} \left( \rho_0^{\theta, j}  \right) \right\vert \right] \leq 2\mathbb{E} \left[ \sup_{\theta \in \Theta, \omega \in \Omega} \sum_{j=0}^{k-1} \varepsilon_j h_{\omega} \left( \rho_0^{\theta, j} \right) \right] \ ,
    \end{align}
    meaning that it suffices to bound 
    $ \mathbb{E} \left[ \sup_{\theta \in \Theta, \omega \in \Omega} \sum_{j=0}^{k-1} \varepsilon_j h_{\omega} \left( \rho_0^{\theta, j} \right) \right] \ . $

    We will do this in three steps: 
    \begin{enumerate}
        \item We show that the process defined by
        \begin{align}
        \label{eq:definition_process}
            Z_{\theta, \omega}^k := \sum_{j=0}^{k-1} \varepsilon_j h_{\omega} \left( \rho_0^{\theta, j}  \right)
        \end{align}
        is sub-gaussian.
        \item We bound the covering number of the parameter space $\Theta \times \Omega$.
        \item We find an upper bound on $\mathbb{E} \left[ \sup_{\theta \in \Theta, \omega \in \Omega} \sum_{j=0}^{k-1} \varepsilon_j h_{\omega} \left( \rho_0^{\theta, j} \right) \right]$ using Dudley's entropy integral. 
    \end{enumerate}

    We begin by showing that the process $Z_{\theta, \omega}^k$ is sub-gaussian.

    \begin{lemma}
    \label{lemma:subgaussian}
        Let $h_{\omega}$ be a readout function from the general readout class $\mathcal{F}_n^{O}$, and let $T^{\theta}$ be a CPTP map from the general class of quantum reservoir maps $\mathcal{F}_n^{\text{QRC}}(\Theta)$, parametrised in the general parameter space $\Theta$, all as defined in Section \ref{sec:general_class}. Then, the process $Z_{\theta, \omega}^k$ as defined in \eqref{eq:definition_process} is sub-Gaussian for the metric $d((\theta, \omega) \, , \, (\theta', \omega')) = \sqrt{k} \left( \Bar{L}_h L'_{\Theta} \Vert \theta - \theta' \Vert_2 + L_{\Omega} \Vert \omega - \omega' \Vert_2 \right) $.
    \end{lemma}
    \begin{proof}
        Note that we have, for any fixed $\theta, \theta' \in \Theta$ and for any $\omega, \omega' \in \Omega$,
    \begin{align}
        \left( h_{\omega} \left( \rho_0^{\theta, j} \right) - h_{\omega'}\left( \rho_0^{\theta', j}  \right) \right)^2 & = \left\vert h_{\omega} \left( \rho_0^{\theta, j} \right) - h_{\omega'} \left( \rho_0^{\theta', j}  \right) \right\vert^2 \\
        & \leq \left( \left\vert h_{\omega} \left( \rho_0^{\theta, j} \right) - h_{\omega}\left( \rho_0^{\theta', j}  \right) \right\vert + \left\vert h_{\omega} \left( \rho_0^{\theta', j} \right) - h_{\omega'} \left( \rho_0^{\theta', j}  \right) \right\vert \right)^2 \\
        & \leq \left( \bar{L}_h\left\Vert \rho_0^{\theta, j} - \rho_0^{\theta', j} \right\Vert_2 + L_{\Omega} \left\Vert \omega - \omega' \right\Vert_2 \right)^2 \\
        & \leq \left( \bar{L}_h L_{\Theta}' \left\Vert \theta - \theta' \right\Vert_2 + L_{\Omega} \left\Vert \omega - \omega' \right\Vert_2 \right)^2 \ ,
    \end{align}
    where we have used the Lipschitz-continuity of the readout function in both the density matrices as well as the parameters $\omega$, and the Lipschitz continuity of the CPTP map in the parameters $\theta$.

    Additionally, for any fixed $\theta_0 \in \Theta$ and any fixed $\omega_0 \in \Omega $, we have
    \begin{align}
    \label{eq:theta_0_omega_0_null}
        \mathbb{E} \left[ Z_{\theta_0, \omega_0}^k \right] &= \mathbb{E} \left[ \sum_{j=0}^{k-1} \varepsilon_j h_{\omega_0} \left( \rho_0^{\theta, j}  \right) \right] = \sum_{j=0}^{k-1} \mathbb{E} \left[ \varepsilon_j h_{\omega_0} \left( \rho_0^{\theta, j}  \right) \right] = \sum_{j=0}^{k-1} \mathbb{E} \left[ \varepsilon_j \right] \mathbb{E}\left[ h_{\omega_0} \left( \rho_0^{\theta, j}  \right) \right] = 0 \ ,
    \end{align}
    where we have used independence between the $\varepsilon_j$ and the $V^{(j)}$ in the third equality and the fact that the $\varepsilon_j$ are centred in the last.

    Thus we have, for any $\omega, \omega' \in \Omega$ and for any $\theta, \theta' \in \Theta$
    \begin{align}
        \mathbb{E} \left[ \exp{\left\{\lambda \left( Z_{\theta, \omega}^k - Z_{\theta', \omega'}^k \right)\right\}} \right] & = \mathbb{E}\left[  \mathbb{E}\left[ \exp{\left\{\lambda \sum_{j=0}^{k=1} \varepsilon_j \left( h_{\omega} \left( \rho_0^{\theta, j} \right) - h_{\omega'} \left( \rho_0^{\theta', j} \right) \right) \right\}} \Bigg\vert \pmb V \right]  \right] \\
        & = \mathbb{E}\left[  \prod_{j=0}^{k-1} \mathbb{E}\left[ \exp{\left\{\lambda  \varepsilon_j  \left( h_{\omega} \left( \rho_0^{\theta, j} \right) - h_{\omega'} \left( \rho_0^{\theta', j} \right) \right) \right\}} \Bigg\vert \pmb V \right]  \right] \\
        & \leq \mathbb{E}\left[  \prod_{j=0}^{k-1} \exp{ \left\{ \frac{4\lambda^2}{8} \left( h_{\omega} \left( \rho_0^{\theta, j} \right) - h_{\omega'} \left( \rho_0^{\theta', j} \right) \right) ^2 \right\}}  \right] \\
        & \leq \mathbb{E}\left[  \prod_{j=0}^{k-1} \exp{ \left\{ \frac{\lambda^2}{2} \left( \bar{L}_h L_{\Theta} \left\Vert \theta - \theta' \right\Vert_2 + L'_{\Omega} \left\Vert \omega - \omega' \right\Vert_2 \right)^2 \right\}}  \right] \\
        & = \exp\left\{ \frac{\lambda^2}{2} k \left( \bar{L}_h L'_{\Theta} \left\Vert \theta - \theta' \right\Vert_2 + L_{\Omega} \left\Vert \omega - \omega' \right\Vert_2 \right)^2 \right\} \ ,
    \end{align}
    where we used the independence between the Rademacher variables and $\Tilde{\pmb V}^{(j)}$ in the second equality, Hoeffding's Lemma in the first inequality (note that $\varepsilon \in \{-1, 1\}$ and the quantity $h_{\omega} \left( \rho_0^{\theta, j} \right) - h_{\omega'} \left( \rho_0^{\theta', j} \right)$ is a constant under the conditional expectation), the second inequality is an application of \eqref{eq:theta_0_omega_0_null}, and the final equality follows from the fact that there is no dependence on any random variables in the second to last line left. Thus, taking the logarithm, the process $Z_{\theta, \omega}$ is sub-Gaussian for the metric 
  \begin{equation}
   \label{eq:metric_subgaussian}
    d((\theta, \omega) \, , \, (\theta', \omega')) = \sqrt{k} \left( \Bar{L}_h L'_{\Theta} \Vert \theta - \theta' \Vert_2 + L_{\Omega} \Vert \omega - \omega' \Vert_2 \right) \ .
  \end{equation}
  
 \end{proof}

   We recall the definitions of $\kappa$-nets and covering numbers.

    Let $T$ be a metric or pseudo-metric space (that is, equipped with a distance $d$ that does not necessarily separate points).

    \begin{definition}(for example \cite[Definition 4.2.1]{vershyn:18})
        For $\kappa > 0$, a subset $\mathcal{N} \subset T$ is called a \textit{$\kappa$-net of $T$} if every point in $T$ is within distance $\kappa$ of some point of $\mathcal{N}$, i.e.
        \begin{equation}
            \forall x \in T \ \exists x_0 \in \mathcal{N} \ : \ d(x, x_0) \leq \kappa \ .
        \end{equation}
    \end{definition}
    
    \begin{definition}\cite[Definition 4.2.2]{vershyn:18}
        The smallest cardinality of a $\kappa$-net of $T$ is called the \textit{covering number} of $T$ ans is denoted $\mathcal{N}(T, d; \kappa)$. Equivalently, $\mathcal{N}(T, d; \kappa)$ is the smallest number of closed balls with centers in $T$ and radii $\kappa$ whose union covers $T$.
    \end{definition}

    We can now bound the covering number of the set $\Theta \times \Omega$ w.r.t. the metric $d$ as defined in \eqref{eq:metric_subgaussian}:
    \begin{lemma}
    \label{lemma:covering_numbers}
    Let $\Theta$ be a parameter space that fulfils the condition $\left(A^{\Theta} \right)$, and let $\Omega$ be a parameter space that fulfils the condition $\left( A_5^h \right)$. Furthermore, let $d$ be the metric on $\Theta \times \Omega$ defined by 
    \begin{equation}
        d((\theta, \omega) \, , \, (\theta', \omega')) = \sqrt{k} \left( \Bar{L}_h L'_{\Theta} \Vert \theta - \theta' \Vert_2 + L_{\Omega} \Vert \omega - \omega' \Vert_2 \right) \ ,
    \end{equation}
    for some $k \geq 2$. Then the covering number of $\Theta \times \Omega$ w.r.t. $d$ is bounded as
    \begin{equation}
        \mathcal{N} \left( \Theta \times \Omega, d; \kappa \right) \leq \left(1 + \frac{2D_{\Theta} \bar{L}_h L'_{\Theta}\sqrt{k}}{\kappa} \right)^{\text{dim}(\Theta) }  \left(1 + \frac{2D_{\Omega} L_{\Omega}\sqrt{k}}{\kappa} \right)^{\text{dim}(\Omega) }\ .
    \end{equation}
    \end{lemma}

    \begin{proof}
        First, let us write $d((\theta, \omega) \, , \, (\theta', \omega')) = d_{\Theta}(\theta, \theta') + d_{\Omega}(\omega, \omega') $, with $d_{\Theta}(\theta, \theta') = \sqrt{k} \Bar{L}_h L'_{\Theta} \Vert \theta - \theta' \Vert_2 $ and $d_{\Omega}(\omega, \omega') = \sqrt{k} L_{\Omega} \Vert \omega - \omega' \Vert_2 $. Suppose that $\mathcal{N}_{\Theta}$ is a $\kappa/2$-covering of $\Theta$ for the metric $d_{\Theta}$, and suppose that $\mathcal{N}_{\Omega}$ is a $\kappa/2$-covering of $\Omega$ for the metric $d_{\Omega}$. Then, for any tuple $(\theta, \omega) \in \Theta \times \mathcal{B}_{\Vert \cdot \Vert_2}$, we can find $\theta^* \in \mathcal{N}_{\Theta}$ such that $d_{\Theta}(\theta, \theta^*) \leq \kappa/2$, and $\omega^* \in \mathcal{N}_{\Omega}$ such that $d_{\Omega}(\omega, \omega^*) \leq \kappa/2$ and thus, $d((\theta, \omega), (\theta^*, \omega^*)) = d_{\Theta}(\theta, \theta^*) + d_{\Omega}(\omega, \omega^*) \leq \kappa/2 = \kappa$. Then note that, if $\mathcal{N} \left( \Theta, d_{\Theta}; \kappa/2 \right)$ is the $\kappa/2$-covering number of $\Theta$ for the metric $d_{\Theta}$, and if $\mathcal{N} \left( \mathcal{B}_{\Vert \cdot \Vert_2}, d_{\Omega}; \kappa/2 \right)$ is the $\kappa/2$-covering number of $\Omega$ for the metric $d_{\Theta}$, then the product of the two coverings will cover the Cartesian product of $\Theta$ and $\Omega$. That is, we can write 
    \begin{equation}
        \mathcal{N} \left( \Theta \times \Omega, d; \kappa \right) \leq \mathcal{N} \left( \Theta, d_{\Theta}; \kappa/2 \right) \cdot \mathcal{N} \left( \Omega, d_{\Omega}; \kappa/2 \right) \ .
    \end{equation}

  
    The covering number of a closed euclidean ball $\mathcal{B}^{\text{dim}_B}_{\Vert \cdot \Vert_2}(R)$ of radius $R$ and dimension $\text{dim}_B$ is well known to be 
    $ \mathcal{N} \left( \mathcal{B}^{\text{dim}_B}_{\Vert \cdot \Vert_2}(R), \Vert \cdot \Vert_2; \kappa \right) \leq \left(1 + \frac{2R}{\kappa} \right)^{\text{dim}_B } $
    (see for example \cite[Corollary 4.2.11]{vershyn:18}). To bound the covering number of $\Theta$, we can use Jung's theorem to conclude that there exists a closed euclidean ball with radius $R \leq D_{\Theta} \sqrt{\frac{\text{dim}(\Theta)}{2 (\text{dim}(\Theta) +1)}} < D_{\Theta}$  so that we can use the above inequality to find
    \begin{align}
        \mathcal{N} \left( \Theta, \Vert \cdot \Vert_2; \frac{\kappa}{\bar{L}_h L'_{\Theta}\sqrt{k}} \right) & \leq \left(1 + \frac{2D_{\Theta} \bar{L}_h L'_{\Theta}\sqrt{k}}{\kappa} \right)^{\text{dim}(\Theta) } \ .
    \end{align}
    Similarly, we find $ \mathcal{N} \left( \Omega, \Vert \cdot \Vert_2; \frac{\kappa}{L_{\Omega}\sqrt{k}} \right) \leq \left(1 + \frac{2D_{\Omega} L_{\Omega}\sqrt{k}}{\kappa} \right)^{\text{dim}(\Omega) }$.
    
    Putting everything together, we find
    \begin{equation}
        \mathcal{N} \left( \Theta \times \Omega, d; \kappa \right) \leq \left(1 + \frac{2D_{\Theta} \bar{L}_h L'_{\Theta}\sqrt{k}}{\kappa} \right)^{\text{dim}(\Theta) }  \left(1 + \frac{2D_{\Omega} L_{\Omega}\sqrt{k}}{\kappa} \right)^{\text{dim}(\Omega) }\ .
    \end{equation}
    \end{proof}

    We will use Dudley's entropy integral to bound the expectation of the supremum of $Z_{\theta, \omega}^k$. We restate the theorem here:
    \begin{theorem}\cite[Corollary 13.2]{boucheron:13}
    \label{thm:boucheron}
        Let $T$ be a finite pseudo-metric space and let $\left( X_t \right)_{t \in T}$ be a sub-Gaussian process, that is
        \begin{equation}
            \log \mathbb{E} \left[ e^{\lambda \left( X_t - X_{t'} \right)}\right] \leq \frac{\lambda^2 d^2(t, t')}{2}
        \end{equation}
        for all $t, t' \in T$ and all $\lambda > 0$. Then for any $t_0 \in T$,
        \begin{equation}
            \mathbb{E} \left[ \sup_{t \in T} X_t - X_{t_0} \right] \leq 12 \int_0^{ \sup_{t \in T} d(t, t_0)/2} \sqrt{\log \mathcal{N} \left( T, d; \kappa \right)} \  d\kappa\ .
        \end{equation} 
    \end{theorem}
    Note that, while the theorem is stated for finite parameter sets, it is easily extended to a separable set (see for example \cite[beginning of chapter 11]{ledoux:91} for a discussion). In particular, under the assumptions $\left(A^{\Theta} \right)$ and $\left( A_5^h \right)$, the set $\Theta \times \Omega$ is separable and of finite covering number for any $\kappa > 0$ so that Theorem \ref{thm:boucheron} is directly applicable to $\Theta \times \Omega$. 

    We are now ready to prove Theorem \ref{thm:general_rademacher_dudley}. First, note that we have, for any fixed $(\theta_0, \omega_0) \in \Theta \times \Omega$,
    \begin{align}
    \label{eq:bound_integral_bound}
        \sup_{(\theta, \omega)} \frac{1}{2} \sqrt{k} \left( \Bar{L}_h L'_{\Theta} \Vert \theta - \theta_0 \Vert_2 + L_{\Omega} \Vert \omega - \omega_0 \Vert_2 \right)& \leq \frac{\sqrt{k}}{2}  \left( \Bar{L}_h L'_{\Theta} D_{\Theta} + L_{\Omega} D_\Omega \right)  \ .
    \end{align}

    Using Theorem \ref{thm:boucheron}, we can now write
    \begin{align}
    \label{eq:dudley_applied}
        &\mathbb{E} \left[ \sup_{\theta \in \Theta, \omega \in \Omega} Z_{\theta, \omega}^k - Z_{\theta_0, \omega_0}^k \right] \\
        & \leq 12 \int_0^{ \frac{\sqrt{k}}{2}  \left( \Bar{L}_h L'_{\Theta} D_{\Theta} + L_{\Omega} D_\Omega \right)} \sqrt{\log{ \left( \left(1 + \frac{2D_{\Theta} \bar{L}_h L'_{\Theta}\sqrt{k}}{\kappa} \right)^{\text{dim}(\Theta) }  \left(1 + \frac{2D_{\Omega} L_{\Omega}\sqrt{k}}{\kappa} \right)^{\text{dim}(\Omega) }\right)}} \  d\kappa \\ 
        & = 12 \int_0^{ \frac{\sqrt{k}}{2}  \left( \Bar{L}_h L'_{\Theta} D_{\Theta} + L_{\Omega} D_\Omega \right)} \sqrt{{\text{dim}(\Theta) } \log{ \left(1 + \frac{2D_{\Theta} \bar{L}_h L'_{\Theta}\sqrt{k}}{\kappa} \right)+  {\text{dim}(\Omega) } \log{ \left(1 + \frac{2D_{\Omega} L_{\Omega}\sqrt{k}}{\kappa} \right)}}} \  d\kappa \ .
    \end{align} 

    To simplify calculations, we can distinguish two cases:

    If $D_{\Theta} \Bar{L}_h L'_{\Theta} \geq D_{\Omega} L_{\Omega}$, we have $\frac{\sqrt{k}}{2}  \left( \Bar{L}_h L'_{\Theta} D_{\Theta} + L_{\Omega} \Omega \right) \leq \sqrt{k} \bar{L}_h L'_{\Theta} D_{\Theta}$ and $\log{ \left(1 + \frac{2D_{\Omega} L_{\Omega}\sqrt{k}}{\kappa} \right)} \leq \log{ \left(1 + \frac{2D_{\Theta} \bar{L}_h L'_{\Theta}\sqrt{k}}{\kappa} \right)} $ and thus
    \begin{align}
        &\mathbb{E} \left[ \sup_{\theta \in \Theta, \omega \in \Omega} Z_{\theta, \omega}^k - Z_{\theta_0, \omega_0}^k \right] \\
        & \leq 12 \int_0^{ \sqrt{k} \bar{L}_h L'_{\Theta} D_{\Theta}} \sqrt{{\text{dim}(\Theta) } \log{ \left(1 + \frac{2D_{\Theta} \bar{L}_h L'_{\Theta}\sqrt{k}}{\kappa} \right)+  {\text{dim}(\Omega) }  \log{ \left(1 + \frac{2D_{\Theta} \bar{L}_h L'_{\Theta}\sqrt{k}}{\kappa} \right)}}} \  d\kappa \\ 
        & \leq 24 \sqrt{k} \bar{L}_h L'_{\Theta} D_{\Theta} \left( \sqrt{\text{dim}(\Theta)} \int_0^{1/2} \sqrt{\log\left( 1 + \frac{1}{u} \right)} du + \sqrt{\text{dim}(\Omega)} \int_0^{1/2} \sqrt{\log\left( 1 + \frac{1}{u} \right)} du \right) \\
        & \leq 24 \sqrt{k} \bar{L}_h L'_{\Theta} D_{\Theta} \left( \sqrt{\text{dim}(\Theta)} + \sqrt{\text{dim}(\Omega)} \right) \cdot 0.89 \ ,
    \end{align}
    
    where we used a change of variables $u = \frac{\kappa}{2 D_{\Theta} \Bar{L}_h L'_{\Theta} \sqrt{k}}$. 

    If $D_{\Theta} \Bar{L}_h L'_{\Theta} \leq D_{\Omega} L_{\Omega}$, we have $\frac{\sqrt{k}}{2}  \left( \Bar{L}_h L'_{\Theta} D_{\Theta} + L_{\Omega} \Omega \right) \leq \sqrt{k} L_{\Omega} D_{\Omega}$ and $\log{ \left(1 + \frac{2D_{\Omega} L_{\Omega}\sqrt{k}}{\kappa} \right)} \geq \log{ \left(1 + \frac{2D_{\Theta} \bar{L}_h L'_{\Theta}\sqrt{k}}{\kappa} \right)} $ and thus, using similar arguments as above, we have
    \begin{align}
        \mathbb{E} \left[ \sup_{\theta \in \Theta, \omega \in \Omega} Z_{\theta, \omega}^k - Z_{\theta_0, \omega_0}^k \right] \leq 24 \sqrt{k} D_{\Omega} L_{\Omega} \left( \sqrt{\text{dim}(\Theta)} + \sqrt{\text{dim}(\Omega)} \right) \cdot 0.89 \ .
    \end{align}
    
    Thus we have
    \begin{align}
        \mathbb{E} \left[ \sup_{\theta \in \Theta, \omega \in \Omega} Z_{\theta, \omega}^k - Z_{\theta_0, \omega_0}^k \right] \leq 24 \cdot 0.89 \sqrt{k} \left( \sqrt{\text{dim} (\Theta)} + \sqrt{\text{dim} (\Omega)} \right) \cdot \max\left( \Bar{L}_h L_{\Theta}' D_{\Theta} \, , \, D_{\Omega} L_{\Omega} \right) \ .
    \end{align}
    Now, because for any fixed $\theta_0 \in \Theta$ and any fixed $\omega_0 \in \Omega $, we have $\mathbb{E}\left[ Z_{\theta_0}^k \right] = 0$ according to \eqref{eq:theta_0_omega_0_null}, we can omit $\mathbb{E}\left[ Z_{\theta_0}^k \right]$ in the l.h.s. of \eqref{eq:dudley_applied}, and we find:
    \begin{align}
        \mathbb{E} \left[ \sup_{\theta \in \Theta, \omega \in \Omega} \sum_{j=0}^{k-1} \varepsilon_j h_{\omega} \left( \rho_0^{\theta, j}  \right) \right] \leq 24 \cdot 0.89 \sqrt{k} \left( \sqrt{\text{dim} (\Theta)} + \sqrt{\text{dim} (\Omega)} \right) \cdot \max\left( \Bar{L}_h L_{\Theta}' D_{\Theta} \, , \, D_{\Omega} L_{\Omega} \right) \ .
    \end{align}
    The final result follows immediately by multiplying this bound by two.
\color{black}
=

\section{Proof of lemma \ref{lemma:lipschitz_poly}}
\label{app:lemma_lipschitz_poly}

In the following we show that all readout functions in the class $\mathcal{F}_{n, R_{\max}, C_{\max}}^{\text{poly}}$ are Lipschitz-continuous in the space of square $2^n \times 2^n$ matrices $\rho$ with $\text{tr} \left[ \sigma_Z^{(i)} \rho \right] \in [-1,1] $ for all $i$, with Lipschitz-constants bounded by $\Bar{L}_h^{\text{poly}}$ as defined in \eqref{eq:L_h_Bar_poly}.

Let $\rho_1, \rho_2 \in \Bplus \cup 0$ be two density matrices or the square $2^n \times 2^n$ matrix with only zero entries. Denote by $g_i(\rho):=\text{tr}\left[ \sigma_Z^{(i)} \rho \right] \in[-1,1]$  the expectation of operator $\sigma_Z^{(i)}$ (i.e. of the measurement operator of the $i$-th qubit on the $Z$-axis). The function $g_i$ is a Lipschitz-continuous function with Lipschitz-constant $\sqrt{2^n}$ for all $i$, since
    \begin{align}
        \left| \text{tr}(\sigma_Z^{(i)} \rho_1) - \text{tr}(\sigma_Z^{(i)} \rho_2) \right| & = \left| \text{tr}(\sigma_Z^{(i)} ( \rho_1 - \rho_2 )) \right| \\ 
        & \leq \sqrt{\text{tr}(\sigma_Z^{(i)^{\dag}} \sigma_Z^{(i)})} \sqrt{\text{ tr } ( (\rho_1 - \rho_2)^{\dag} (\rho_1 - \rho_2) )} \\
        & = \sqrt{2^n} \lVert \rho_1 - \rho_2 \rVert_2 \label{eq:cs_lipsch}
    \end{align}
    where we have used linearity of the trace operator and matrix multiplication in the first step, the Cauchy-Schwarz inequality in the second step, and the fact that $\sigma_Z^{(i)}$ is hermitian and unitary in the last step.

    Next, define $f(x_{i_1}, \ldots x_{i_n})=w_{i_1,\ldots,i_n}^{r_{i_1},\ldots,r_{i_n}} (x_{i_1})^{r_{i_1}} \cdots (x_{i_n})^{r_{i_n}}$, where $ x_{i_k} \in [-1, 1]$ for all $i_k$.  
    Then for each $x_{i_k}$, the partial derivative of $f(x_{i_1}, \ldots x_{i_n})$ with respect to $x_{i_k}$ is
    \begin{equation}
        D_{x_{i_k}}f(x_{i_1},\ldots x_{i_n}) = w_{i_1,\ldots,i_n}^{r_{i_1},\ldots,r_{i_n}} r_{i_k} (x_{i_1})^{r_{i_1}}\cdots (x_{i_k})^{r_{i_k-1}} \cdots (x_{i_n})^{r_{i_n}} \ .
    \end{equation}
    
    It follows that for each $x_{i_k}$, $\rVert D_{x_{i_k}}f(x_{i_1},\ldots x_{i_n})\lVert_2 \leq \mathop{\max_{\substack{i_1,\ldots,i_n \\ r_{i_1},\ldots,r_{i_n}}}} |w_{i_1,\ldots,i_n}^{r_{i_1},\ldots,r_{i_n}}| \cdot R_{\max} \leq R_{\max}$. Recall that an everywhere differentiable multivariate function is Lipschitz continuous if it has bounded partial derivatives, in which case the Lipschitz constant is given by the square root of the sum of the square of the upper bounds of the partial derivatives. That means that the Lipschitz constant $L_f$ of $f$ is given by $L_f = \sqrt{S_{i_1}^2+\ldots + S_{i_n}^2}$, where $S_{i_k}$ is the upper bound of the $i$th partial derivative, i.e. $R$ in our case. This can be shown by applying the mean-valued theorem in multiple variables as well as the Cauchy-Schwarz inequality, see for example \cite[Theorem 54.2 and following discussion]{AppliedMaths}. Thus, the function $f$ is Lipschitz-continuous with Lipschitz constant $L_f = \sqrt{n} \cdot R $. Then we can write

    \begin{align}
    \label{eq:ineq_lipschitz_poly}
        \lVert & f( g_1(\rho_1), \ldots, g_n(\rho_1) ) - f( g_1(\rho_2), \ldots, g_n(\rho_2) ) \rVert_2 \\
        & \leq L_f \left\lVert ( g_1(\rho_1)-  g_1(\rho_2), \ldots, g_n(\rho_1) - g_n(\rho_2) ) \right\rVert_2 \\
        & = L_f \sqrt{ ( g_1(\rho_1) - g_1(\rho_2) )^2 + \dots +  ( g_n(\rho_1) - g_n(\rho_2) )^2} \\
        & \leq L_f \sqrt{n \cdot 2^n \lVert \rho_1 - \rho_2 \rVert_2^2} =  n \cdot R_{\max} \cdot \sqrt{2^n} \lVert \rho_1 - \rho_2 \rVert_2 \ .
    \end{align}
    Note that the sum of Lipschitz-continuous functions is Lipschitz-continuous, where the Lipschitz constant is the sum of the Lipschitz constants of the individual summands. In our case, all summands have the same Lipschitz-constant $L_{f,g} = R_{\max} \cdot n \sqrt{2^n} $ so that the Lipschitz-constant $L_h$ of $h$ is simply $L_{f,g}$ multiplied by the number of summands. To find this number, note that the way the $r_{i_1}, \ldots, r_{i_n} $ are partitioned is in fact a known combinatorics problem, called stars and bars or multichoose. The problem is formulated as follows: For a fixed number $d$ of stars, and a fixed number $n$ of bins (delimited by $n-1$ bars), how many ways are there to distribute the $d$ stars into the $n$ bins (allowing empty bins). Then the number of ways to partition the integer $d$ into an $n$-tuple is given by $\binom{n+d-1}{d}$, see for example \cite[Chapter II.5]{Feller:68}. Summing over $d = 1, \cdots, R_{\max}$, we can then bound the complete Lipschitz-constant of $h$ by 
    \begin{equation}
        \Bar{L}_h^{\text{poly}} = R_{\max} \cdot n \sqrt{2^n} \left( \binom{n+R_{\max}}{R_{\max}} - 1 \right) \ .
    \end{equation}


\section{Proof of Proposition \ref{cor:rademacher_polynomial}}
\label{app:proof_rademacher_polynomial}

For the class $\mathcal{H}_{n, R_{\max}, C_{\max}}^{\text{poly}} $ of quantum reservoir functionals, begin by writing

\begin{align}
    \label{eq:rademacher_poly_first_steps}
        & \mathbb{E} \left[ \sup_{H \in \mathcal{H}_{n, R_{\max}, C_{\max}}^{\text{poly}}} \left\vert \sum_{j=0}^{k-1} \varepsilon_j H \left( \Tilde{\pmb V}^{(j)} \right) \right\vert \right] \\
        & = \mathbb{E} \left[ \sup_{h \in \mathcal{F}_{n, R_{\max}, C_{\max}}^{\text{poly}} \atop T^{\theta} \in \mathcal{F}_n^{\text{QRC}} (\Theta)} \left\vert \sum_{j=0}^{k-1} \varepsilon_j h \left( \overrightarrow{ \prod}_{i=0}^{\infty} T^{\theta} \left( \Tilde{V}^{(j)}_{-i} \right) \rho_{-\infty}  \right)\right\vert \right] \\
        & = \mathbb{E} \Bigg[ \sup_{ \substack{ w_{i_1,\ldots,i_n}^{r_{i_1},\ldots,r_{i_n}} \in[0,1] \\  0 \leq |C| \leq C_{\max} } } \sup_{\theta \in \Theta} \Bigg\vert \sum_{j=0}^{k-1} \varepsilon_j \Bigg( C  \\
        & \hspace{10mm} +   \sum_{d=1}^{R_{\max}}\sum_{i_1=1}^n\sum_{i_2=i_1+1}^n \! \cdots\! \sum_{i_n = i_{n-1}+1}^n \sum_{r_{i_1}+\cdots + r_{i_n}=d} w_{i_1,\ldots,i_n}^{r_{i_1},\ldots,r_{i_n}} \text{tr}\left[ \sigma_Z^{(i_1)} \rho_0^{\theta, j} \right]^{r_{i_1}} \cdots \text{tr}\left[ \sigma_Z^{(i_n)} \rho_0^{\theta, j} \right]^{r_{i_n}} \Bigg) \Bigg\vert \Bigg] 
    \end{align}
where we have written $\rho_0^{\theta, j} := \overrightarrow{ \prod}_{i=0}^{\infty} T^{\theta} \left( \Tilde{V}^{(j)}_{-i} \right) \rho_{-\infty} $. Using the triangle inequality as well as the subadditivity of the supremum, we can bound the last term as follows:

\begin{align}
    \label{eq:rademacher_poly_triangle}
        & \mathbb{E} \Bigg[ \sup_{ \substack{ w_{i_1,\ldots,i_n}^{r_{i_1},\ldots,r_{i_n}} \in[0,1] \\ 0 \leq |C| \leq C_{\max} } } \sup_{\theta \in \Theta}  \Bigg\vert \sum_{j=0}^{k-1} \varepsilon_j \Bigg( C + \\
        & \hspace{10mm}  \sum_{d=1}^{R_{\max}} \sum_{i_1=1}^n\sum_{i_2=i_1+1}^n \cdots \sum_{i_n = i_{n-1}+1}^n \sum_{r_{i_1}+\cdots + r_{i_n}=d} w_{i_1,\ldots,i_n}^{r_{i_1},\ldots,r_{i_n}} \text{tr}\left[ \sigma_Z^{(i_1)} \rho_0^{\theta, j} \right]^{r_{i_1}} \cdots \text{tr}\left[ \sigma_Z^{(i_n)} \rho_0^{\theta, j} \right]^{r_{i_n}} \Bigg) \Bigg\vert \Bigg] \\
        & \leq \mathbb{E} \left[ \sup_{ \substack{ 0 \leq |C| \leq C_{\max} } } \left\vert \sum_{j=0}^{k-1} \varepsilon_j C \right\vert \right] \\
        & + \mathbb{E} \Bigg[ \sum_{d=1}^{R_{\max}} \sum_{i_1=1}^n \\
        & \hspace{10mm}\! \cdots\! \sum_{i_n = i_{n-1}+1}^n \sum_{r_{i_1}+\cdots + r_{i_n}=d} \sup_{w_{i_1,\ldots,i_n}^{r_{i_1},\ldots,r_{i_n} } \in[0,1] \atop \theta \in \Theta} \left\vert  \sum_{j=0}^{k-1} \varepsilon_j w_{i_1,\ldots,i_n}^{r_{i_1},\ldots,r_{i_n}} \text{tr}\left[ \sigma_Z^{(i_1)} \rho_0^{\theta, j} \right]^{r_{i_1}}\! \cdots\! \text{tr}\left[ \sigma_Z^{(i_n)} \rho_0^{\theta, j} \right]^{r_{i_n}} \right\vert \Bigg] \label{eq:poly_triangle_second_term} \ .
    \end{align}
For the first term, we can write 

    \begin{align}
    \label{eq:simple_bound_poly_C}
        & \left(  \mathbb{E} \left[ \sup_{ \substack{ 0 \leq |C| \leq C_{\max} } } \left\vert \sum_{j=0}^{k-1} \varepsilon_j C \right\vert \right] \right)^2 \leq  \mathbb{E} \left[ \sup_{0 \leq |C| \leq C_{\max}} \left\vert C \right\vert^2 \left\vert \sum_{j=0}^{k-1} \varepsilon_j \right\vert^2  \right] = \sum_{j=0}^{k-1} \mathbb{E} \left[ \varepsilon_j^2 \right] C_{\max}^2  = k  C_{\max}^2
    \end{align}
    where we used Jensen's inequality twice in the first inequality, as well as the fact that $C$ is independent of $j$ and the submultiplicativity of the norm; and the fact that $\mathbb{E}[\varepsilon_j \varepsilon_m] = 0$ if $m \neq j$ in the first equality.

    For the second term in \eqref{eq:poly_triangle_second_term}, we can once again take arguments that are independent of $j$ out of the sum and use the submultiplicativity of the norm to bound the term behind the sums:

\begin{align}
    & \mathbb{E} \left[ \sup_{w_{i_1,\ldots,i_n}^{r_{i_1},\ldots,r_{i_n} } \in[0,1] \atop \theta \in \Theta}  \left\vert  \sum_{j=0}^{k-1} \varepsilon_j w_{i_1,\ldots,i_n}^{r_{i_1},\ldots,r_{i_n}} \text{tr}\left[ \sigma_Z^{(i_1)} \rho_0^{\theta, j} \right]^{r_{i_1}} \cdots \text{tr}\left[ \sigma_Z^{(i_n)} \rho_0^{\theta, j} \right]^{r_{i_n}} \right\vert \right] \\
    & = \mathbb{E} \left[ \sup_{w_{i_1,\ldots,i_n}^{r_{i_1},\ldots,r_{i_n} } \in[0,1] \atop \theta \in \Theta}  \left\vert w_{i_1,\ldots,i_n}^{r_{i_1},\ldots,r_{i_n}}  \sum_{j=0}^{k-1} \varepsilon_j  \text{tr}\left[ \sigma_Z^{(i_1)} \rho_0^{\theta, j} \right]^{r_{i_1}} \cdots \text{tr}\left[ \sigma_Z^{(i_n)} \rho_0^{\theta, j} \right]^{r_{i_n}} \right\vert \right] \\
    & \leq \mathbb{E} \left[ \sup_{w_{i_1,\ldots,i_n}^{r_{i_1},\ldots,r_{i_n} } \in[0,1] \atop \theta \in \Theta}  \left\vert w_{i_1,\ldots,i_n}^{r_{i_1},\ldots,r_{i_n}} \right\vert \left\vert \sum_{j=0}^{k-1} \varepsilon_j  \text{tr}\left[ \sigma_Z^{(i_1)} \rho_0^{\theta, j} \right]^{r_{i_1}} \cdots \text{tr}\left[ \sigma_Z^{(i_n)} \rho_0^{\theta, j} \right]^{r_{i_n}} \right\vert \right] \\
    & \leq \mathbb{E} \left[ \sup_{\theta \in \Theta} \left\vert \sum_{j=0}^{k-1} \varepsilon_j  \text{tr}\left[ \sigma_Z^{(i_1)} \rho_0^{\theta, j} \right]^{r_{i_1}} \cdots \text{tr}\left[ \sigma_Z^{(i_n)} \rho_0^{\theta, j} \right]^{r_{i_n}} \right\vert \right] \ .
\end{align}

Using the fact that the parameter space is finite, we write

\begin{align}
\label{eq:poly_bound_finite_theta}
    & \mathbb{E} \left[ \sup_{\theta \in \Theta} \left\vert \sum_{j=0}^{k-1} \varepsilon_j  \text{tr}\left[ \sigma_Z^{(i_1)} \rho_0^{\theta, j} \right]^{r_{i_1}} \cdots \text{tr}\left[ \sigma_Z^{(i_n)} \rho_0^{\theta, j} \right]^{r_{i_n}} \right\vert \right] \\
    & \leq \sum_{\theta \in \Theta} \mathbb{E} \left[ \left\vert \sum_{j=0}^{k-1} \varepsilon_j  \text{tr}\left[ \sigma_Z^{(i_1)} \rho_0^{\theta, j} \right]^{r_{i_1}} \cdots \text{tr}\left[ \sigma_Z^{(i_n)} \rho_0^{\theta, j} \right]^{r_{i_n}} \right\vert \right] \ .
\end{align}

    We can bound the term behind the sum over the parameter space as follows: 

    \begin{align}
    \label{eq:jensen_poly_j}
        & \left( \mathbb{E} \left[ \left\vert \sum_{j=0}^{k-1} \varepsilon_j \text{tr}\left[ \sigma_Z^{(i_1)} \rho_0^{\theta, j} \right]^{r_{i_1}} \cdots \text{tr}\left[ \sigma_Z^{(i_n)} \rho_0^{\theta, j} \right]^{r_{i_n}} \right\vert \right] \right)^2 \\
        & \leq  \mathbb{E} \left[ \left\vert \sum_{j=0}^{k-1} \varepsilon_j \text{tr}\left[ \sigma_Z^{(i_1)} \rho_0^{\theta, j} \right]^{r_{i_1}} \cdots \text{tr}\left[ \sigma_Z^{(i_n)} \rho_0^{\theta, j} \right]^{r_{i_n}} \right\vert^2  \right] \\
        & = \sum_{j=0}^{k-1} \mathbb{E} \left[ \varepsilon_j^2 \right] \mathbb{E} \left[ \left\vert \text{tr}\left[ \sigma_Z^{(i_1)} \rho_0^{\theta, j} \right]^{r_{i_1}} \cdots \text{tr}\left[ \sigma_Z^{(i_n)} \rho_0^{\theta, j} \right]^{r_{i_n}} \right\vert^2  \right] \\
        & \leq k 
    \end{align}
    where we used Jensen's inequality in the first inequality, independence between the Rademacher variables and the input variables and the fact that $\mathbb{E}[\varepsilon_j \varepsilon_m] = 0$ if $m \neq j$ in the first equality, as well as the fact that $ \text{tr}\left[ \sigma_Z^{(i)} \rho_0^{\theta, j} \right]^{r_{i}} \in [-1, 1]$ .

 Finally, combining \eqref{eq:poly_bound_finite_theta} and \eqref{eq:poly_triangle_second_term}, and using the same combinatorial argument as in Appendix \ref{app:lemma_lipschitz_poly} to bound the number of terms of sums in \eqref{eq:poly_triangle_second_term}, we have

\begin{align}
     & \sum_{d=1}^{R_{\max}} \sum_{i_1=1}^n \!\cdots\! \sum_{i_n = i_{n-1}+1}^n \sum_{r_{i_1}+ \cdots + r_{i_n}=d} \mathbb{E} \left[ \sup_{w_{i_1,\ldots,i_n}^{r_{i_1},\ldots,r_{i_n} } \in[0,1] \atop \theta \in \Theta}  \left\vert  \sum_{j=0}^{k-1} \varepsilon_j w_{i_1,\ldots,i_n}^{r_{i_1},\ldots,r_{i_n}} \text{tr}\left[ \sigma_Z^{(i_1)} \rho_0^{\theta, j} \right]^{r_{i_1}} \!\cdots\! \text{tr}\left[ \sigma_Z^{(i_n)} \rho_0^{\theta, j} \right]^{r_{i_n}} \right\vert \right] \\
     & \leq R_{\max} \left( \binom{n+R_{\max}}{R_{\max}} - 1 \right) |\Theta| \sqrt{k} \ .
\end{align}

Combining this with \eqref{eq:simple_bound_poly_C}, we get the final result.

\section{Proof of Corollary \ref{cor:rademacher_linear}}
\label{app:rademacher_linear}

    We begin by calculating the Lipschitz constants of the readout function. In the following, we write
    \begin{equation}
    \label{eq:omega_lin}
        \Omega_{\text{lin}} := [-C_{\max}, C_{\max}] \times [0, 1]^n \ .
    \end{equation}

    \textbf{Lipschitz-continuity in the density matrices.} For any $(C, \pmb w) \in \Omega_{\text{lin}}$, for all $\rho_1, \rho_2 \in \Bplus$ we have
    \begin{align}
        \left\vert h_{C, \pmb w}(\rho_1) - h_{C, \pmb w} \left( \rho_2 \right) \right\vert & = \left\vert \sum_{i=1}^n w_i \text{tr} \left[\sigma_Z^{(i)} \rho_1 \right] - \sum_{i=1}^n w_i \text{tr} \left[\sigma_Z^{(i)} \rho_2 \right] \right\vert = \left\vert \sum_{i=1}^n w_i \text{tr} \left[\sigma_Z^{(i)} \left( \rho_1 - \rho_2 \right) \right] \right\vert \\
        & \leq \sum_{i=1}^n w_i \left\vert \text{tr} \left[\sigma_Z^{(i)} \left( \rho_1 - \rho_2 \right) \right] \right\vert \leq n \sqrt{2^n} \left\Vert \rho_1 - \rho_2 \right\Vert_2 \ ,
    \end{align}
    where we have used the fact that $w_i \in [0, 1]$ and the inequality \eqref{eq:cs_lipsch}. Thus we have $\Bar{L}_h^{\text{lin}} = n \sqrt{2^n}$.

    \textbf{Lipschitz-continuity in the parameters.} For any $\rho \in \Bplus$, for all $(C, \pmb w), (C', \pmb w') \in \Omega_{\text{lin}}$, we have
    \begin{align}
        \left\vert h_{C, \pmb w}(\rho) - h_{C', \pmb w'} (\rho) \right\vert & = \left\vert C - C' + \sum_{i=1}^n \left(w_i - w'_i\right) \text{tr} \left[ \sigma_Z^{(i)} \rho \right] \right\vert \\
        & \leq \left\vert C - C' \right\vert + \sum_{i=1}^n \left\vert w_i - w_i' \right\vert \left\vert \text{tr} \left[ \sigma_Z^{(i)} \rho \right] \right\vert \\
        & \leq \sqrt{n+1} \left\Vert (C, \pmb w) - \left( C', \pmb w' \right) \right\Vert_2 \ .
    \end{align}
    Thus we have $ L_{\Omega}^{\text{lin}} = \sqrt{n+1}$.

    It is clear that $\text{dim} \left( \Omega_{\text{lin}} \right) = n+1$ and, using the same arguments as in \eqref{eq:diameter_omega_poly} we find $D_{\Omega}^{\text{lin}} = \sqrt{n + 4 C_{\max}^2}$. We can then use Theorem \ref{thm:general_rademacher_dudley} to find the final result.

\section{Proof of Lemma \ref{proposition:lipschitz_sm_inputs}}
\label{app:lipschitz_sm}

The proof of Lemma \ref{proposition:lipschitz_sm_inputs} proceeds similarly to the proof of Lemma \ref{lemma:lipschitz_poly}, with a few modifications in the constants.

Let $h_{C, \pmb W, \pmb b} \in \mathcal{F}_{n, \ell_{\max}, C_{\max}}^{\text{SM}} $. Consider functions $g_i (\rho)$, $1 \leq i \leq n$ as defined in Appendix \ref{app:lemma_lipschitz_poly}. We write
\begin{equation}
    h_{C, \pmb W, \pmb b}(\rho) = f_{C, \pmb W, \pmb b} \left( g_1(\rho), \ldots, g_n(\rho) \right) \ ,
\end{equation}
where we define
\begin{equation}
    f_{C, \pmb W, \pmb b}(x_1, \ldots, x_n) := \prod_{j=1}^{\ell_{\max}} \left( \sum_{i=1}^n W_{j, i} x_i + b_j \right) + C \ .
\end{equation}

Write $a_j := \sum_{i=1}^n W_{j,i} x_i + b_j$ and note that we have $\vert a_j \vert \leq n+1 \ , \ 1 \leq j \leq \ell_{\max}$.

We have, for any $1 \leq i \leq n$,
\begin{equation}
    \left\vert D_{x_i}f_{C, \pmb W, \pmb b} \left( x_1, \ldots, x_n \right) \right\vert = \left\vert \sum_{j=1}^{\ell_{\max}} W_{j,i} \prod_{k=1 \atop k \neq j}^{\ell_{\max}} a_k \right\vert \leq \sum_{j=1}^{\ell_{\max}} \vert W_{j,i} \vert \vert a_k \vert^{\ell_{\max}-1} = \ell_{\max}(n+1)^{\ell_{\max}-1} \ .
\end{equation}

Through the same calculations as in \eqref{eq:ineq_lipschitz_poly}, we obtain, for all $\rho_1, \rho_2 \in \Bplus \cup 0$,
\begin{equation}
    \vert h(\rho_1) - h(\rho_2) \vert \leq n(n+1)^{\ell_{\max}-1}\sqrt{2^n}\ell_{\max} \Vert \rho_1 - \rho_2 \Vert_2 \ .
\end{equation}

\section{Proof of Lemma \ref{proposition:lipschitz_sm_omega}}
\label{app:lipschitz_sm_parameters}

    The proof of Lemma \ref{proposition:lipschitz_sm_omega} proceeds similarly to that of Lemma \ref{proposition:lipschitz_sm_inputs}.

    Reusing the same notation and definitions as in Appendix \ref{app:lipschitz_sm}, we have for all $1 \leq j \leq \ell_{\max}, 1 \leq i \leq n$,
    \begin{equation}
        \left\vert D_{W_{j,i}}f_{C, \pmb W, \pmb b} \left( x_1, \ldots, x_n \right) \right\vert = \left\vert x_i \prod_{k=1 \atop k \neq j}^{\ell_{\max}} a_k \right\vert \leq (n+1)^{\ell_{\max}-1} \ .
    \end{equation}
    Additionally, for all $1 \leq j \leq \ell_{\max}$,
    \begin{equation}
        \left\vert D_{b_j}f_{C, \pmb W, \pmb b} \left( x_1, \ldots, x_n \right) \right\vert = \left\vert \prod_{k=1 \atop k \neq j}^{\ell_{\max}} a_k \right\vert \leq (n+1)^{\ell_{\max}-1} \ .
    \end{equation}
    Finally, we have $\left\vert D_{b_j}f_{C, \pmb W, \pmb b} \left( x_1, \ldots, x_n \right) \right\vert = 1$.

    Putting everything together, we find the Lipschitz constant:
    \begin{align}
        L_{\Omega}^{\text{SM}} & = \sqrt{ \sum_{j=1}^{\ell_{\max}} \sum_{i=1}^n \left( (n+1)^{\ell_{\max}-1} \right)^2 + \sum_{j=1}^n \left( (n+1)^{\ell_{\max}-1} \right)^2 + 1} \\
        & = \sqrt{(n+1)^{2 \ell_{\max} - 1} \ell_{\max} + 1} \ .
    \end{align}


\section{Proof of Lemma \ref{lemma:T_PTR_contractive_Lipschitz}}
\label{app:T_PTR_contractive_Lipschitz}

We begin by proving contractivity in the space of density operators. 

\textbf{Contractivity in the density matrices.}
The proof makes use of \cite[Proposition 1]{rastegin:12}, which states that for an operator $\Tilde{Q}$ defined on a composite system $\mathcal{H}_A\otimes\mathcal{H}_B$ with $\text{dim}(\mathcal{H}_A) = m$ and $\text{dim}(\mathcal{H}_B) = n$ we have

\begin{equation}
\label{eq:partialtracebound}
    \left\lVert \text{tr}_B \left[ \Tilde{Q} \right] \right\rVert_2 \leq \sqrt{n} \left\lVert \Tilde{Q} \right\rVert_2 \ .
\end{equation}

To show that there exists an $r_{\text{PTR}} \in (0,1)$ such that for fixed parameters $(J, \gamma, \tau) \in \Theta_{\text{PTR}}$, for any two quantum states $\rho_1, \rho_2$ and for all $v \in D_v$

\begin{equation}
    \left\lVert T^{J, \gamma, \tau} (v, \rho_1) - T^{J, \gamma, \tau} (v, \rho_2) \right\rVert_2 \leq r_{\text{PTR}} \left\lVert \rho_1 - \rho_2 \right\rVert_2 \ ,
\end{equation}

note that $T^{J, \gamma, \tau}$ is a linear map, and $\rho_1, \rho_2$ being quantum states, we have $\text{tr}[\rho_1-\rho_2]=0$ so that we can restrict $T^{J, \gamma, \tau}$ to the hyperplane $\mathcal{B}_0$ of $2^n\times 2^n$ traceless Hermitian operators. In the remainder of this proof we write $T = T^{J, \gamma, \tau} $ and $H = H(J, \gamma) \tau$ for ease of notation, as the parameters do not intervene in the calculations.
Define
$ \mathcal{V}=\begin{pmatrix}
v & 0 \\
0 & 1 - v
\end{pmatrix}$. 
Using (\ref{eq:partialtracebound}) we have for all $ A \in \mathcal{B}_0$,

\begin{equation}
    \left\lVert T(v, A) \right\rVert_2 = \left\lVert \text{tr}_0\left[e^{-iH} A\otimes \mathcal{V} e^{iH}\right] \right\rVert_2 \leq \sqrt{2} \left\lVert e^{-iH} A \otimes \mathcal{V} e^{iH}  \right\rVert_2 \ .
\end{equation}

Notice also that

\begin{align}
    \left\lVert e^{-iH} A \otimes \mathcal{V} e^{iH} \right\rVert_2 & = \left\lVert A \otimes \mathcal{V} \right\rVert_2 = \sqrt{\text{tr}\left[ \left[ A \otimes \mathcal{V} \right]^* \left[ A \otimes \mathcal{V} \right] \right] } \\
    & = \sqrt{\text{tr} \left[ A^* A \otimes \mathcal{V}^* \mathcal{V} \right] } = \sqrt{\text{tr} \left[ A^* A \right ] \text{tr} \left[ \mathcal{V}^* \mathcal{V} \right] } = \left\lVert A \right\rVert_2 \left( v^2 + (1-v) ^ 2 \right) \ .
\end{align}

We thus have, for $v \in \left[ \frac{1}{2}(1-\sqrt{\sqrt{2}-1}) + \epsilon_{\text{PTR}}, \frac{1}{2}(1+\sqrt{\sqrt{2}-1}) - \epsilon_{\text{PTR}} \right]$, 

\begin{equation}
\sup_{ A \in \mathcal{B}_0 \atop \left\lVert A \right\rVert_2 = 1 } \left\lVert T(v, A) \right\rVert_2 \leq  r_{\text{PTR}} (\epsilon_{\text{PTR}})
\end{equation}

with 

\begin{align}
    r_{\text{PTR}} (\epsilon_{\text{PTR}}) & = \sqrt{2} \left( \left( \frac{1}{2}(1+\sqrt{\sqrt{2}-1}) - \epsilon_{\text{PTR}} \right)^2 + \left( 1 - \frac{1}{2}(1+\sqrt{\sqrt{2}-1}) + \epsilon_{\text{PTR}} \right)^2 \right) \\
    & = 2 \sqrt{2} \left( \epsilon_{\text{PTR}}^2 - \epsilon_{\text{PTR}} \sqrt{ \sqrt{2} -1 } \right) +1
\end{align}

where we indicate the dependence of the contractivity constant on the choice of $\epsilon_{\text{PTR}}$

\textbf{Lipschitz continuity in the inputs.}
We proceed similarly to prove Lipschitz-continuity: For any inputs $v_1, v_2 \in D_v$ and any density matrix $\rho \in \Bar{\mathcal{B}}^+$ we have

    \begin{align}
        & \left\lVert T(v_1, \rho) - T(v_2, \rho) \right\rVert_2 \\
        & = \left\lVert \text{tr}_0 \left[ e^{-iH} \rho \otimes \begin{pmatrix} v_1 & 0 \\ 0 & 1 - v_1 \end{pmatrix} e^{iH} \right] - \text{tr}_0 \left[ e^{-iH} \rho \otimes \begin{pmatrix} v_2 & 0 \\ 0 & 1 - v_2 \end{pmatrix} e^{iH} \right] \right\rVert_2 \\
        & = \left\lVert \text{tr}_0 \left[ e^{-iH} \rho \otimes \begin{pmatrix} v_1 & 0 \\ 0 & 1 - v_1 \end{pmatrix} e^{iH} - e^{-iH} \rho \otimes \begin{pmatrix} v_2 & 0 \\ 0 & 1 - v_2 \end{pmatrix} e^{iH} \right] \right\rVert_2 \\
        & = \left\lVert \text{tr}_0 \left[ e^{-iH} \left( \rho \otimes \begin{pmatrix} v_1 & 0 \\ 0 & 1 - v_1 \end{pmatrix} - \rho \otimes \begin{pmatrix} v_2 & 0 \\ 0 & 1 - v_2 \end{pmatrix} \right) e^{iH} \right] \right\rVert_2 \\
        & = \left\lVert \text{tr}_0 \left[ e^{-iH} \rho \otimes \begin{pmatrix} v_1 - v_2 & 0 \\ 0 & v_2 - v_1 \end{pmatrix} e^{iH} \right] \right\rVert_2 \\
        & \leq \sqrt{2} \left\lVert \rho \otimes \begin{pmatrix} v_1 - v_2 & 0 \\ 0 & v_2 - v_1 \end{pmatrix} \right\rVert_2 \\
        & = \sqrt{2 \text{tr} \left[ \left( \rho^{\dag} \otimes \begin{pmatrix} v_1 - v_2 & 0 \\ 0 & v_2 - v_1 \end{pmatrix} \right) \left (\rho \otimes \begin{pmatrix} v_1 - v_2 & 0 \\ 0 & v_2 - v_1 \end{pmatrix} \right) \right]} \\
        & = \sqrt{2 \text{tr} \left( \rho^{\dag} \rho \right) \text{tr} \left (\begin{pmatrix} v_1 - v_2 & 0 \\ 0 & v_2 - v_1 \end{pmatrix}^2 \right) } \\
        & \leq \sqrt{2 \cdot 2 (v_1 - v_2)^2 } = 2 \lVert v_1 - v_2 \rVert_2
    \end{align}

where the second equality follows from the linearity of the partial trace operator, the first inequality follows from (\ref{eq:partialtracebound}), and the second inequality follows from the fact that the purity of any quantum state is bounded by one. 

\textbf{Lipschitz continuity in the parameters.}
To show Lipschitz continuity in the parameters, let $v \in D_v$ and $\rho \in \Bplus$ be a fixed input and density matrix respectively. Let us write $\theta = (J, \gamma)$ and $H(\theta) = H(J, \gamma)$ for ease of notation, as well as $U(\theta) := e^{iH(\theta)}$. Then, for any two sets of parameters $\theta_1, \theta_2 \in \Theta_{\text{PTR}}$, we have
\begin{align}
    & \left\Vert T^{\theta_1}(v, \rho) - T^{\theta_2}(v, \rho) \right\Vert_2 \\
    & = \left\Vert \text{tr}_0\left[U^{\dag}(\theta_1) \rho \otimes \mathcal{V} U(\theta_1) - U^{\dag}(\theta_2) \rho \otimes \mathcal{V} U(\theta_2)\right] \right\Vert_2 \\
    & \leq \sqrt{2} \left\Vert U^{\dag}(\theta_1) \rho \otimes \mathcal{V} U(\theta_1) - U^{\dag}(\theta_2) \rho \otimes \mathcal{V} U(\theta_2) \right\Vert_2 \\
    & = \sqrt{2} \left\Vert U^{\dag}(\theta_1) \rho \otimes \mathcal{V} \left(U(\theta_1) - U(\theta_2) \right) + \left(U^{\dag}(\theta_1) - U^{\dag}(\theta_2)  \right)\rho \otimes \mathcal{V} U(\theta_2) \right\Vert_2 \\
    & \leq \sqrt{2^{n+1}} \left( \left\Vert  U^{\dag}(\theta_1) \right\Vert_2 \left\Vert \rho \otimes \mathcal{V}\right\Vert_2 \left\Vert U(\theta_1) - U(\theta_2)\right\Vert_2 + \left\Vert U^{\dag}(\theta_1) - U^{\dag}(\theta_2)\right\Vert_2 \left\Vert \rho \otimes \mathcal{V} \right\Vert_2 \left\Vert U(\theta_2) \right\Vert_2 \right) \\
    & = \sqrt{2^{n+1}}  \left\Vert \rho \otimes \mathcal{V}\right\Vert_2 \left(\left\Vert U(\theta_1) - U(\theta_2)\right\Vert_2 + \left\Vert U^{\dag}(\theta_1) - U^{\dag}(\theta_2)\right\Vert_2 \right)\\
    & = 2 \sqrt{2^{n+1}}  \left\Vert \rho \right\Vert_2 \left\Vert \mathcal{V} \right\Vert_2 \left\Vert U(\theta_1) - U(\theta_2)\right\Vert_2 \\
    & \leq 2 \sqrt{2^{n}}  \left\Vert U(\theta_1) - U(\theta_2)\right\Vert_2 \ ,
\end{align}
where we once again used \eqref{eq:partialtracebound} in the first inequality and the fact that $\left\Vert\mathcal{V}\right\Vert_2 < \frac{1}{\sqrt{2}}$ in the last.

In the following, we write $\theta = (\theta^{1}, \ldots, \theta^{n(n-1) + 1})$, where $(\theta^{1}, \ldots, \theta^{n(n-1)}) = \left\{ J^{ij} \right\} $ and $\theta^{(n(n-1)+1)} = \gamma$. Using the integral identity of the derivative of $e^{iH(\theta)}$ for bounded operators,
\begin{equation}
    \frac{\partial}{\partial \theta^{j}} e^{iH(\theta)} = \int_0^1 e^{i(1-s)H(\theta)} \frac{\partial i H(\theta)}{\partial \theta^{j}} e^{i \, sH(\theta)} ds \ ,
\end{equation}
(see for example \cite{wilcox:67}), we can write
\begin{equation}
    \left\Vert \frac{\partial}{\partial \theta^{j}} e^{iH(\theta)} \right\Vert_2 \leq \int_0^1 \left\Vert e^{i(1-s)H(\theta)} \frac{\partial i H(\theta)}{\partial \theta^{j}} e^{i \, sH(\theta)} \right\Vert_2 ds =  \int_0^1 \left\Vert \frac{\partial i H(\theta)}{\partial \theta^{j}} \right\Vert_2 ds =  \left\Vert \frac{\partial H(\theta)}{\partial \theta^{j}} \right\Vert_2 \ ,
\end{equation}
where we used the fact that $ e^{isH(\theta)}$ and $e^{i(1-s)H(\theta)} $ are unitary. Using the same argument as in the proof of Lemma \ref{lemma:lipschitz_poly}, we conclude that the map $\theta \mapsto e^{iH(\theta)}$ is Lipschitz-continuous w.r.t. $\theta$, where we have
\begin{equation}
    \left\Vert e^{iH(\theta_1)} - e^{iH(\theta_2)} \right\Vert_2 \leq \sqrt{ \sum_{j=1}^{n(n-1) + 1} \left(\sup_{\theta} \left\Vert \frac{\partial H(\theta)}{\partial \theta^{j}} \right\Vert_2\right)^2 }\Vert \theta_1 - \theta_2 \Vert_2 \ .
\end{equation}

It remains to calculate the partial derivatives of the Hamiltonian. We find
\begin{align}
    \left\Vert \frac{\partial H(\theta)}{\partial \gamma} \right\Vert_2 & = \left\Vert \sum_{i=1}^n \sigma_Z^{(i)} \right\Vert_2 = \sqrt{\text{tr}\left[ \left( \sum_{i=1}^n \sigma_Z^{(i)} \right) \left( \sum_{i=1}^n \sigma_Z^{(i)}\right) \right] } \\
    & = \sqrt{\text{tr} \left[ \sum_{i=1}^n \left( \sigma_Z^{(i)} \right)^2 + \sum_{i=1}^n \sum_{j \neq i} \sigma_Z^{(i)} \sigma_Z^{(j)} \right] } \\
    & = \sqrt{\sum_{i=1}^n \text{tr} \left[ I \right] + \sum_{i=1}^n \sum_{j \neq i} \text{tr}\left[ I^{\otimes (n-i)} \otimes \sigma_Z \otimes I^{\otimes(j-i-1)} \otimes \sigma_Z \otimes I^{\otimes(i-1)} \right] } \\
    & = \sqrt{n 2^n} \ ,
\end{align}
where we have used the fact that $\text{tr}\left[ A \otimes B\right] =\text{tr} \left[A\right] \text{tr} \left[B\right] $ as well as the fact that the trace of Pauli matrices is zero.

For the remaining parameters, we find, for any $1 \leq i, j \leq n$,
\begin{align}
    \left\Vert \frac{\partial H(\theta)}{\partial J^{i,j}} \right\Vert_2 & = \left\Vert \sigma_X^{(i)}\sigma_X^{(j)} + \sigma_Y^{(i)}\sigma_Y^{(j)} \right\Vert_2 \\
    & = \left\Vert I^{\otimes (i-1)} \otimes \sigma_X \otimes I^{\otimes (j-i-1)} \otimes \sigma_X \otimes I^{(n-i)} +  I^{\otimes (i-1)} \otimes \sigma_Y \otimes I^{\otimes (j-i-1)} \otimes \sigma_Y \otimes I^{(n-i)}   \right\Vert_2 \\
    & = \left\Vert I^{\otimes (i-1)} \otimes \left( \sigma_X \otimes I^{\otimes (j-i-1)} \otimes \sigma_X +\sigma_Y \otimes I^{\otimes (j-i-1)} \otimes \sigma_Y \right) \otimes I^{(n-i)} \right\Vert_2 \\
    & \leq \left\Vert I^{\otimes (i-1)} \right\Vert_2 \left( \left\Vert I^{(j-i-1)} \otimes \sigma_X \otimes \sigma_X  \right\Vert_2 + \left\Vert I^{(j-i-1)} \otimes \sigma_Y \otimes \sigma_Y  \right\Vert_2  \right) \left\Vert I^{\otimes (n-i)} \right\Vert_2 \\
    & \leq \sqrt{2^{n-2}}\left( \left\Vert \sigma_X \otimes \sigma_X  \right\Vert_2 + \left\Vert \sigma_Y \otimes \sigma_Y  \right\Vert_2  \right) \leq \sqrt{2^{n-2}} \left( \left\Vert \sigma_X \right\Vert_2^2 + \left\Vert \sigma_Y \right\Vert_2^2 \right) = 4 \cdot \sqrt{2^{n-2}} = \sqrt{2^{n+2}} \ ,
\end{align}
where we have used the fact that $\left\Vert A \otimes B \right\Vert \leq \left\Vert A \right\Vert \left\Vert B\right\Vert$ as well as $\left\Vert A \otimes B \right\Vert_2 = \left\Vert B \otimes A \right\Vert_2$
For the various tensor product identities used in this proof, see for example \cite[Chapter 13]{laub:04}.

Putting everything together, we find that the CPTP map $T^{(J, \gamma)}$ is Lipschitz-continuous w.r.t. the parameters with Lipschitz constant $2 \sqrt{n 2^{2n} + n(n-1)2^{2n+2}} $ and thus all CPTP maps in $\mathcal{F}_n^{\text{PTR}}\left( \Theta_{\text{PTR}} \right)$ are Lipschitz-continuous with Lipschitz constants bounded by $L_{\Theta}^{\text{PTR}} = 2^{n+1}\sqrt{ n (4n - 3)}$.

\section{Proof of Lemma \ref{lemma:RRR_AT}}
\label{app:T_RRR_contractive_Lipschitz}

\textbf{Contractivity in the density matrices.} We begin by proving contractivity in the space of density operators: For any $\rho_1, \rho_2 \in \bar{\mathcal{B}}^+$ and for any fixed $v \in D_v^{\text{RRR}}$ we have

\begin{align}
    & \left\lVert T^{\alpha, \sigma} (v, \rho_1) - T^{\alpha, \sigma} (v, \rho_2) \right\rVert_2 \\
    & = (1 - \alpha) \left\lVert (v T_0 + (1 - v) T_1) \rho_1 - (v T_0 + (1 - v) T_1) \rho_2 \right\rVert_2 \\
    & \leq (1 - \alpha) ( v \left\lVert T_0(\rho_1) - T_0(\rho_2) \right\rVert_2 + (1-v) \left\lVert T_1(\rho_1) - T_1(\rho_2) \right\rVert_2 ) \\
    & \leq (1 - \alpha) ( v r_0  + (1 - v) r_1) \left\lVert \rho_1 - \rho_2 \right\rVert_2 \\
    & = (1 - \alpha) ( v (r_0 - r_1) + r_1) \left \lVert \rho_1 - \rho_2 \right\rVert_2
\end{align}

 where we used the fact that $T_0$ and $T_1$ are $r_0$- and $r_1$-contractions, respectively. If $r_0 = r_1 = r_{0,1}$, the result is immediate noting that $\alpha \in [\alpha_{\min}, \alpha_{\max}] \subset (0,1)$. If $r_0 > r_1$, we have $1 > r_0 + r_1$ which implies $0 < r_0 - r_1 < 1 - 2 r_1$, thus 

 \begin{align}
     v (r_0 - r_1) + r_1 < v (1 - 2 r_1) + r_1 = r_1 (1 - 2 v) + v < 1 - v < 1 - \epsilon_{\text{RRR}}
 \end{align}

where the second to last inequality follows from the fact that $r_1 < 1$ and $v < 1/2$.

\textbf{Lipschitz continuity in the inputs.}
The proof of the Lipschitz continuity is a straightforward application of the triangle inequality. For any $v_1, v_2 \in D_v^{\text{RRR}}$ and for any fixed $\rho \in \Bar{\mathcal{B}}^+$ we have

\begin{align}
    & \left\lVert T^{\alpha, \sigma} (v_1, \rho) - T^{\alpha, \sigma} (v_2, \rho) \right\rVert_2 \\
    & = \left\lVert (1 - \alpha) \left( v_1 T_0 + (1 - v_1) T_1 \right) \rho + \alpha \sigma - (1 - \alpha) \left( v_2 T_0 + (1 - v_2) T_1 \right) \rho - \alpha \sigma \right\rVert_2 \\
    & \leq (1 - \alpha) \left\lVert (v_1 - v_2) T_0(\rho) + (v_2 - v_1) T_1(\rho)  \right\rVert_2 \\
    & \leq (1 - \alpha) \left| v_1 - v_2 \right| \left\lVert T_0(\rho) + T_1(\rho) \right\rVert_2 \\
    & \leq (1 - \alpha) \left| v_1 - v_2 \right|\left( \left\lVert T_0(\rho) \right\rVert_2 + \left\lVert T_1(\rho_1) \right\rVert_2 \right) \\
    & \leq 2 (1 - \alpha_{\min}) \left| v_1 - v_2 \right| \\
\end{align}

where the last inequality comes from the fact that $T_0$ and $T_1$ are CPTP maps and thus produce a density matrix with purity bounded by $1$.

\textbf{Lipschitz continuity in the parameters.}
Lastly, we verify Lipschitz continuity in the parameters: Let $v \in D_v^{\text{RRR}}$ and $\rho \in \Bplus$ be a fixed input and density matrix respectively. Let us write $\mathcal{T} := \left( v T_0 + (1-v) T_1 \right)(\rho)$ as well as $\theta = (\alpha, \sigma) \in \Theta_{\text{RRR}}$. Then we have, for any $\theta_1, \theta_2 \in \Theta_{\text{RRR}}$
\begin{align}
    \left\Vert T^{\theta_1} (v, \rho) - T^{\theta_2} (v, \rho) \right\Vert_2 & = \left\Vert (1 - \alpha_1) \mathcal{T} + \alpha_1 \sigma_1 - (1-\alpha_2) \mathcal{T} - \alpha_2\sigma_2  \right\Vert_2 \\
    & = \left\Vert (\alpha_2 - \alpha_1) \mathcal{T} + \alpha_1 \sigma_2 - \alpha_2 \sigma_2 \right\Vert_2 \\
    & = \left\Vert (\alpha_2 - \alpha_1) \mathcal{T} + \alpha_1(\sigma_1 - \sigma_2) + (\alpha_1 - \alpha_2) \sigma_2  \right\Vert_2 \\
    & = \left\Vert (\alpha_2 - \alpha_1) (\mathcal{T} - \sigma_2) + \alpha_1 (\sigma_1 - \sigma_2) \right\Vert_2 \\
    & \leq \vert \alpha_1 - \alpha_2 \vert \left\Vert \mathcal{T} - \sigma_2 \right\Vert_2 + \alpha_1 \left\Vert \sigma_1 - \sigma_2 \right\Vert_2 \\
    & \leq \sqrt{2} \vert \alpha_1 - \alpha_2 \vert + \alpha_1 \Vert \sigma_1 - \sigma_2 \Vert_2 \\
    & \leq \sqrt{2} \left( \vert \alpha_1 - \alpha_2 \vert + \Vert \sigma_1 - \sigma_2 \Vert_2 \right) \\
    & \leq 2 \left\Vert \theta_1 - \theta_2 \right\Vert_2 \ ,
\end{align}
where we have used the fact that for two density matrices we have $\left\Vert \mathcal{T} - \sigma_2 \right\Vert_2 \leq \sqrt{2}$ in the second inequality, the fact that $\alpha < 1$ in the third, and the fact that for any positive numbers $a, b$ we have $a + b \leq \sqrt{2} \sqrt{a^2 + b^2}$. Thus, the map $T^{\alpha, \sigma}$ is Lipschitz continuous w.r.t the parameters $(\alpha, \sigma)$, with Lipschitz constant $ L_{\Theta}^{\text{RRR}} = 2 $, for all values of $\theta \in \Theta_{\text{RRR}}$.

\section{Proof of Theorem \ref{cor:risk_bound_ptr_O}}
\label{app:risk_bound_ptr_O}

The proof of this theorem hinges on the application of \cite[Theorem 14]{gonon:20} which we restate here while adapting to our particular case:

\begin{theorem} 
\label{theorem:14}
    Let $\mathcal{H}_{n}^{\text{QRC}}$ be the hypothesis class of quantum reservoir functionals specified in \eqref{eq:general_reservoirclass} associated to a class $\mathcal{F}_n^{\text{QRC}} (\Theta)$ of reservoir maps verifying the conditions $A^T$ and a class $ \mathcal{F}_n^O$ of readout maps verifying conditions $A^h$. Suppose that both the input $V$ and the target $Y$ processes verify the hypotheses $A^{IO}$. Assume additionally that there exists a constant $C_{\text{QRC}}$ such that the Rademacher complexity satisfies $ \mathcal{R}_k(\mathcal{H}_{n}^{\text{QRC}}) \leq C_{\text{QRC}} / \sqrt{k}$. Furthermore, define $\zeta_{\max} := \max (r, D_{w^y}, D_{w^v})$. Then there exist constants $C_0, C_1, C_2, C_3, C_{\text{bd}} > 0$ such that for all $m \in \mathbb{N}^+$ satisfying $\log{m} < m \log{\zeta_{\max}^{-1}}$ and for all $\delta \in (0, 1)$ with probability at least $1 - \delta$ we have

    \begin{align}
    \label{eq;general_risk_bound}
         \sup_{H \in \mathcal{H}_n^{\text{QRC}} } \left| R(H) - \hat{R}_m(H) \right| \leq \frac{\left( 1 - r^m \right) C_0^{\text{QRC}} + C_1}{m} + \frac{C_2 \log{m}}{m} + \frac{C_3 \sqrt{\log{m}}}{\sqrt{m}} + \frac{C_{\text{bd}} \sqrt{\log{ \frac{4}{\delta}} }}{\sqrt{2 m}}
    \end{align}

    with constants

    \begin{align}
    \label{eq:constants_th_14}
        & C_1 = \frac{L_{\ell} (2 \Bar{L}_h + C_y)}{\zeta_{\max}} \\
        & C_2 = \frac{2 S}{ \log{\zeta_{\max}^{-1}}} + \frac{L_{\ell} L_R \bar{L}_h C_v}{\zeta_{\max} \log{\zeta_{\max}^{-1}}} \\
        & C_3 = \frac{2 L_{\ell} C_{\text{QRC}}}{\sqrt{\log{\zeta_{\max}^{-1}}}} \\
        & C_{\text{bd}} = 2 L_{\ell} \left( \frac{\Bar{L}_h}{1-r} \left( r + L_R M_{\xi} L_v \left\lVert w^v \right\rVert_1 \right) + M_{\xi} L_y \left\lVert w^y \right\rVert_1 \right)
    \end{align}

    where $ S = L_{\ell} \Bar{L}_h + \mathbb{E}\left[ \left| \ell (0, Y_0) \right| \right] + L_{h,0} L_{\ell} \ , \ C_I = \frac{2 L_I \mathbb{E} \left[ \left\lVert \xi_0^I\right\rVert_2 \right] }{1 - D_{w^I}} \ , \ I = v, y$ and $C_0^{\text{QRC}} = \frac{ 2 r L_{\ell} \Bar{L}_h }{1 - r}$.
    
\end{theorem}

Note that in the adaptation of the theorem we have used the fact that any convergent reservoir functional is bounded by one in the space of density matrices, and have replaced the constant $M_{\mathcal{F}}$ by one.  The dependence in the upper bound of the Rademacher complexity intervenes in the third summand in \eqref{eq;general_risk_bound}.

Then we can calculate the constant factor in the Rademacher complexity  $C_0^{\text{PTR}}$ by plugging the constants from Lemma \ref{lemma:T_PTR_contractive_Lipschitz} into Theorem \ref{theorem:14}:

\begin{align}
    C_0^{\text{PTR}} & =  \frac{ 2  r_{\text{PTR}} (\epsilon_{\text{PTR}}) L_{\ell} \Bar{L}_h^{\text{poly}} }{1 -  r_{\text{PTR}} (\epsilon_{\text{PTR}})}\\
    & = \frac{\left( 2 \sqrt{2} \left( \epsilon_{\text{PTR}}^2 - \epsilon_{\text{PTR}} \sqrt{ \sqrt{2} -1 } \right) + 1 \right) L_{\ell}  R_{\text{max}} \cdot n \sqrt{2^n} \left( \binom{n+R_{\text{max}}}{R_{\text{max}}} - 1 \right)}{\sqrt{2} \left( \epsilon_{\text{PTR}}^2 - \epsilon_{\text{PTR}} \sqrt{ \sqrt{2} -1 } \right)} \ .
\end{align}

Combining Lemma \ref{lemma:T_PTR_contractive_Lipschitz} with Proposition \ref{prop:rademacher_polynomial}, we calculate the constant factor in the bound on the Rademacher complexity:
\begin{align}
    & C_{\text{PTR}} = 48 \cdot 0.89 \cdot \left(\sqrt{n(n-1)} + \sqrt{\mathcal{N}_{\text{poly}} + 1} \right) \\
     \cdot \max\Bigg( & \frac{ R_{\max} \mathcal{N}_{poly}}{r_{\text{PTR}} (\epsilon_{\text{PTR}})} \sqrt{\left( 2^{n+1} n \sqrt{n (4n -3)} \right)\left( n(n-1) \left(J_{\max} - J_{\min} \right)^2 + \left(\gamma_{\max} - \gamma_{\min} \right)^2 \right)} \ , \\
    & \sqrt{\left(\mathcal{N}_{\text{poly}} + 4 C_{\max}^2 \right) \left( \mathcal{N}_{\text{poly}} + 1 \right)} \Bigg) \ .
\end{align}

 We can then calculate the explicit bound which we state in Appendix \ref{app:explicit_risk_bounds} by injecting the above constants and all relevant constants derived in this document into the bound in \eqref{eq;general_risk_bound}.

\section{Proof of Theorem \ref{cor:risk_bound_rrr_O}}
\label{app:risk_bound_rrr_O}

The proof of this theorem is analogous to that of Theorem \ref{cor:risk_bound_ptr_O}, but distinguishing the case where $r_0 = r_1$ from the case where $r_0 > r_1$.

As in Appendix \ref{app:risk_bound_ptr_O}, we calculate $C_0^{\text{RRR}}$ by plugging the constants from Lemma \ref{lemma:RRR_AT} into Theorem \ref{theorem:14} while distinguishing the case when $r_0 = r_1 = r_{0,1}$ from the case when $r_0 > r_1$ :

    \begin{align}
    \label{eq:C_0_RRR}
        C_0^{\text{RRR}} = \begin{cases}
            \frac{2(1 - \alpha_{\min}) r_{0,1} L_{\ell}  R_{\max} \cdot n\sqrt{2^n} \left( \binom{n+R_{\max}}{R_{\max}} - 1 \right)}{ 1 - (1 - \alpha_{\min}) r_{0,1} } & \text{ if } r_0 = r_1 = r_{0,1} \\
            \frac{2 (1 - \epsilon_{\text{RRR}}) L_{\ell}  R_{\max} \cdot n \sqrt{2^n} \left( \binom{n+R_{\max}}{R_{\max}} - 1 \right)}{\epsilon_{\text{RRR}}} & \text{ if } r_0 > r_1
            \end{cases}
    \end{align} 

Using Proposition \ref{prop:rademacher_polynomial} and Lemma \ref{lemma:RRR_AT}, we calculate 
\begin{align}
    C_{\text{RRR}} = 48 \cdot & 0.89\left( \sqrt{4^n} + \sqrt{\mathcal{N}_{\text{poly}} + 1} \right) \cdot \\
    & \max \left\{ \frac{2}{r_{\text{RRR}}}n\sqrt{2^n} R_{\max} \mathcal{N}_{\text{poly}} \sqrt{(\alpha_{\max} - \alpha_{\min})^2 + 2} \, , \, \sqrt{\left( \mathcal{N}_{\text{poly}} + 4 C_{\max}^2 \right) \left( \mathcal{N}_{\text{poly}} + 1 \right)} \right\} \ .
\end{align}

 We can then calculate the explicit bound which we state in Appendix \ref{app:explicit_risk_bounds} by injecting the above constants and all relevant constants derived in this document into the bound in \eqref{eq;general_risk_bound}. 

\section{Explicit risk bounds for $\mathcal{H}_{n, R_{\text{max}}, C_{\text{max}}}^{\text{PTR}}$ and $\mathcal{H}_{n, R_{\text{max}}, C_{\text{max}}}^{\text{RRR}}$}
\label{app:explicit_risk_bounds}

Replacing the general constants $r$, $L_R$ and $C_0^{\text{QRC}}$ in Theorem \ref{theorem:14} with their more explicit forms for the PTR reservoir class $\mathcal{H}_{n, R_{\max}, C_{\max}}^{\text{PTR}}$ as well as the RRR reservoir class $\mathcal{H}_{n, R_{\max}, C_{\max}}^{\text{RRR}}$ that we establish in Appendix \ref{app:risk_bound_ptr_O} and Appendix \ref{app:risk_bound_rrr_O} respectively, we obtain the generalisation bounds, which we state in the following with the explicit constants. 

Under the same hypotheses as in Theorem \ref{cor:risk_bound_ptr_O}, for all $\delta \in (0, 1)$, with probability at least $1 - \delta$ we have

 \begin{align}
    \label{eq:T_PTR_risk_bound}
         \sup_{H \in \mathcal{H}_{n, R_{\max}, C_{\max}}^{\text{PTR}}} & \left| R(H) - \hat{R}_m(H) \right| \\
        & \leq \left( C_0^P \left( n, R_{\max}, \epsilon_{\text{PTR}} \right) + C_1 \left( n, R_{\max}, \zeta_{\max} \right)
        \right) \frac{1}{m} \\
        & + C_2^P \left(  n, R_{\text{max}}, \zeta_{\text{max}}, C_{\text{max}} \right) \frac{ \log{m} }{m} \\
        & +  C_3^P \left( n, R_{\text{max}}, \zeta_{\text{max}}, \epsilon_{\text{PTR}}, J_{\max}, J_{\min}, \gamma_{\max}, \gamma_{\min}, C_{\max} \right) \sqrt{\frac{ \log{m} }{m}} \\
        & + C_4^P \left(  n, R_{\text{max}}, \zeta_{\text{max}}, \epsilon_{\text{PTR}} \right) \sqrt{ \frac{\log{4/\delta}}{2 m} }
    \end{align}

    where 

    \begin{align}
        &  C_0^P \left(  n, R_{\max}, \epsilon_{\text{PTR}} \right) = \frac{ 2\left( 1 - r_{\text{PTR}} \left( \epsilon_{\text{PTR}} \right)^m \right) r_{\text{PTR}} \left( \epsilon_{\text{PTR}} \right) L_{\ell} \Bar{L}_h^{\text{poly}} }{1 - r_{\text{PTR}} \left( \epsilon_{\text{PTR}} \right)} \\ \label{eq:constants_PTR_C_1}
        &  C_1 \left(  n, R_{\max}, \zeta_{\max} \right) = L_{\ell} ( 2 \Bar{L}_h^{\text{poly}} + C_y ) \zeta_{\text{max}}^{-1} \\
        &  C_2^P \left(   n, R_{\text{max}}, \zeta_{\text{max}}, C_{\text{max}} \right) \\
        & \hspace{5mm} = \frac{ 2 \left( L_{\ell} \Bar{L}_h^{\text{poly}} + \mathbb{E} \left[ | \ell (0, Y_0) | \right] + C_{\text{max}} L_{\ell} \right) }{\log{\zeta_{\text{max}}^{-1}}} + \frac{2 L_{\ell} \Bar{L}_h^{\text{poly}}C_v}{ \zeta_{\text{max}} \log{\zeta_{\text{max}}^{-1}} } \\ \label{eq:constants_PTR_C_3}
        & C_3^P \left( n, R_{\text{max}}, \zeta_{\text{max}}, \epsilon_{\text{PTR}}, J_{\max}, J_{\min}, \gamma_{\max}, \gamma_{\min}, C_{\max} \right) \\
        & = \frac{2 \cdot 48 \cdot 0.89 \cdot \left(\sqrt{n(n-1)+1} + \sqrt{\mathcal{N}_{\text{poly}} + 1} \right) L_{\ell}}{\sqrt{\log{\zeta_{\text{max}}^{-1}}}} \cdot \\
        & \max\Bigg(\frac{R_{\max} \mathcal{N}_{poly}}{r_{\text{PTR}} (\epsilon_{\text{PTR}})} n 2^{n+1} \sqrt{ 2^n n (4n -3) \left( n(n-1) \left(J_{\max} - J_{\min} \right)^2 + \left(\gamma_{\max} - \gamma_{\min} \right)^2 \right)} \ , \\
        & \sqrt{\left(\mathcal{N}_{\text{poly}} + 4 C_{\max}^2 \right) \left( \mathcal{N}_{\text{poly}} + 1 \right)} \Bigg) \\
        & C_4^P \left(  n, R_{\text{max}}, \zeta_{\text{max}}, \epsilon_{\text{PTR}} \right) \\
        & \hspace{5mm} = 2 L_{\ell} \left( \frac{\Bar{L}_h^{\text{poly}}}{ 1 - r_{\text{PTR}} \left( \epsilon_{\text{PTR}} \right) } \left( r_{\text{PTR}} \left( \epsilon_{\text{PTR}} \right) + 2 M_{\xi} L_v  \lVert w^v \rVert_1 \right) + M_{\xi} L_y \lVert w^y \rVert_1 \right)
    \end{align}

    and $\Bar{L}_h^{\text{poly}}$ is as in \eqref{eq:L_h_Bar_poly}, that is, $\Bar{L}_h^{\text{poly}} =  R_{\text{max}} \cdot n \sqrt{2^n} \left( \binom{n+R_{\text{max}}}{R_{\text{max}}} - 1 \right)$, and  $ r_{\text{PTR}}\left( \epsilon_{\text{PTR}} \right)$ as in Lemma \ref{lemma:T_PTR_contractive_Lipschitz}, i.e. $r_{\text{PTR}} (\epsilon_{\text{PTR}}) = 2 \sqrt{2} \left( \epsilon_{\text{PTR}}^2 - \epsilon_{\text{PTR}} \sqrt{ \sqrt{2} -1 } \right) + 1$.

    Here we have highlighted the dependence in the parameters of the reservoir class in the constants $C_0^P, C_1, C_2^P, C_3^P$ and $C_4^P$.

\medskip

Similarly, under the same hypotheses as in Theorem \ref{cor:risk_bound_rrr_O}, for all $\delta \in (0, 1)$, with probability at least $1 - \delta$ we have

    \begin{align}
    \label{eq:T_RRR_risk_bound_a}
       \sup_{H \in \mathcal{H}_{n, R_{\max}, C_{\max}}^{\text{RRR}}} & \left| R(H) - \hat{R}_m(H) \right| \\
       & \leq \left( C_0^{a,R} \left(  n, R_{\max}, \alpha_{\min}, r_{0,1} \right) + C_1 \left(  n, R_{\max}, \zeta_{\max} \right) \right) \frac{1}{m} \\
        & + C_2^R \left(   n, R_{\max}, \zeta_{\max}, C_{\max}, \alpha_{\min} \right) \frac{ \log{m} }{m} \\
        & + C_3^{a,R} \left( n, R_{\text{max}}, \zeta_{\text{max}}, \alpha_{\min}, \alpha_{\max}, r_{0,1}, C_{\max} \right) \sqrt{\frac{ \log{m} }{m}} \\
        & +  C_4^{a, R} \left(  n, R_{\max}, \zeta_{\max}, \alpha_{\min}, r_{0,1} \right) \sqrt{ \frac{\log{4/\delta}}{2 m} }
    \end{align}

    if $r_0 = r_1 = r_{0,1}$, and 

    \begin{align}
    \label{eq:T_RRR_risk_bound_b}
       \sup_{H \in \mathcal{H}_{n, R_{\max}, C_{\max}}^{\text{RRR}}} & \left| R(H) - \hat{R}_m(H) \right| \\
       & \leq \left( C_0^{b, R} \left(  n, R_{\max}, \epsilon_{\text{RRR}} \right) + C_1 \left(  n, R_{\max}, \zeta_{\max} \right) \right) \frac{1}{m} \\
        & + C_2^R \left(   n, R_{\max}, \zeta_{\max}, C_{\max}, \alpha_{\min} \right) \frac{ \log{m} }{m} \\
        & +C_3^{b,R} \left( n, R_{\text{max}}, \zeta_{\text{max}}, \alpha_{\min}, \alpha_{\max}, \epsilon_{\text{RRR}}, C_{\max} \right) \sqrt{\frac{ \log{m} }{m}} \\
        & + C_4^{b, R} \left( n, R_{\max}, \zeta_{\max}, \alpha_{\min}, \epsilon_{\text{RRR}} \right) \sqrt{ \frac{\log{4/\delta}}{2 m} }
    \end{align}

    if $r_0 > r_1$, where 

    \begin{align}
        & C_0^{a, R} \left(  n, R_{\max}, \alpha_{\min}, r_{0,1} \right) = 2 L_{\ell} \Bar{L}_h^{\text{poly}} \frac{\left( 1 - \left( 1 - \alpha_{\min} \right)^m r_{0,1}^m \right) \left( 1 - \alpha_{\min} \right)  r_{0,1}}{ 1 - \left( 1 - \alpha_{\min} \right) r_{0,1} } \\
        & C_0^{b, R} \left(  n, R_{\max}, \epsilon_{\text{RRR}} \right) = 2 L_{\ell} \Bar{L}_h^{\text{poly}} \frac{\left( 1 - \left( 1 - \epsilon_{\text{RRR}} \right)^m \right) \left( 1 - \epsilon_{\text{RRR}} \right) }{\epsilon_{\text{RRR}} } \\
        &  C_2^R \left(   n, R_{\max}, \zeta_{\max}, C_{\max}, \alpha_{\min} \right) \\
        & \hspace{5mm} = \frac{ 2 \left( L_{\ell} \Bar{L}_h^{\text{poly}} + \mathbb{E} \left[ | \ell (0, Y_0) | \right] + C_{\max} L_{\ell} \right) }{\log{\zeta_{\max}^{-1}}} + \frac{ 2 (1 - \alpha_{\min}) L_{\ell} \Bar{L}_h^{\text{poly}}C_v}{ \zeta_{\max} \log{\zeta_{\max}^{-1}} } \\
        &  C_3^{a,R} \left( n, R_{\text{max}}, \zeta_{\text{max}}, \alpha_{\min}, \alpha_{\max}, r_{0,1}, C_{\max} \right) \\ 
        & = \frac{2 L_{\ell}}{\sqrt{\log{\zeta_{\text{max}}^{-1}}}} 48 \cdot 0.89\left( \sqrt{4^n} + \sqrt{\mathcal{N}_{\text{poly}} + 1} \right) \cdot \\
        & \max \left\{ \frac{2}{(1 - \alpha_{\min})r_{0,1}} n\sqrt{2^n} R_{\max} \mathcal{N}_{\text{poly}} \sqrt{(\alpha_{\max} - \alpha_{\min})^2 + 2} \, , \, \sqrt{\left( \mathcal{N}_{\text{poly}} + 4 C_{\max}^2 \right) \left( \mathcal{N}_{\text{poly}} + 1 \right)} \right\} \\
        &  C_3^{b,R} \left( n, R_{\text{max}}, \zeta_{\text{max}}, \alpha_{\min}, \alpha_{\max}, \epsilon_{\text{RRR}}, C_{\max} \right) \\ 
        & = \frac{2 L_{\ell}}{\sqrt{\log{\zeta_{\text{max}}^{-1}}}} 48 \cdot 0.89\left( \sqrt{4^n} + \sqrt{\mathcal{N}_{\text{poly}} + 1} \right) \cdot \\
        & \max \left\{ \frac{2}{(1 - \epsilon_{\text{RRR}})} n\sqrt{2^n} R_{\max} \mathcal{N}_{\text{poly}} \sqrt{(\alpha_{\max} - \alpha_{\min})^2 + 2} \, , \, \sqrt{\left( \mathcal{N}_{\text{poly}} + 4 C_{\max}^2 \right) \left( \mathcal{N}_{\text{poly}} + 1 \right)} \right\} \\
        & C_4^{a, R} \left(  n, R_{\max}, \zeta_{\max}, \alpha_{\min}, r_{0,1} \right) \\
        & \hspace{5mm} = 2 L_{\ell} \frac{\Bar{L}_h^{\text{poly}}}{1 - (1 -  \alpha_{\min}) r_{0,1}} \left(  (1 -  \alpha_{\min}) r_{0,1} + 2 \left( 1 - \alpha_{\min} \right) M_{\xi} L_v \lVert w^v \rVert_1 \right) + M_{\xi} L_y \lVert w^y \rVert_1  \\
        & C_4^{b, R} \left( n, R_{\max}, \zeta_{\max}, \alpha_{\min}, \epsilon_{\text{RRR}} \right) \\
        & \hspace{5mm} = 2 L_{\ell} \frac{\Bar{L}_h^{\text{poly}}}{\epsilon_{\text{RRR}}} \left(  1 - \epsilon_{\text{RRR}} + 2 \left( 1 - \alpha_{\min} \right)  M_{\xi} L_v \lVert w^v \rVert_1 \right) + M_{\xi} L_y \lVert w^y \rVert_1 \\
    \end{align}

    and $\Bar{L}_h^{\text{poly}}$ is as in \eqref{eq:L_h_Bar_poly}, i.e. $\Bar{L}_h^{\text{poly}} =  R_{\max} \cdot n \sqrt{2^n} \left( \binom{n+R_{\max}}{R_{\max}} - 1 \right)$, and the constant $C_1$ is as in \eqref{eq:constants_PTR_C_1}.

    The Big-O bounds follows straightforwardly by considering the parameters that do not explicitly depend on the reservoir choice as constants (i.e. constants related to the Input-Output distribution as well as the choice of loss function).

\end{document}